%% file: TIP-DCV_2015_double_v1.tex
\documentclass[journal]{IEEEtran}

\usepackage{times}
\usepackage{epsfig}
\usepackage{epstopdf}
\usepackage{graphicx}
\usepackage{amsmath}
\usepackage{amssymb}
\usepackage{tabularx}
\usepackage{multirow}
\usepackage{subfigure}
\usepackage[justification=justified, font=small]{caption}
\usepackage{bm}
\usepackage{array}
\usepackage{multirow}
\usepackage{txfonts}
\usepackage{float}
\usepackage{stfloats}
\usepackage{color, soul}
\usepackage{color}
\usepackage{setspace}
 \usepackage{longtable}
 \usepackage{rotating}
 \usepackage{multirow}
\usepackage{cite}

\ifCLASSINFOpdf
\else
\fi
\graphicspath{{figures/}}

\begin{document}

\setstcolor{green}
%
\title{Detail-preserving and Content-aware Variational Multi-view Stereo Reconstruction}
%
%
%
\author{Zhaoxin~Li,
        Kuanquan~Wang,
        Wangmeng~Zuo,
        Deyu~Meng
        and~Lei Zhang 
\thanks{Z. Li is with School of Computer Science and Technology, Harbin Institute of Technology, Harbin, China, and Department of Computing, The Hong Kong Polytechnic University, Kowloon, Hong Kong. }
\thanks{K. Wang and W. Zuo is with School of Computer Science and Technology, Harbin Institute of Technology, Harbin, China.}
\thanks{D. Meng is with Institute for Information and System Sciences, Xi'an Jiaotong University, Xi'an, China.}
\thanks{L. Zhang is with Department of Computing, The Hong Kong Polytechnic University, Kowloon, Hong Kong.}
}
\maketitle

\begin{abstract}
Accurate recovery of 3D geometrical surfaces from calibrated 2D multi-view images is a fundamental yet active research area in computer vision. Despite the steady progress in multi-view stereo reconstruction, most existing methods are still limited in recovering fine-scale details and sharp features while suppressing noises, and may fail in reconstructing regions with few textures. To address these limitations, this paper presents a \emph{D}etail-preserving and \emph{C}ontent-aware \emph{V}ariational (DCV) multi-view stereo method, which reconstructs the 3D surface by alternating between reprojection error minimization and mesh denoising. In reprojection error minimization, we propose a novel inter-image similarity measure, which is effective to preserve fine-scale details of the reconstructed surface and builds a connection between guided image filtering and image registration. In mesh denoising, we propose a content-aware $\ell_{p}$-minimization algorithm by adaptively estimating the $p$ value and regularization parameters based on the current input. It is much more promising in suppressing noise while preserving sharp features than conventional isotropic mesh smoothing. Experimental results on benchmark datasets demonstrate that our DCV method is capable of recovering more surface details, and obtains cleaner and more accurate reconstructions than  state-of-the-art methods. In particular, our method achieves the best results among all published methods on the Middlebury {\it dino ring} and {\it dino sparse ring} datasets in terms of both completeness and accuracy.
\end{abstract}

%
\IEEEpeerreviewmaketitle

\section{INTRODUCTION}
\label{section1}
%
%
%
%
\IEEEPARstart{M}{ulti-view} stereo (MVS), which aims at inferring a scene's 3D geometric surface from a set of calibrated 2D images captured in different views, is a fundamental problem in computer vision. Due to its capability of high-quality reconstruction for both indoor and outdoor scenes, MVS has been widely used in science and engineering \cite{3,4,5}. Driven by the MVS benchmark datasets in \cite{1} and \cite{2}, various MVS algorithms have been proposed to gradually improve the accuracy and completeness of MVS reconstruction \cite{31a,49,20,2d}, and MVS remains an active research area that attracts considerable attentions \cite{2a,2d,2e}.

The performance of existing MVS methods is limited due to factors such as violation of the Lambertian reflectance model, inaccurate camera calibration, lack of textures on the object, and false matches. Therefore, noises are inevitable for the reconstructed 3D surface, resulting in degraded accuracy and  visually unpleasant artifacts. A number of methods, e.g., weighted minimal surface models~\cite{6,7}, have been proposed to suppress noises. However, this line of methods usually impose isotropic smoothness prior on 3D models, and tend to over-smooth fine-scale details and sharp features. 

To overcome these limitations, various methods have been developed to suppress noise while preserving sharp features. Based on the 3D model representation, these methods can be grouped into three categories, i.e., point cloud-based, volumetric-based, and mesh-based. For point cloud-based methods, smooth prior is introduced in \cite{48} to improve the accuracy of local matches on each stereo pairs. In~\cite{48a,31a}, accurate point clouds on high-textured regions are generated by deploying reliable features, and then propagated to the neighbouring regions. Besides, several heuristic strategies \cite{48b,48d} are suggested to evaluate the reliability of each point and remove noises based on local geometry orientation, photometric and visibility. For these methods \cite{48,48a,31a,48b,48d}, meshing point clouds are usually required to generate the final 3D surface, which may lead to over-smoothing in thinly protruding structures. Besides, noises and missing data of point clouds could be propagated to the meshing step, resulting in artifacts in the final reconstruction.

In volumetric-based methods, a photoflux term of photoconsistency \cite{16a} is introduced to provide data-driven ballooning force toward maximal photo-consistent surface. Such an energy term is helpful in segmenting thin structures, but fail to recover the structures on concave regions. Kolev et al.~\cite{2d} added a stereo regional term to enforce the background constraint based on a set of depth-maps. The regional term can be updated along with iterations to infer the occluded regions, making the method work well in recovering both the protrusion structure and concave regions. Kostrikov et al.~\cite{16b} further improved the method in~\cite{2d} by proposing a robust camera selection algorithm for labelling voxels as interior or exterior. However, high memory requirements of volumetric-based methods hamper their applications in large-volume and high-quality MVS reconstruction.

For mesh-based MVS, a number of variational methods \cite{2g,12,18,19,20,21,2e} have been proposed to improve the reconstruction quality. They can also be employed as a refinement step of other methods for high-quality reconstruction \cite{20,21,21a,31a}. However, most existing mesh-based methods adopt isotropic mesh smoothing, where the photoconsistency is computed by the zero-mean normalized cross-correlation (ZNCC). This makes them often fail in recovering the fine details and sharp features of object surface.

In addition to the above methods, other cues, e.g., silhouettes~\cite{12,13,14,15} and surface orientation~\cite{16}, can also be incorporated to help 3D reconstruction. The silhouettes of an object can be fused to ensure that the reconstructed surface preserves the protrusions and indentions.  They can be adopted in either volumetric-based \cite{13,14,15} or mesh-based methods~\cite{2i, 12, 12a}. However, the incorporation of silhouettes cannot guarantee the preservation of fine-scale surface details and sharp features that are not on the contour generators of surface. The surface orientation of the observed shape can be employed to design an anisotropic weight surface \cite{16}. However, the computation of surface orientation needs accurate second-order surface derivative, and the constant albedo assumption may not hold \cite{17,17a}, making surface orientation only be applicable to some restricted scenarios.

 \begin{figure*}[tbp]
 \centering
\scalebox{1.35}[1.35]{\includegraphics[height=28mm]{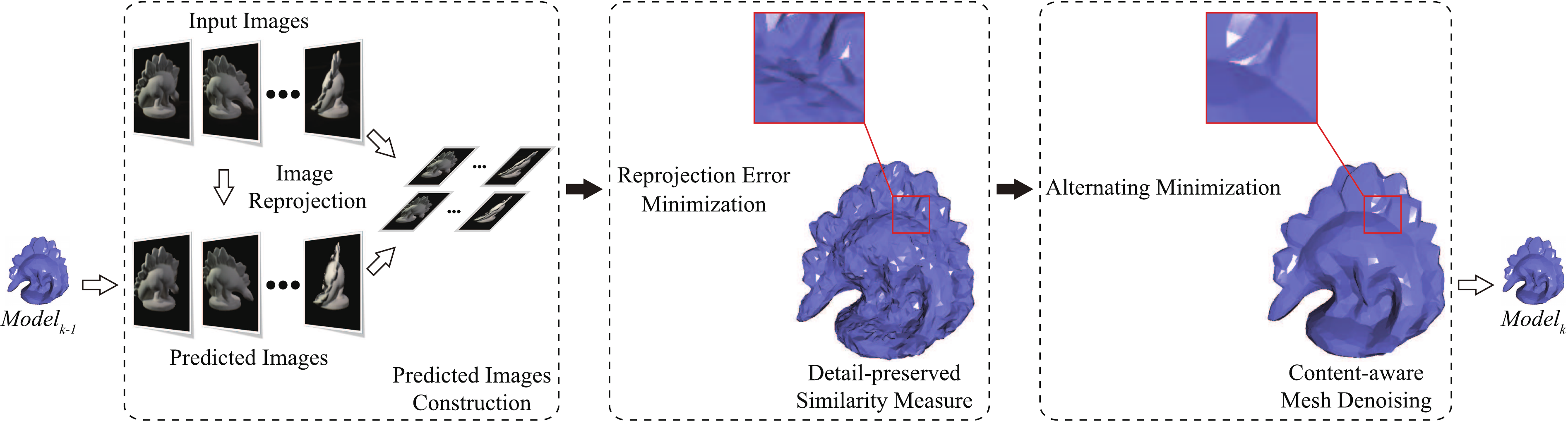}}
 \caption{Overview of the proposed DCV method.}
 \label{fig:DeCoSfig}
\vspace{-2mm}
 \end{figure*}
Compared with point cloud-based and volumetric-based methods, mesh-based methods are more feasible for reconstructing high-resolution surface with low memory requirements, but the accuracy of the mesh-based methods is generally limited, partially due to the use of isotropic mesh smoothing and ZNCC-based inter-image similarity measure. To address these issues, in this paper we propose a novel inter-image similarity measure and a content aware mesh denoising algorithm, resulting in a detail-preserving and content-aware variational (DCV) method for MVS. As shown in Fig. \ref{fig:DeCoSfig}, the contribution of this work is two-fold:
\begin{itemize}
    \item An inter-image similarity measure is proposed to preserve fine-scale details of the reconstructed surface. The proposed similarity measure also builds a connection between guided image filtering~\cite{22} and image registration, making our measure have promising edge-preserving performance.

    \item A content-aware $\ell_{p}$-minimization algorithm is proposed for mesh denoising. By adaptively estimating a suitable $p$ value and regularization parameters, our algorithm works very well in mesh smoothing while preserving sharp features.

\end{itemize}
Extensive experimental results on benchmark datasets validate the superiority of our DCV method in accurate 3D reconstruction. Moreover, DCV achieves the best results among all published methods on the Middlebury {\it dino ring} and {\it dino sparse ring} datasets in terms of both completeness and accuracy.

The paper is organized as follows. Section \ref{section2} introduces the related work. Section \ref{section3} briefly introduces the concept of reprojection error and its minimization. Section \ref{section4} presents the pipeline of our method and its two major components, i.e., detail-preserving similarity measure and content-aware mesh denoising, respectively. Section \ref{section5} presents the experimental results. Finally, the paper is concluded in \ref{section6}.


\section{Related Work}
\label{section2}

This section gives a brief survey on mesh-based MVS methods according to the two key elements which decide the quality of MVS reconstruction: data fidelity and regularization. The mesh-based MVS methods can provide high-resolution reconstruction with low memory requirements, and they are convenient to accelerate by using graphic hardware. Due to these advantages, mesh-based representation has been widely adopted in state-of-the-art MVS methods for 3D reconstruction of indoor \cite{2e, 12} and outdoor scenarios \cite{20, 21}, and surface refinement~\cite{20,21,31a}.

\textbf{Data fidelity.} Data fidelity is used to measure photometric consistency between images. In some early works \cite{2g, 2h, 12}, data fidelity is measured by comparing projections of surface points (or a planar patch tangent to surface point) with the corresponding neighbouring images, i.e., photoconsistency. The total consistency of the mesh surface is a summation of photoconsistency over all the mesh vertices. The main drawback of this measure is the projective distortion occurred in the high curvature regions of objects.  Another line of methods compare the image pixels with rendered surface textures \cite{2i, 17a} by implicitly assuming controlled lighting environments. The reprojection error minimization framework, which is also known as reprojection error functional\cite{2e, 23, 24, 12a}, attempts to solve the MVS problem by comparing the observed and predicted values of pixels generated from the reconstructed surface. The total consistency of the mesh surface is measured on image space instead of 3D surface space to alleviate projective distortion. It can also be considered as an image registration problem~\cite{25}, i.e., registering the input images and their predicted images. 

A similarity measure is needed to measure the reprojection error. In previous works, differentiable and isotropic similarity measures have been widely used, such as Zero-mean Normalized Cross Correlation (ZNCC)~\cite{18,20,21,25} and Sum of Square Difference (SSD)\cite{19,23,24}. However, these isotropic similarity measures tend to flatten or smoothen the sharp features of surface, and are limited in recovering fine-scale details. Edge-aware anisotropic methods can be used to replace the isotropic ones. Actually, anisotropic methods have been independently proposed in binocular stereo vision\cite{8,9,10,11}. Among them, guided image filtering has been used~\cite{9,10} due to its effectiveness and efficiency. However, the guided image filter in stereo vision is employed to filter discrete disparity space images (DSI), and cannot be directly adopted in the reprojection error minimization framework, where a variational measure is necessary. The proposed method fills the gap between guided image filtering-based anisotropic measure and variational-based image registration, and it is effective in reconstructing fine-scale details.

\textbf{Surface regularization:} Mesh regularization methods are introduced to improve the smoothness while preserving details of 3D surface, which can be divided into two categories, i.e., surface smoothing and denoising. For surface smoothing, several mesh smoothing operators, e.g., discrete Laplace-Beltrami operator~\cite{26}, have been adopted as band-pass filters in MVS. Other smoothing methods, such as mean curvature motion \cite{23, 24} and gradient flow \cite{2f}, have been studied and applied to mesh-based MVS \cite{2e,19}. To improve the computational efficiency, different approximations, such as Laplacian approximation \cite{2g, 12a,18}, umbrella operator, \cite{12} and paraboloid approximation \cite{17a}, have been proposed. Higher order derivatives, e.g., combination  of the first- and second-order Laplace \cite{31a}, thin-plate energy~\cite{20,21}, have been suggested to handle artificial shrinkage of small components and to penalize strong bending. However, higher order derivatives are not well defined at regions with sharp features and their computation is sensitive to noise.


Mesh denoising aims to remove the noises or spurious details while preserving sharp edge and corner features, which can be further classified into three sub-categories. The first one is based on bilateral filtering on vertexes~\cite{27}; the second one combines normal filtering and vertex position updating~\cite{28,29}; and the third one is based on optimizing an $\ell_{0}$ norm based non-convex energy function~\cite{30}. In this work, we propose a novel mesh-denoising method by considering the gradient distribution of surface meshes, which is formulated as an $\ell_{p}$-minimization problem in the {\it maximum a posteriori} (MAP) framework. Moreover, we adaptively select the $p$ value and regularization parameters, making our method content-aware to preserve sharp features.
\section{Prerequisites: Reprojection Error and Its Minimization}
\label{section3}
\begin{figure}[tbp]
\centering
 \subfigure[]{
  \label{fig:subfig:visiblefiga}
\includegraphics[height=26mm]{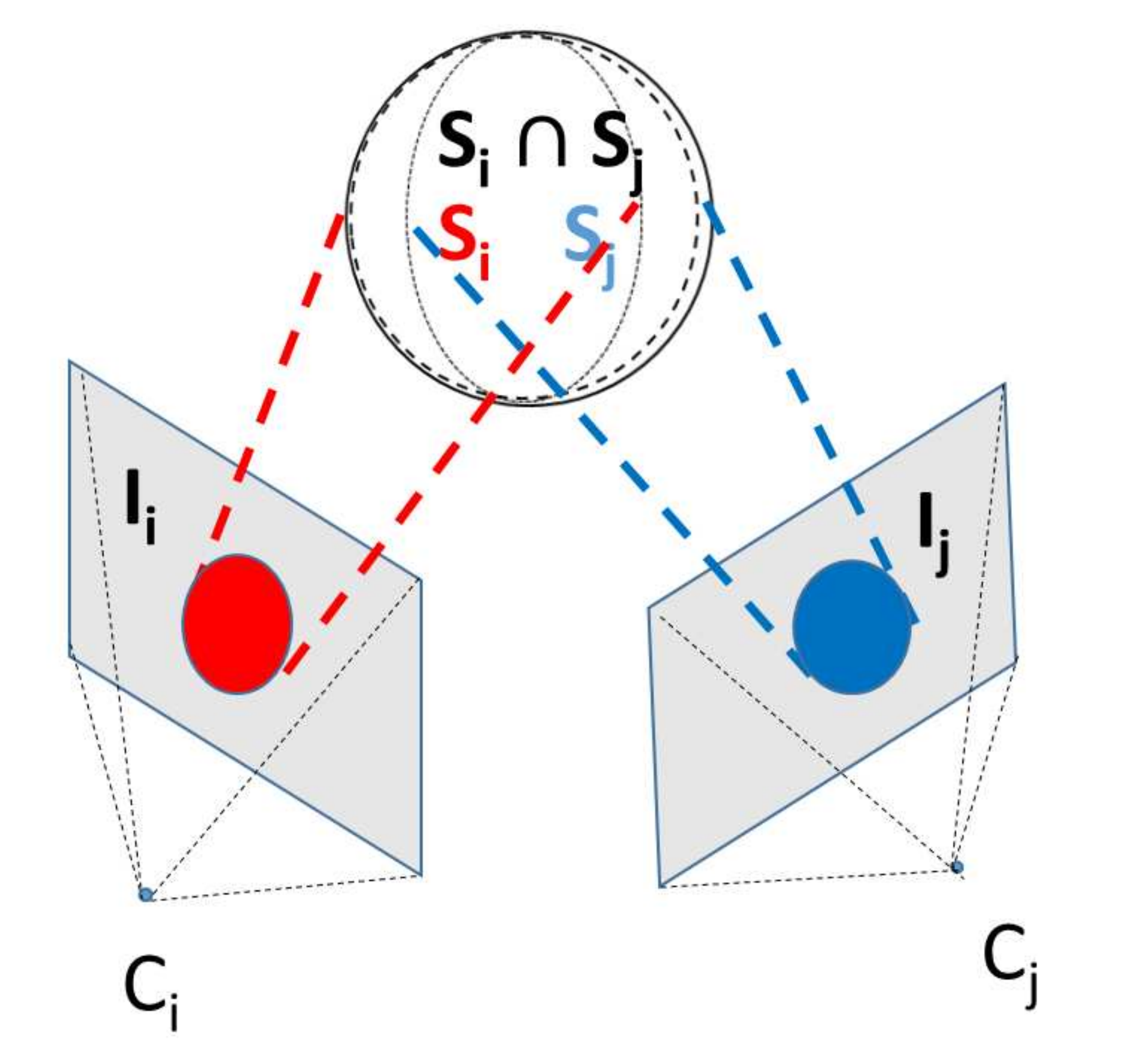}}
 \subfigure[]{
  \label{fig:subfig:visiblefigb}
\includegraphics[height=29mm]{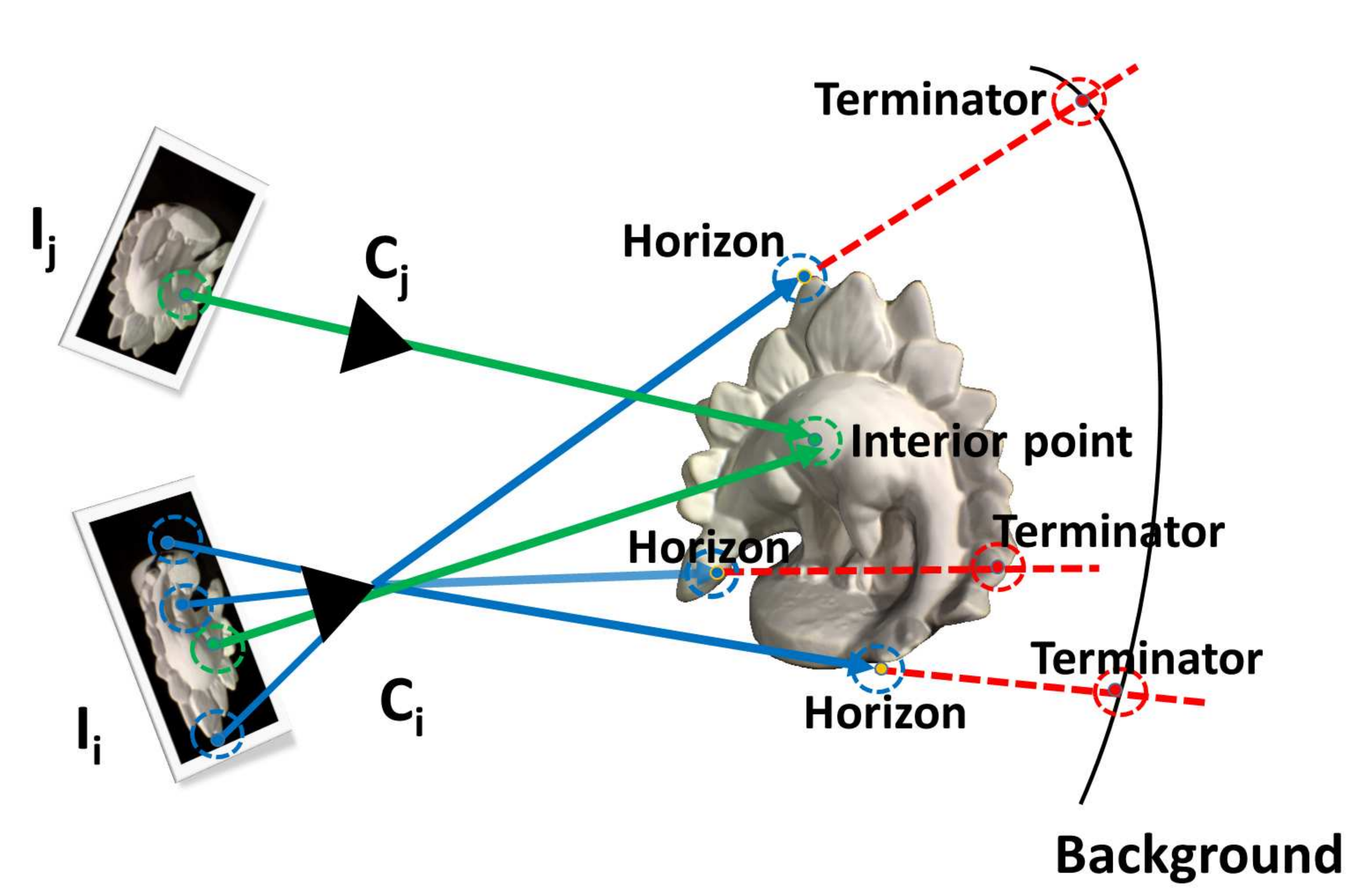}}
\vspace{-2mm}
\caption{Illustration of visibility of the surface. (a) Visible parts of surface for each camera: $S_{i}$ for camera $i$ (center $C_{i}$) and $S_{j}$ for camera $j$ (center $C_{j}$). $S_{i,j}$ is the shared visible part for cameras $i$ and $j$. (b) The interior points are these surface points which are visible from both stereo pair and not in the contour generators of the surface. The horizons are the points in the contour generators for a specific camera. The terminators are occluded by horizons and are behind the horizons along with the camera ray tracing.}
\label{fig:visiblefig}
\vspace{-3mm}
\end{figure}

Let $S\subset\mathbb{R}^{3}$ denote a reconstructed surface of object, $B\subset\mathbb{R}^{3}$ stand for its background, and $I_{i}: \Omega_{i} \subset \mathbb{R}^{2} \to \mathbb{R}^{d}$ denote the observed (input) image in camera $i$ ($d=1$ for grayscale images, and $d=3$ for color images). In image formation, the observed image only records the visible (i.e., unoccluded) part of a real scene, which includes foreground from both the interested object and the irrelevant background. As shown in Fig. \ref{fig:subfig:visiblefiga}, $S_{i}$ is the visible part of surface for camera $i$. We define $S_{i,j}$ as the shared visible surface of camera $i$ and camera $j$.  Let $\pi_{i}:\mathbb{R}^{3}\to\Omega_{i}$ be the perspective projection which projects 3D point $\textbf{x}$ to 2D pixel $\textbf{p}$. Let $\hat{I}_{i,S,B}$ be the predicted image of $I_{i}$ via surface and background. $\hat{I}_{i,S}$ is the predicted image for object part and $\hat{I}_{i,B}$ is the predicted image for background part.
With the desired 3D reconstruction of object $S$ and background $B$, it is natural to assume that the image predicted by 3D object and background models should be similar to the observed image. Therefore, the minimization framework of reprojection errors 
adopts the following functional\cite{19,23,24,2e}:
\begin{equation}
\begin{aligned}
&E_{\text{im}}(S)= \sum\nolimits_{i}\int_{\Omega_{i}}g(I_{i}(\textbf{p}),\hat{I}_{i,S,B}(\textbf{p}))d\textbf{p}\\
&=  \sum\nolimits_{i}\Big[ \int_{\pi_{i}\circ  S_{i}}g^i_{F}(I_{i}(\textbf{p}),\hat{I}_{i,S}(\textbf{p}))d\textbf{p} 
+\int_{\Omega_{i}-\pi_{i}\circ  S_{i}}g^i_{B}(I_{i}(\textbf{p}),\hat{I}_{i,B}(\textbf{p}))d\textbf{p} \Big],
\end{aligned}
\label{eq:reprojectionfunc1}
\end{equation}
where $\pi_{i}\circ S_{i}$ denotes the projection of surface $S_{i}$ onto $I_{i}$, the reprojection error $g(I,J)(\textbf{p})$ denotes the similarity measure between images $I$ and $J$ at pixel $\textbf{p}$. The reprojection error $g^i_{F}(I_{i}(\textbf{p}),\hat{I}_{i,S}(\textbf{p}))$ measures the similarity between image $I_{i}$ and its predicted image via surface of the object, and the reprojection error $g^i_{B}(I_{i}(\textbf{p}),\hat{I}_{i,B}(\textbf{p}))$ measures similarity  between image $I_{i}$ and predicted background image.

The predicted image can be generated via rendering surface and background. In particular, $\hat{I}_{i,S}$ is defined based on stereo pairs and is usually not a single image. One of the predicted images of $I_{i}$ can be computed by first projecting its neighbouring image $I_{j}$ onto the reconstructed surface $S$ and then projecting to image space of camera $i$, which actually defines a predicted image $\hat{I}_{i,j,S}$. The valid definition domain for $\hat{I}_{i,j,S}$ is the projection of shared visible surface $S_{i,j}$, i.e., $\pi_{i}\circ S_{i,j}$.  By counting all the neighbouring images of $I_{i}$, $g^i_{F}$ is defined as:
\begin{equation}
g^i_{F}(\textbf{p})=\sum\nolimits_{j}m(I_{i},\hat{I}_{i,j,S})(\textbf{p})=\sum\nolimits_{j}m_{i,j}(\textbf{p}),
\label{eq:similarity_F}
\end{equation}
where $m$ is a similarity measure of two pixels in a small squared window centered on $\textbf{p}$. The definition domain of predicted image $\hat{I}_{i,B}$ of  $I_{i}$ via background is defined by $\Omega_{i}-\pi_{i}\circ  S_{i}$, i.e., the supplementary set of $\pi_{i}\circ S_{i}$. To simplify the computation, we assume that the background is uniformly black, and this can be implemented by segmenting silhouettes from observed images. Based on the fact that  $d\textbf{u}=-{\textbf{x}\cdot\textbf{n}(\textbf{x})}/ {\textbf{x}^3_{z}}d\textbf{s}$, with simple algebra, Eq. \eqref{eq:reprojectionfunc1} can be rewritten as an integral over the surface by counting only the visible points (see \cite{24} for details):
\begin{equation}
E_{\text{im}}(S)=\sum\nolimits_{i}\int_{S}-\dfrac{\textbf{x}\cdot\textbf{n}(\textbf{x})}{\textbf{x}^3_{z}}\left[ g^i_{F}(\textbf{x},\textbf{n}(\textbf{x}))-g^i_{B}(\textbf{x},\textbf{n}(\textbf{x}))\right]  {\Lambda}_{i,S}(\textbf{x})d\textbf{s},
\label{eq:reprojectionfunc2}
\end{equation}
where $\textbf{n}(\textbf{x})$ is the outward normal of surface $S$ on point $\textbf{x}$, and $\Lambda_{i,S}: \mathbb{R}^{3}\to\left [ 0,1\right]$ is the visibility function which equals to 1 if \textbf{x} is visible from camera $i$ and 0 otherwise. 

The functional of reprojection errors in Eq. \eqref{eq:reprojectionfunc2} can be reformulated on  mesh-based discrete representation. Let's  parametrize surface $S$ to a triangle mesh $M$ with a set of indexes $\mathcal{V}=\{v_{1},v_{2},\dots v_{n}\}$ and a set of triangular faces  $\mathcal{F}=\{f_{1},f_{2},\dots,f_{n}\}$, $f_{i}\in \mathcal{V}\times\mathcal{V}\times\mathcal{V}$. The geometric embedding of a triangle mesh into $\mathbb{R}^3$ is specified by associating each vertex to a 3D position. Let $\textbf{x}_{i}$ denote the position of a vertex $v_{i}$. Over each triangular face, points are parametrized using barycentric coordinates $\textbf{x}(\textbf{u}):\textbf{u}=(\it u,\it v)\in T=\{(u,v)|u \in [0,1], v \in [0,1-u]\}$. The energy functional on the triangle mesh is formulated as follows:
\begin{equation}
E_{\text{im}}(M)=\sum\nolimits_{i}\sum\nolimits_{k}A_{k}\int_{T}G(\textbf{x})N_{k} \Lambda_{i,S}(\textbf{x})d\textbf{u}
\label{eq:reprojectionfunc3},
\end{equation} 
where $G(\textbf{x})=-(\textbf{x}/\textbf{x}^3_{z})\cdot \left[ g^i_{F}(\textbf{x},\textbf{n}(\textbf{x}))-g^i_{B}(\textbf{x},\textbf{n}(\textbf{x}))\right]$, $N_{k}$ and $A_{k}$ are the normal and area of the triangle $f_{k}$, respectively, and the term $d\textbf{u}=2A_{k}ds$ corresponds to the unit surface area element in the triangle mesh.

The energy functional in Eq. \eqref{eq:reprojectionfunc3} can be optimized by using gradient decent over all the vertices of the mesh. According to \cite{19,23,24,2e}, the evolution equation for gradient decent flow is:
\begin{equation}
\begin{cases} 
\textbf{x}_{k}(0)=\textbf{x}^0\\
d\textbf{x}_{k}/dt=-(1/A_{k})\left[M_{k}^\text{int}+M_{k}^\text{horiz}\right]
\end{cases},
\label{eq:evolution_eq}
\end{equation}
where  $M_{k}^\text{int}$ is defined as:
\begin{equation}
V_{k}\sum\nolimits_{k}2A_{k}N_{k}\int_{T}\bigtriangledown{G} (\textbf{x})(1-u-v)dudv
\label{eq:evolution_eq_int},
\end{equation}
and $M_{k}^\text{horiz}$ is defined as:
\begin{equation}
\sum\nolimits_{\text{horizon edges}H_{k,j}}\int_{u\in[0,1]}\dfrac{1}{2}[G(T(\textbf{y}))-G(\textbf{y})]\dfrac{\textbf{y}\wedge H_{k,j}}{|\textbf{y}|[\textbf{y}]^3_{z}}(1-u)du
\label{eq:evolution_eq_horiz},
\end{equation}
where $V_{k}$ is the velocity vector, $H_{k,j}$ is the vector such that $\left [x_{k},x_{k}+H_{k,j}\right ]$ is the edge of the triangular face $f_{j}$ generating the horizon, $\textbf{y}$ is defined as $\textbf{y}=x_{k}+uH_{k,j}$, and $T(x)$ is the terminator of $\textbf{x}$. The definitions of Horizon and Terminator are illustrated in Fig. \ref{fig:subfig:visiblefigb}. $M_{k}^\text{int}$ is the gradient for the vertex (interior point) that does not change its visibility state, and $M_{k}^\text{horiz}$ is the gradient for the vertex that exihibits strong changes in  visibility  during the evolution.

The term $M_{k}^\text{horiz}$ is used to confine the horizons of the surface in different cameras. Although its influence will be considerably decreased by the introduction of surface regularization, this term is very useful for persevering thin protruding structures on the border between object and background. This naturally corresponds to a silhouette constriant \cite{12,12a,13,14,15}. The form of $g^i_{B}$ decides the consistency of reconstructed model with silhouettes, and SSD can be used to measure this error.

The term $M_{k}^\text{int}$ is crucial to the reconstruction quality. To evolve the current surface $S$, we should estimate the derivative of $g^i_{F}(\textbf{x})$. As shown in \cite{25}, the gradient of the similarity measure $m_{i,j}$ with respect to an infinitesimal vector displacement $\delta S$ of 3D surface point $\textbf{x}$ can be computed using the chain rule:
\begin{equation}
\lim\limits_{\epsilon \rightarrow 0}\dfrac{\partial m_{i,j}(S+\epsilon\delta S)}{\partial\epsilon}=\lim\limits_{\epsilon \rightarrow 0}\int_{\pi_{i}\circ S_{i,j}}\dfrac{\partial m(I_{i},\hat{I}_{i,j,S})}{\hat{I}_{i,j,S}(\textbf{p}_{j})}\times\dfrac{\partial \hat{I}_{i,j,S}}{\partial \textbf{p}_{j}}
\times\dfrac{\partial \textbf{p}_{j}}{\partial \textbf{x}}\times\dfrac{\partial\pi_{i,S}^{-1}}{\partial\epsilon}d\textbf{p}_{i},
\label{eq:pf3}
\end{equation}
and we have
\begin{equation}
\bigtriangledown g^i_{F}(S)(\textbf{x})=- \sum_{j,i\ne j}\eta_{\pi_{i}\circ S_{i,j}}\dfrac{\partial m(I_{i},\hat{I}_{i,j,S})}{\hat{I}_{i,j,S}(\textbf{p}_{j})}\times\dfrac{\partial \hat{I}_{i,j,S}}{\partial \textbf{p}_{j}}\times\dfrac{\partial \textbf{p}_{j}}{\partial \textbf{x}}\times\dfrac{\textbf{d}_{i}}{z_{i}^3} \textbf{n},
\label{eq:pf4}
\end{equation}
where $\textbf{p}_{i}$ and $\textbf{p}_{j}$  are the pixel positions in images $I_{i}$ and $\hat{I}_{i,j,S}$, respectively, $\pi_{i,S}^{-1}:\mathbb{R}^{3}\to\Omega_{i}$ is the inverse projection which projects pixels from camera $i$ onto the surface, and  $\textbf{d}_{i}$ is the vector joining the center of camera $i$ and $\textbf{x}$, $\eta$ is the Kronecker symbol which cancels the gradient computation in the region outside the shared visible surface of both cameras. When the surface moves, the predicted image tends to be changed. Hence, the variation of reprojection errors involves the derivative of the similarity measure with respect to its second argument $\hat{I}_{i,j,S}$, i.e.,  $\partial_{2}m_{i,j}$,   as shown in the first derivative term of the right part of Eq. \eqref{eq:pf4}. Therefore, the variation of predicted images will affect much the 3D shape of surface.

\section{Proposed Method}
\label{section4}

Following the mesh-based MVS framework, our DCV model consists of two terms, i.e., data fidelity $E_{\text{im}}$ and surface regularization $E_{\text{reg}}$. The energy functional of our model can be formulated as:
\begin{equation}
E(S)= E_{\text{im}}(S) + \lambda E_{\text{reg}}(S)
\label{eq:energy}
\end{equation}
where $S$ denotes the reconstructed surface of the object, and $\lambda$ is the trade-off parameter. Note that $E_{\text{im}}$ usually is differentiable while $E_{\text{reg}}$ is non-smooth. The model can be solved by extending the proximal gradient algorithm \cite{Beck09fista}, which iteratively performs the following two steps.

\textbf{Step 1. Gradient Descent.} Given the current estimate $S^{k}$, the gradient descent algorithm is adopted to minimize the data fidelity term $E_{\text{im}}$:
\begin{equation}
S^{k+0.5} = S^{k+0.5} - \eta\partial E_{\text{im}}(S)/\partial S,
\label{eq:update_im}
\end{equation}
where $\eta$ is the stepsize.

\textbf{Step 2. Surface Denoising.} Given $S^{k+0.5}$, the reconstructed surface $S$ is further refined by solving the following mesh denoising problem:
\begin{equation}
S^{k+1} = \arg \min_{S} \frac{1}{2} \| S - S^{k+0.5} \|^2 + \lambda \eta E_{\text{reg}}(S).
\label{eq:update_denoising}
\end{equation}

Given the nonsmooth convex function $E_{\text{reg}}$ and the smooth convex function $E_{\text{im}}$ with Lipschitz constant $L$, when the stepsize $\eta \leq 1/L$ and the surface denoising problem has the global solution, the algorithm can converge to the global optimum \cite{Beck09fista}. For our case, even $E_{\text{reg}}$ is nonconvex, our algorithm empirically converges to a satisfactory solution. In this work, we propose a detail-preserving similarity measure for \textbf{Step 1} and propose a content-aware mesh denoising algorithm for \textbf{Step 2}, which will be described in detail in the following two sub-sections, respectively.

\subsection{Detail-preserving Inter-image Similarity Measure}
\label{section4.1}

The similarity measure $m(I_{i}, \hat{I}_{i,j,S})$ is critical for the minimization of reprojection error between  $I_{i}$ and its predicted image $\hat{I}_{i,j,S}$ in $\pi_{i}\circ S_{i,j}$. In the variational framework, it is desirable that the similarity measure $m(I_{i}, \hat{I}_{i,j,S})$ is differentiable. Among the existing similarity measures~\cite{33,33a,33b}, zero-mean normalized cross correlation (ZNCC) is the most commonly used one due to the following advantages: (1) it is robust to inter-image affine illumination variation; and (2) its derivative can be efficiently computed. However, the isotropic property of ZNCC treats all pixels equally and prefers to flatten the details of surface. In this section, we first review the derivative of ZNCC-based  similarity measure and then propose a detail-preserving similarity measure based on the principle of guided image filtering.

\subsubsection{Derivative of ZNCC-based Similarity Measure}
\label{section4.1.1}

The ZNCC measure is defined as follows:
\begin{equation}
m(I_{1},I_{2})(\textbf{p})=v_{1,2}(\textbf{p})/\sqrt{v_{1}((\textbf{p}))v_{2}(\textbf{p})},
\label{eq:De1}
\end{equation}
where $v_{1}$, $v_{2}$ and $v_{1,2}$ are given by
\begin{equation}
v_{i}(\textbf{p})=G_{\sigma} \star I_{i}^2(\textbf{p})/\omega(\textbf{p})-\mu_{i}^2(\textbf{p})+\epsilon,
\label{eq:De2}
\end{equation}
\begin{equation}
v_{1,2}(\textbf{p})=G_{\sigma}\star I_{1}I_{2}(\textbf{p})/\omega((\textbf{p}))-\mu_{1}\mu_{2}(\textbf{p}),
\label{eq:De3}
\end{equation}
\begin{equation}
\mu_{i}(\textbf{p})=G_{\sigma}\star I_{i}(\textbf{p})/\omega(\textbf{p}).
\label{eq:De4}
\end{equation}
where $G_{\sigma}$ is a Gaussian kernel with standard deviation $\sigma$, $\omega$ is a normalization coefficient accounting for the shape of support domain: $\omega=\int_{\pi_{i}\circ S_{i,j}}G_{\sigma}(p-q)dq$, and the small positive constant $\epsilon$ is introduced to prevent the denominator from being zero. The derivative of $m(I_{1},I_{2})$ with respect to any entry of $I_{2}$ at pixel position $\textbf{p}$ has the following form \cite{25}:
\begin{equation}
{\partial_{2}m(\textbf{p})} = \alpha I_{1}(\textbf{p})+\beta I_{2}(\textbf{p})+\gamma(\textbf{p}),
\label{eq:De5}
\end{equation}
\begin{equation}
\alpha(\textbf{p})=G_{\sigma} \star\dfrac{-1}{\omega\sqrt{v_{1}v_{2}}}(\textbf{p}),
\label{eq:De6}
\end{equation}
\begin{equation}
\beta(\textbf{p})=G_{\sigma}\star\dfrac{g}{\omega v_{2}}(\textbf{p}),
\label{eq:De7}
\end{equation}
\begin{equation}
\gamma(\textbf{p})=G_{\sigma}\star(\dfrac{\mu}{\omega \sqrt{v_{1}v_{2}}}-\dfrac{\mu_{2}g}{\omega v_{2}})(\textbf{p}).
\label{eq:De7_a}
\end{equation}
Note that the variation at $\textbf{p}$ also tends to affect the similarity measure of its neighboring position. Actually, if we restrict ZNCC in a local square window of size $w$, the variation of pixel $\textbf{p}$ will affect its entire neighbouring pixels in the region of size $2w\times 2w$.

\subsubsection{Detail-preserving Similarity Measure Based on Guided Image Filtering}
\label{section4.1.2}
Let $I_{1}$ be the filtering input, and $I_{2}$ be the guidance image. The principle of guided image filtering is to assume a local linear transformation between filtering output $Q$ and a guidance image $I_{2}$ for any pixel $\textbf{p}$ belonging to a local window  with size of $w_{k}$ ($k$ is the center of window):
\begin{equation}
Q(\textbf{p})=a(\textbf{p})I_{2}(\textbf{p})+b(\textbf{p}).
\label{eq:De8}
\end{equation}
By minimizing the difference between $Q$ and $I_{1}$, we can obtain parameters $a(\textbf{p})$ and $b(\textbf{p})$:
\begin{equation}
a(\textbf{p})=(v_{12}/v_{2})(\textbf{p}),
\label{eq:De9}
\end{equation}
\begin{equation}
b(\textbf{p})=(\mu_{1}-a\mu_{2})(\textbf{p})=(\mu_{1}-v_{12}\mu_{2}/v_{2})(\textbf{p}).
\label{eq:De10}
\end{equation}
Note that the tolerance $\epsilon$ in Eq. \eqref{eq:De2} can also be included to penalize large $a(\textbf{p})$ in \eqref{eq:De9} and \eqref{eq:De10}. The role of $\epsilon$ in the guided filter is similar to the range variance $\sigma^2_{r}$ in the bilateral filter, which determines the edge patch that should be preserved. Finally, the filtering output $Q$ has the following form:

\begin{equation}
Q(\textbf{p})=\frac{G_{\sigma}\star a}{\omega}(\textbf{p})I_{2}(\textbf{p})+\frac{G_{\sigma}\star b}{\omega}(\textbf{p}).
\label{eq:De11}
\end{equation}

We can then have an interesting connection between the derivatives of reprojection errors and guided image filtering. Based on \eqref{eq:De1}, \eqref{eq:De6}-\eqref{eq:De7_a}, Eq. \eqref{eq:De5} can be reformulated as:
\begin{equation}
\begin{aligned}
\partial_{2}m(\textbf{p})=(G_{\sigma}\star\dfrac{-1}{\omega\sqrt{v_{1}v_{2}}}(\textbf{p}))I_{1}(\textbf{p})+(G_{\sigma}\star\dfrac{v_{1,2}}{\omega v_{2}\sqrt{v_{1}v_{2}}}(\textbf{p}))I_{2}(\textbf{p})\\
+G_{\sigma}\star (\dfrac{\mu_{1}}{\omega\sqrt{v_{1}v_{2}}}-\dfrac{\mu_{2}v_{1,2}}{\omega  v_{2}\sqrt{v_{1}v_{2}}})(\textbf{p}).
\end{aligned}
\label{eq:De12}
\end{equation}
Based on \eqref{eq:De9}-\eqref{eq:De10}, Eq. \eqref{eq:De11} can be rewritten as
\begin{equation}
Q(\textbf{p})=(G_{\sigma}\star \dfrac{v_{12}}{\omega v_{2}}(\textbf{p}))I_{2}(\textbf{p})+G_{\sigma}\star\dfrac{(v_{2}\mu_{1}-v_{12}\mu_{2})}{\omega v_{2}}(\textbf{p}).
\label{eq:De13}
\end{equation}
Let $v_{1}(\textbf{p})=v_{2}(\textbf{p})$, Eq. \eqref{eq:De12} becomes:
\begin{equation}
\begin{aligned}
\partial_{2}m(\textbf{p})=(G_{\sigma}\star\dfrac{-1}{\omega v_{2}}(\textbf{p}))I_{1}(\textbf{p})+(G_{\sigma}\star\dfrac{v_{1,2}}{\omega v_{2}^2}(\textbf{p}))I_{2}(\textbf{p}) \\
+G_{\sigma}\star (\dfrac{\mu_{1}}{\omega v_{2}}-\dfrac{\mu_{2}v_{1,2}}{\omega  v_{2}^2})(\textbf{p}).
\end{aligned}
\label{eq:De14}
\end{equation}
We can then have:
\begin{equation}
\begin{aligned}
\partial_{2}m(\textbf{p})+(G_{\sigma}\star\dfrac{1}{\omega v_{2}}(\textbf{p}))I_{1}(\textbf{p})=(G_{\sigma}\star\dfrac{v_{1,2}}{\omega v_{2}^2}(\textbf{p}))I_{2}(\textbf{p}) \\
+G_{\sigma}\star\dfrac{(v_{2}\mu_{1}-v_{12}\mu_{2})}{\omega v_{2}^2}(\textbf{p}).
\end{aligned}
\label{eq:De15}
\end{equation}

Suppose that $v_{2}(\textbf{p})$ varies more slowly than $G_{\sigma}$ and $I_{1}(\textbf{p})$ in spatial domain, and $(G_{\sigma}\star\dfrac{1}{\omega }(\textbf{p})) \approx 1$. Then, Eq. \eqref{eq:De15} can be approximately rewritten as:
\begin{equation}
v_{2} \partial_{2}m(\textbf{p}) + I_{1}(\textbf{p}) \approx (G_{\sigma}\star\dfrac{v_{1,2}}{\omega v_{2}}(\textbf{p}))I_{2}(\textbf{p})\\
+G_{\sigma}\star\dfrac{(v_{2}\mu_{1}-v_{12}\mu_{2})}{\omega v_{2}}(\textbf{p}).
\label{eq:De15_app}
\end{equation}
Note that the right sides of Eq. \eqref{eq:De13} and Eq. \eqref{eq:De15} are the same, and the minimization of reprojection error is actually the maximization of similarity measure. Therefore, we have
\begin{equation}
I_{1}(\textbf{p}) + v_{2} \partial_{2}m(\textbf{p}) \approx Q(\textbf{p}),
\label{eq:Equivalent}
\end{equation}
and guided image filtering can be approximately interpreted as one step of variational image registration of $I_{1}(\textbf{p})$ and $I_{2}(\textbf{p})$ with constraint $v_{1}(\textbf{p})=v_2(\textbf{p})$ and stepsize $v_{2}$.

\begin{figure}[tbp]
\centering
 \subfigure[]{
  \label{fig:subfig:detailenhancefiga}
\includegraphics[height=18mm]{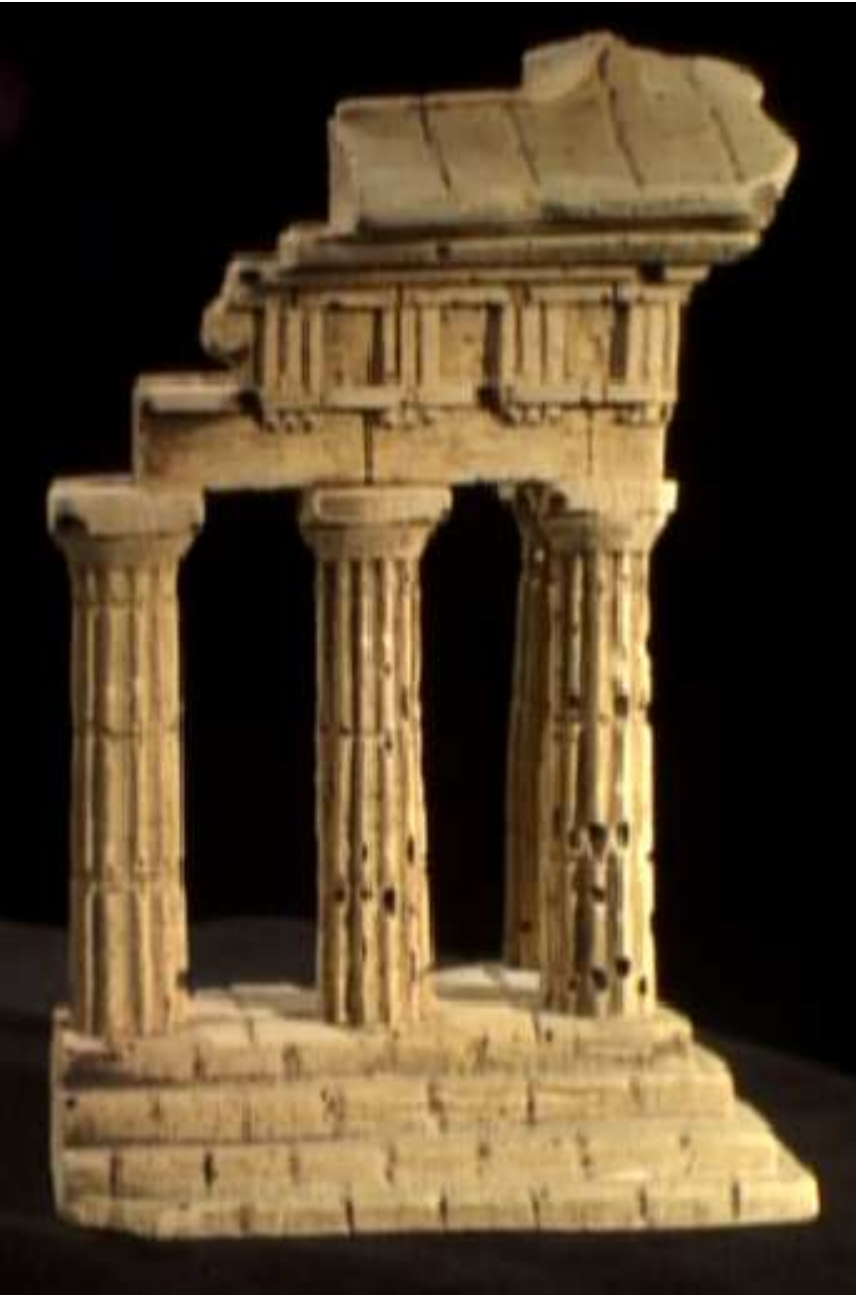}}
 \subfigure[]{
  \label{fig:subfig:detailenhancefigb}
\includegraphics[height=25mm]{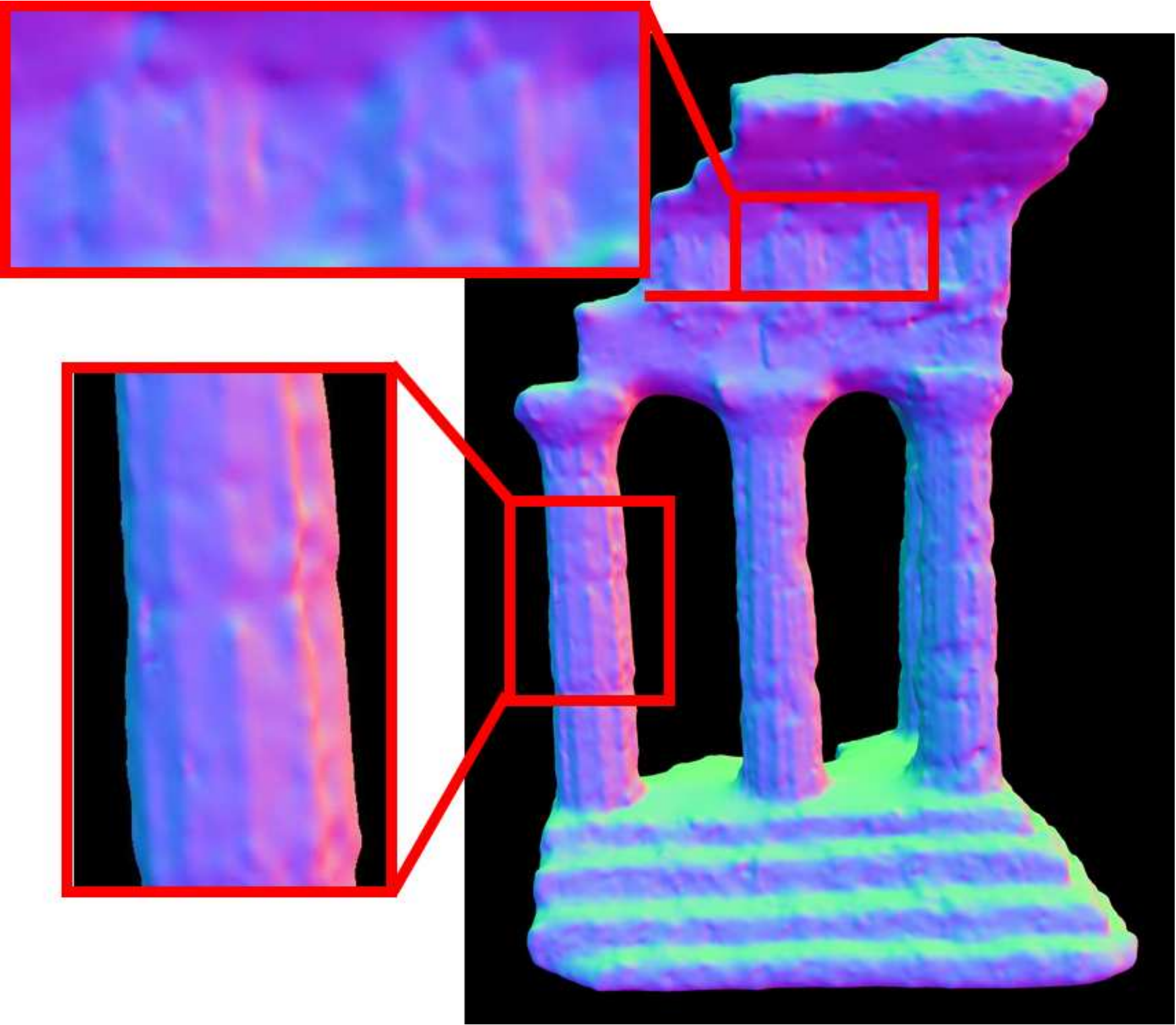}}
 \subfigure[]{
  \label{fig:subfig:detailenhancefigc}
\includegraphics[height=25mm]{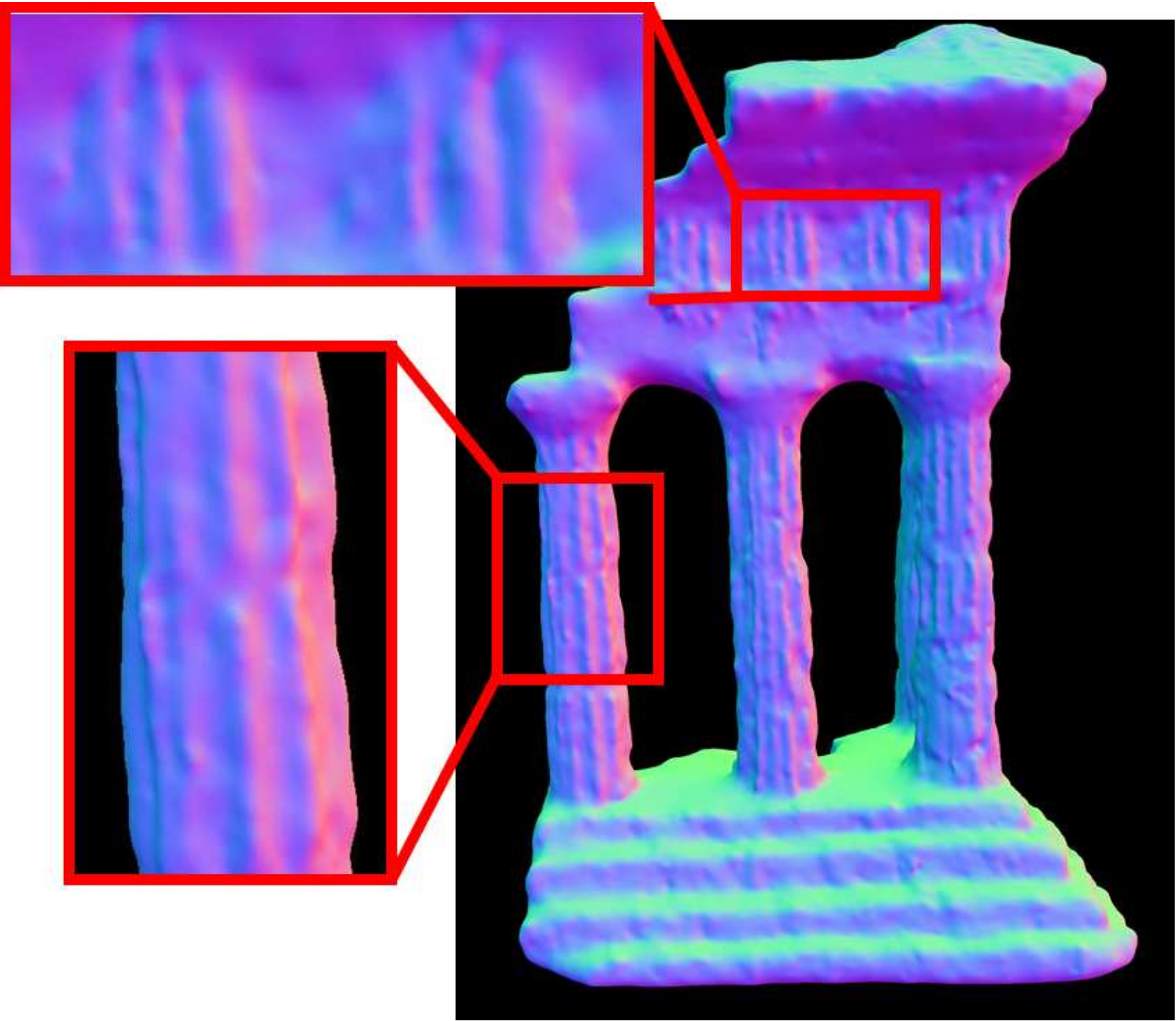}}
 \subfigure[]{
  \label{fig:subfig:detailenhancefigd}
\includegraphics[height=20mm]{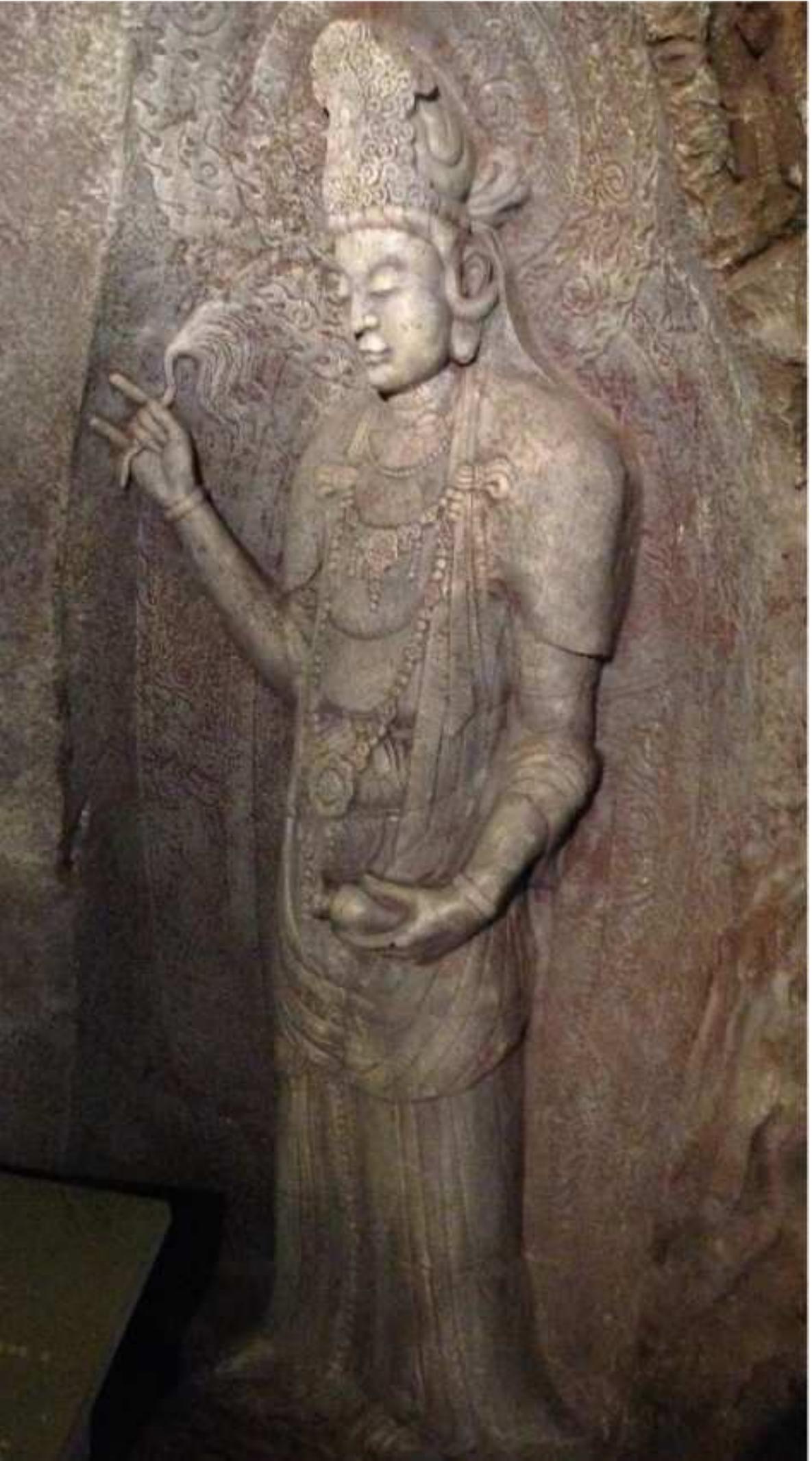}}
 \subfigure[]{
  \label{fig:subfig:detailenhancefige}
\includegraphics[height=26mm]{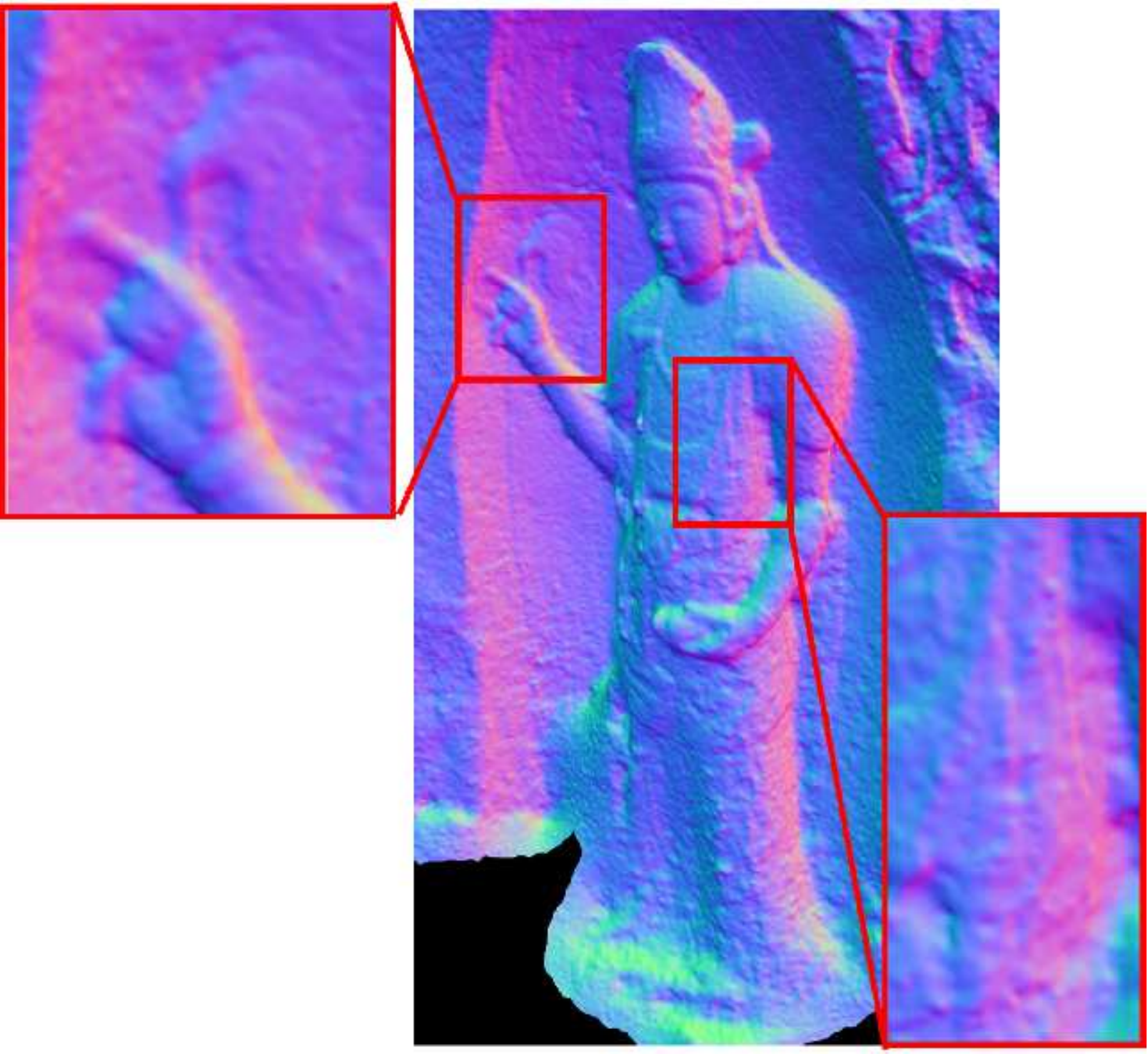}}
 \subfigure[]{
  \label{fig:subfig:detailenhancefigf}
\includegraphics[height=26mm]{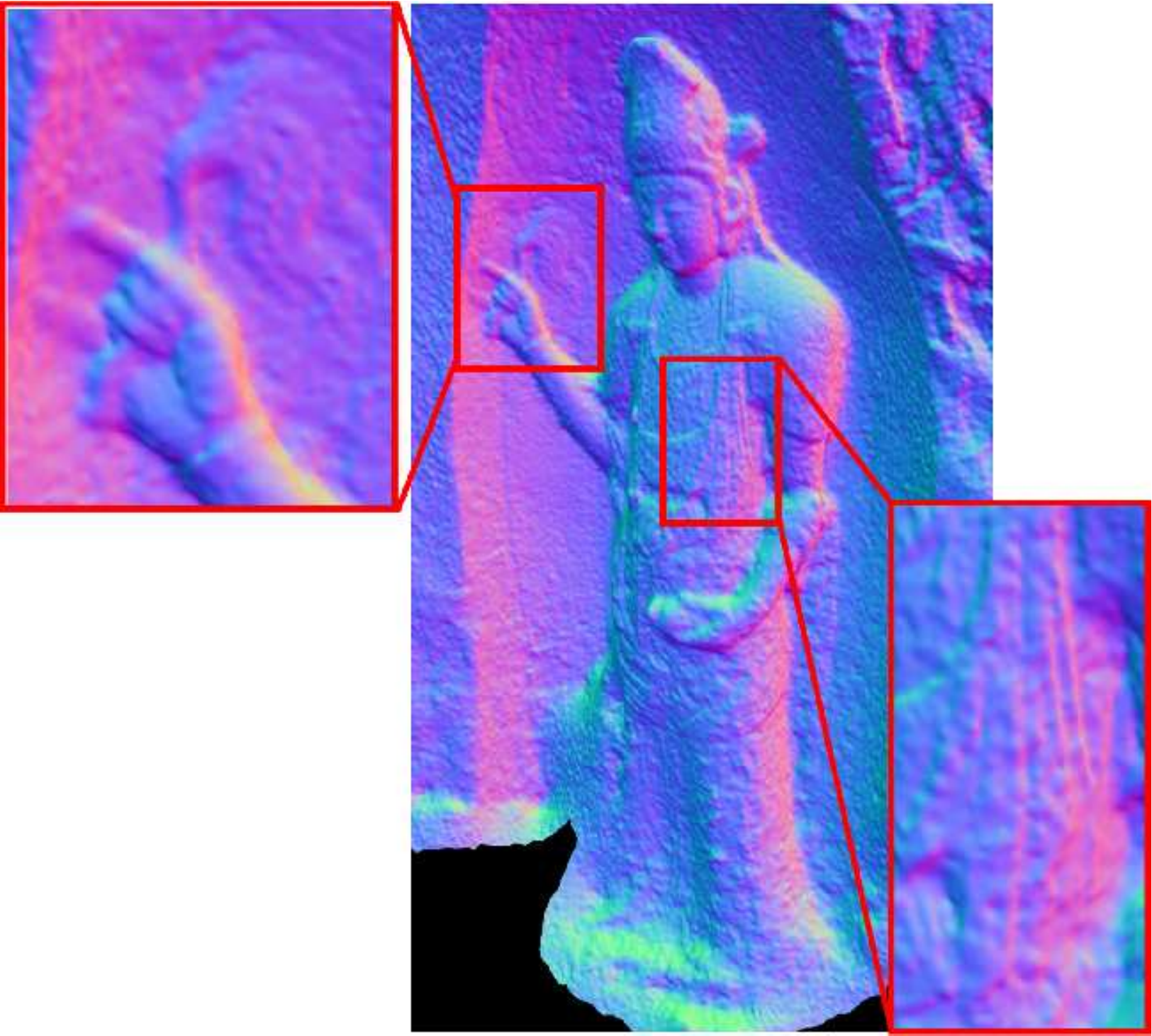}}
\vspace{-1mm}
\caption{Illustration of the proposed similarity measure. (a) One sample image from the {\it temple sparse ring} dataset. (b) The reconstruction result obtained by isotropic ZNCC similarity measure. (c) The result obtained by the proposed detail-preserving similarity measure. Comparing (c) with (b), one can see that the proposed method can better recovers the fine-scale details. (d)-(f) show another example from the Buddha dataset.}
\label{fig:detailenhancefig}
\vspace{-4mm}
\end{figure}

Motivated by the connection between guided image filtering and image registration, and to enhance the edge preservation of ZNCC-based similarity measure, we modify the derivative $\partial_{2} m$ in Eq. (\ref{eq:De5}) by adding a term to enforce the constraint $v_{1}(\textbf{p})=v_2(\textbf{p})$:
\begin{equation}
\partial \tilde{m}(\textbf{p})=\alpha I_{1}(\textbf{p})+\beta I_{2}(\textbf{p})+\gamma(\textbf{p})+\kappa G_{\sigma}\star\dfrac{(v_{2}-v_{1})}{v_{2}}(\textbf{p}),
\label{eq:De17}
\end{equation}
where $\kappa$ is a tradeoff parameter to adjust the influence of the variance constraint. In practice, we initialize $\kappa$ with a small value in the beginning of surface evolution, and gradually increase it until convergence. By Eq. \eqref{eq:De17}, the predicted image $\hat{I}_{i,j,S}$ is implicitly set as the guidance image. As shown in Fig. \ref{fig:detailenhancefig}, the proposed similarity measure can recover the fine-scale details and largely extend the edge preservation capability of the original isotropic measure.

\subsection{Content-aware Mesh Denoising via $L_{p}$-norm Minimization}
\label{section4.2}
Let matrix $X_0=(x_{0,i})_{i=1}^n\in\mathbb{R}^{3\times n}$ store the positions of the $n$ vertices of the reconstructed noisy surface mesh $S$, and matrix $X=(x_{i})_{i=1}^n\in\mathbb{R}^{3\times n}$ store the positions of the $n$ vertices of the noisy-free surface mesh $S^{k+0.5}$. The mesh denoising problem in Eq. \eqref{eq:update_denoising} can be reformulated as:
\begin{equation}
\min_{X} \frac{1}{2} \| X - X_0 \|^2 + \lambda R(X).
\label{eq:denoising_X}
\end{equation}
In this section, we first present the proposed content-aware model based on the MAP framework, and then propose an alternating minimization algorithm for content-aware mesh denoising.

\subsubsection{MAP-based Mesh Denoising with Hyper-Laplacian Prior}
\label{section4.2.1}
Denote by $q(X)$ the prior on the sharpness of noise-free mesh, and by $q(X_0|X)$ the likelihood of noisy mesh. The MAP framework estimates $X$ by maximizing a posterior probability $q(X|X_{0}) \propto q(X_0|X)q(X)$. By assuming that the noise is additive white Gaussian noise with standard deviation $\sigma$, the likelihood of noisy mesh can be modeled as:
 \begin{equation}
 q(X_0\vert X,\sigma^2) = \prod_{i}\dfrac{1}{\sqrt{2\pi\sigma^2}} \exp \left( - \dfrac{(x_{i}-x_{0,i})_{i}^2}{2\sigma^2}\right).
 \label{eq:ca2}
 \end{equation}


For surface mesh, the edge-based discrete Laplacian operator $D\in\mathbb{R}^{m\times n}$ proposed in He et al.~\cite{30} can be adopted for computing surface gradients (where $m$ is the number of edges in the mesh). In image restoration, it has been empirically verified that the natural image gradients generally follow a heavy-tailed distribution and can be well described by hyper-Laplacian~\cite{34}. Therefore, we suggest using hyper-Laplacian to model surface gradients:
  \begin{equation}
  q(X\vert \theta,p)=\prod_{i}\dfrac{p}{2}(\dfrac{\theta}{2})^\frac{1}{p}\dfrac{1}{\Gamma(\frac{1}{p})}exp(-\dfrac{\theta}{2}\vert (DX)_{i}\vert_{p}^p),
  \label{eq:ca1}
  \end{equation}
where $\Gamma$ is the Gamma function, and $p$ and $\theta$ are the shape parameters. $p\in [0,1]$ determines the peakiness and $\theta$ determines the width of a hyper-Laplacian distribution.
\begin{figure*}[tbp]
\centering
 \subfigure[]{
  \label{fig:subfig:redbox}
\includegraphics[height=21mm]{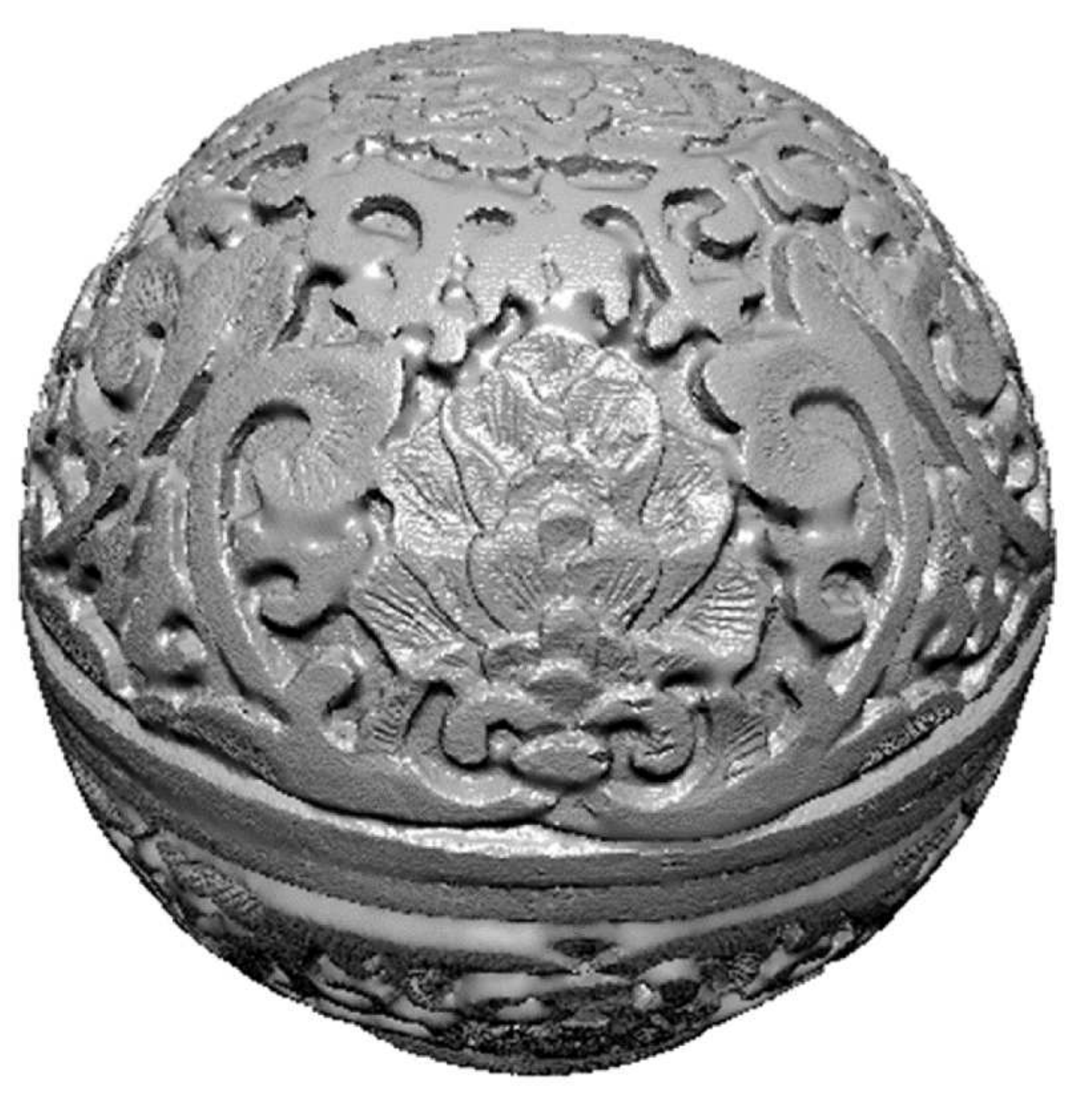}}
 \subfigure[]{
  \label{fig:subfig:hand}
\includegraphics[height=21mm]{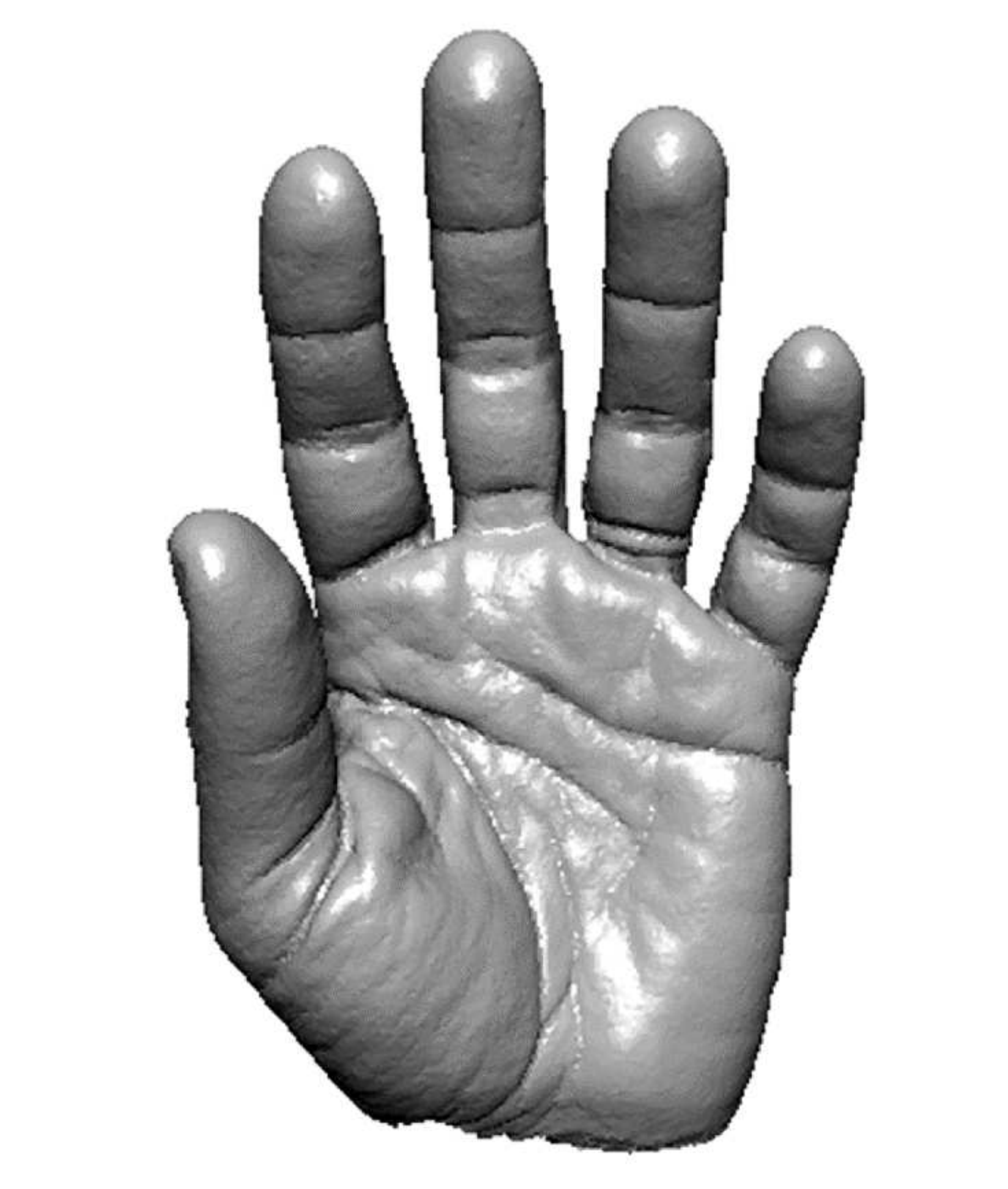}}
 \subfigure[]{
  \label{fig:subfig:gargo}
\includegraphics[height=21mm]{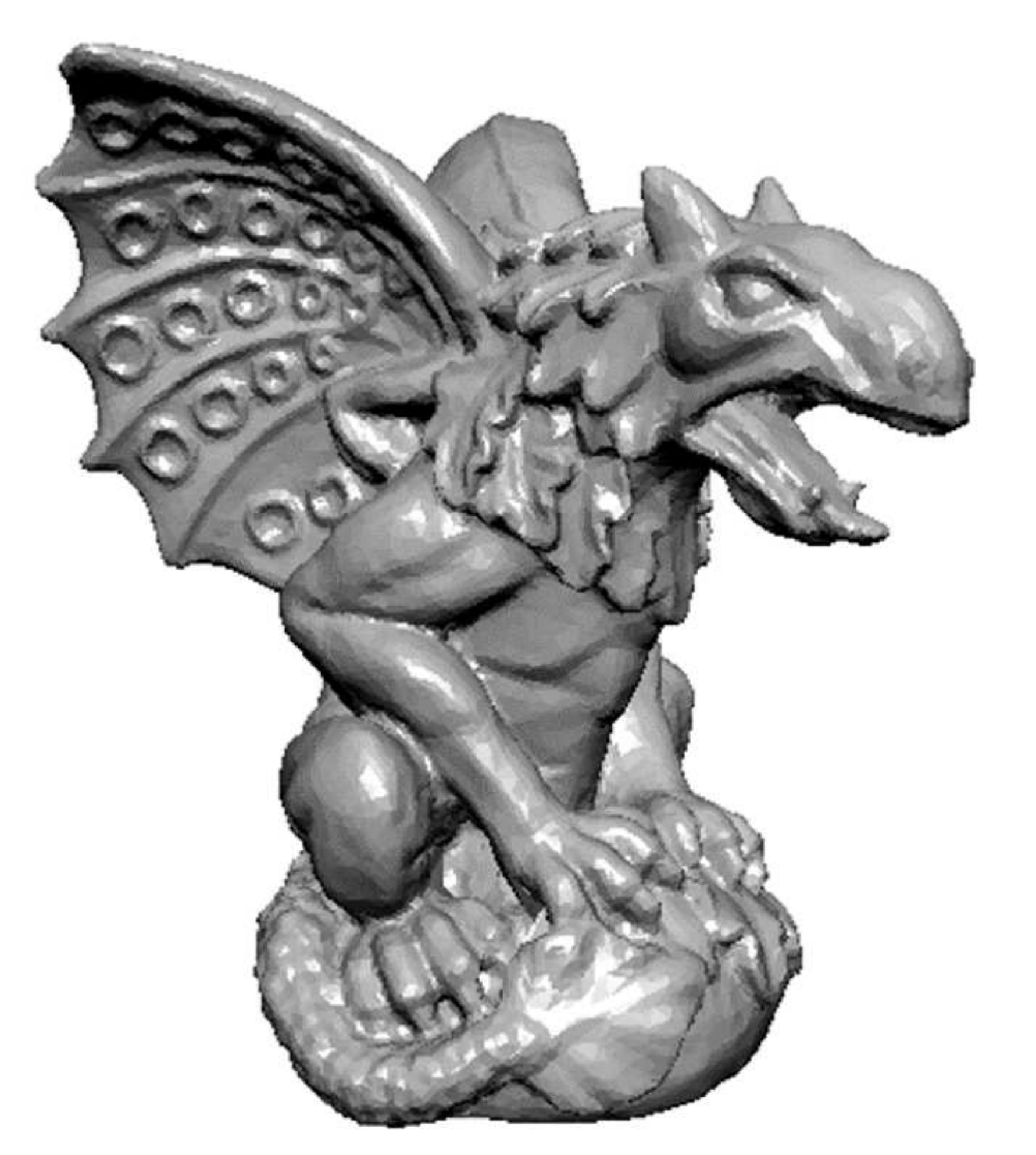}}
 \subfigure[]{
  \label{fig:subfig:redbox_distribution}
\includegraphics[height=25mm]{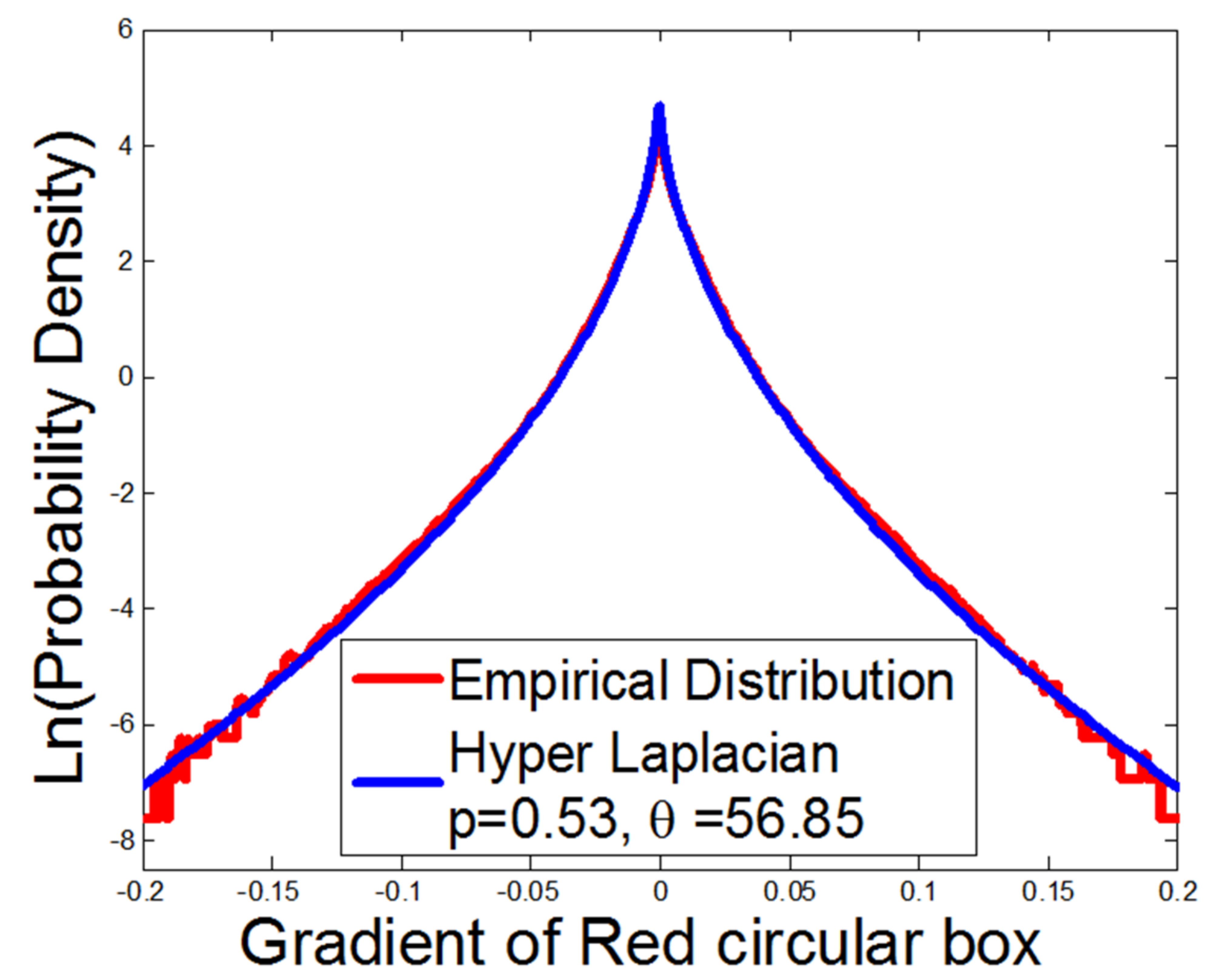}}
 \subfigure[]{
  \label{fig:subfig:hand_distribution}
\includegraphics[height=25mm]{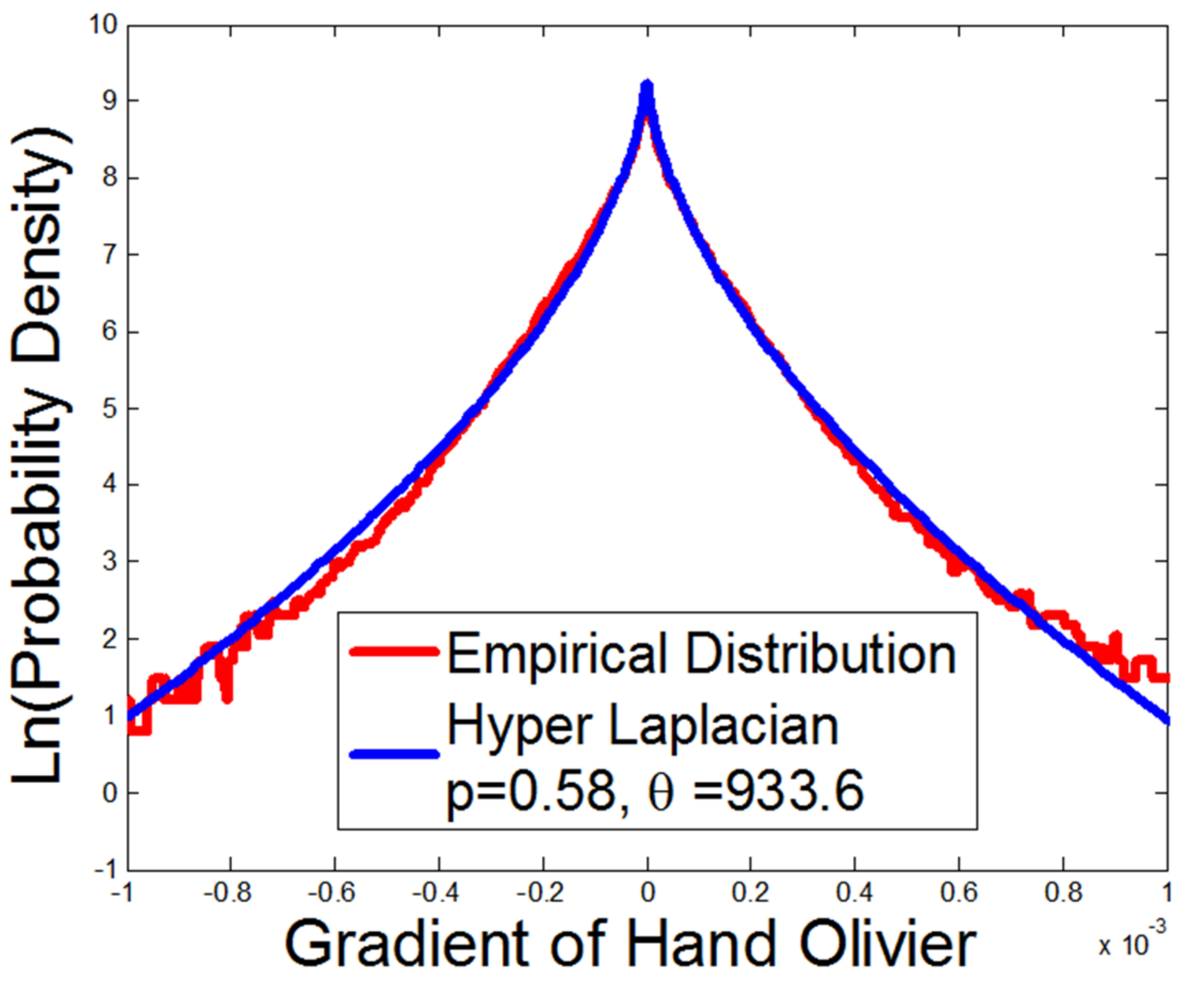}}
 \subfigure[]{
  \label{fig:subfig:gargo_distribution}
\includegraphics[height=25mm]{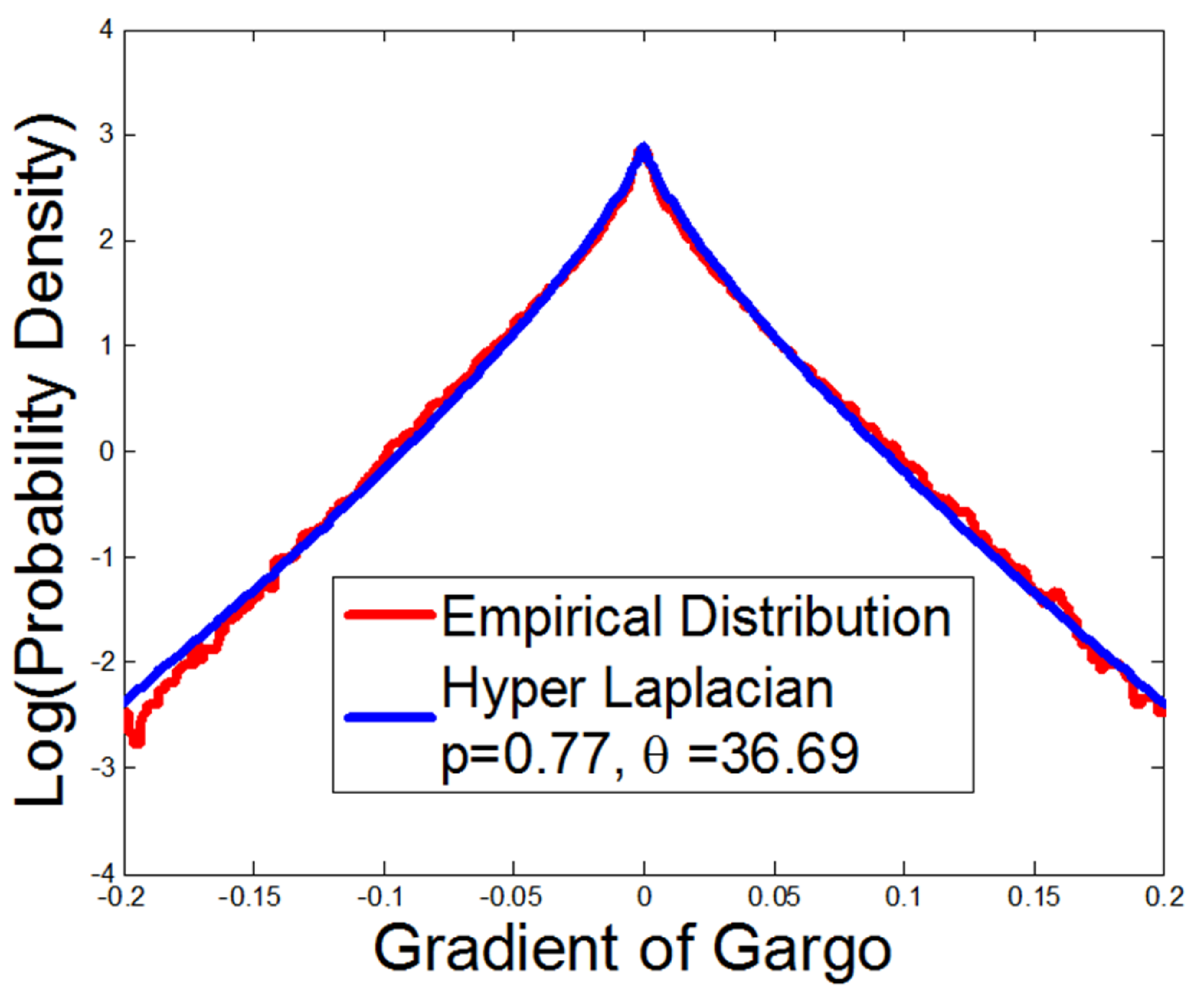}}
\vspace{-3mm}
\caption{Surface gradient distributions of three real 3D models: (a) the Red circular box model, (b) the Hand Olivier model, and (c) the Gargo model. (d)-(f) are their empirical distributions of sharpness (red) and the corresponding fitted hyper-Laplacian profiles (blue).}
\label{fig:priorfig}
\vspace{-5mm}
\end{figure*}

One concern is that whether surface gradients of real 3D models follow the hyper-Laplacian distribution. Fig.~\ref{fig:priorfig} shows the empirical distributions and the corresponding hyper-Laplacian fits of the surface gradients of three real models. One can see that hyper-Laplacian fits the empirical distribution very well, which validates that the empirical distribution can be well modeled by hyper-Laplacian.
 \begin{figure}[tbp]
 \centering
 \scalebox{1.35}[1.35]{\includegraphics[height=21mm]{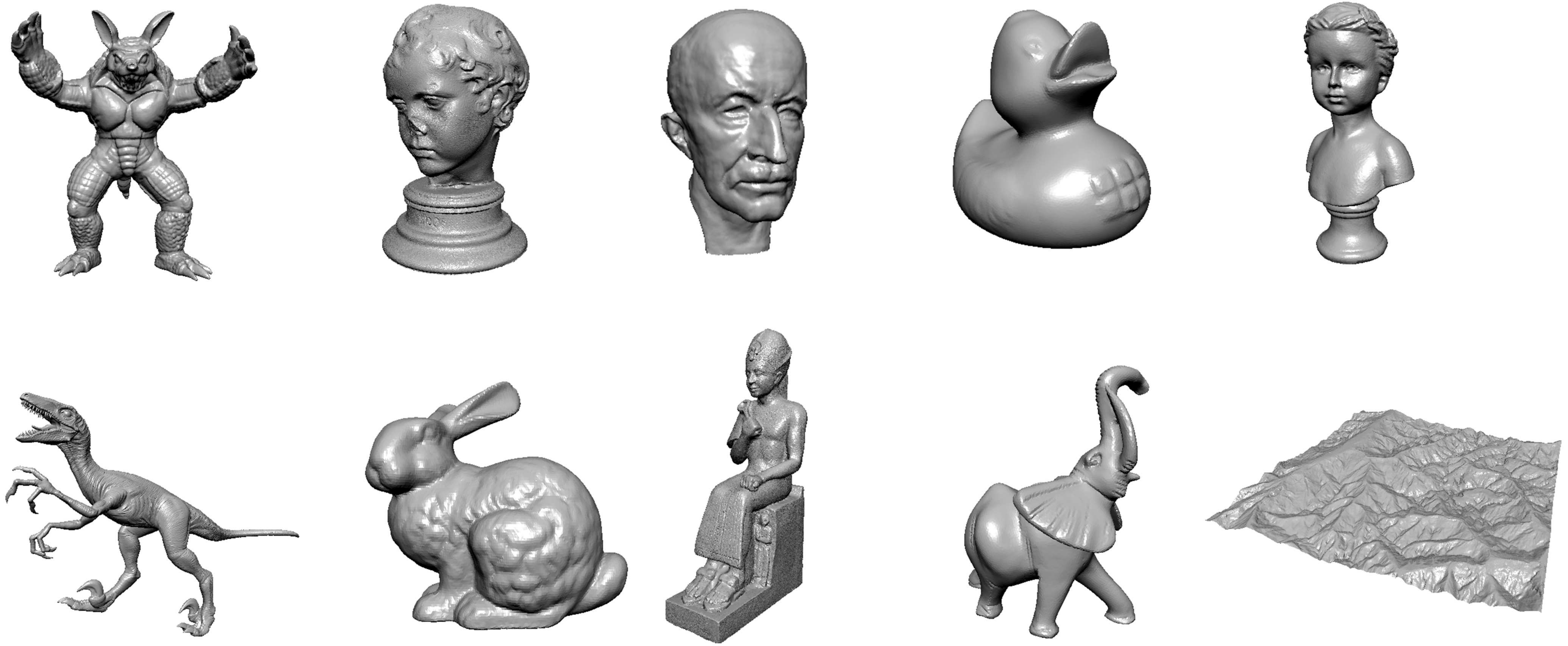} }%
 \caption{Shape parameters of different 3D models. From left to right and from top to down, Armadillo ($p=0.5$, $\theta=45.63$), Eros ($p=0.5$, $\theta=46.62$), Max Planck ($p=0.2$, $\theta=11.64$), Duck ($p=0.5$, $\theta=17.64$), Buste ($p=0.75$, $\theta=34.31$), Raptor ($p=0.3$, $\theta=195.3$), Bunny ($p=0.42$, $\theta=483.9$), Ramesses ($p=0.35$, $\theta=25.96$), Elephant ($p=0.5$, $\theta=357.3$), Cervino ($p=0.35$, $\theta=5.506$)}
 \label{fig:Lp_distribution}
 \vspace{-6mm}
 \end{figure}
It should be noted that, for different 3D models the shape parameters $p$ and $\theta$ will vary. To illustrate this, we compute the shape parameters on more than twenty public models, including Armadillo, Bunny in the Stanford repository~\cite{45} and models in the AIM@SHAPE repository~\cite{46}. We provide the shape parameters of ten models in Fig.~\ref{fig:Lp_distribution}. One can see that the $p$ values vary from $0.2$ to $0.75$ and the $\theta$ values vary from $5.506$ to $483.9$. Therefore, instead of fixing shape parameters, $p$ and $\theta$ values should be adaptively estimated for different 3D models. We propose the following content-aware mesh denoising model that jointly estimates the mesh $X$, noise level $\sigma^2$, and and the shape parameters $\theta$ and $p$ from the observation $X_0$:
 \begin{equation}
 \begin{aligned}
 ({\bar{X},\bar{\sigma^2},\bar{\theta},\bar{p}}) = \arg \min_{X,\sigma^2,\theta,p}\left\{- \log (q(X,\sigma^2,\theta,p\vert X_0))\right\}\\
 = \arg \min_{X,\sigma^2,\theta,p} \Bigg\{  \dfrac{1}{2}n\log(2\pi \sigma ^2)+\dfrac{\|X - X_0\|^2}{2\sigma^2}+\dfrac{\theta}{2}\vert (DX)_{i}\vert_{p}^p+ \\
  3m \left( \log(\Gamma (\dfrac{1}{p}))- \log(\dfrac{p}{2})-\dfrac{1}{p} \log(\dfrac{\theta}{2})\right) \Bigg\}.
 \end{aligned}
 \label{eq:ca2a}
 \end{equation}

\subsubsection{Alternating Minimization}
\label{section4.2.2}
We propose an alternating minimization algorithm to minimize Eq. (\ref{eq:ca2a}) for the joint estimation of noisy-free (denoised) mesh $X$, noise standard deviation $\sigma$, and hyper-Laplacian parameters $\theta $ and $p$ by iteratively solving the following two subproblems.

(1) Given $X$, the optimization problem w.r.t. $\sigma$, $\theta$, and $p$ can be reformulated as
\begin{equation}
  \bar{\sigma} = \arg \min_{\sigma }\Bigg\{ \frac{n}{2} \log (\sigma^2) + \frac{\|X - X_0\|^2}{2 \sigma^2} \Bigg\},
  \label{eq:ca8}
\end{equation}
\begin{equation}
\begin{aligned}
   &\bar{\theta},\bar{p}=\arg\min_{\theta,p}
   \Bigg\{\dfrac{\theta}{2}\vert (DX)_{i}\vert_{p}^p+  \\
     &3m \left( \log(\Gamma (\dfrac{1}{p}))- \log(\dfrac{p}{2})-\dfrac{1}{p} \log(\dfrac{\theta}{2})\right)\Bigg\}.
   \label{eq:ca7}
\end{aligned}
\end{equation} 
The $\sigma$-subproblem has the closed-form solution $\sigma^2 = \Vert X_0 -X\Vert^2/n$. The problem in Eq. \eqref{eq:ca7} can be solved by: ($a$) finding the optimal $\theta$ for given $p$, which results in a closed-form solution $\theta=3m/p\vert (DX)_{i}\vert_{p}^p$, and ($b$) using a simple 1D exhaustive searching strategy to obtain the estimation of $p$ for given $\theta$.

(2) Given $\sigma$, $\theta$ and $p$, we define $\lambda=\theta\sigma^2/2$, and then we have
 \begin{equation}
 \bar{X} =\arg\min_{X}\dfrac{1}{2}\Vert X_0 -X\Vert_{2}^2+\lambda \Vert DX\Vert_{p}^p
 \label{eq:ca3}
 \end{equation}
 where $\lambda$ is the regularziation parameter. Using the variable splitting approach, Eq. \eqref{eq:ca3} can be reformulated as:
  \begin{equation}
  \bar{X}=\arg\min_{X}\dfrac{1}{2}\Vert X_0 -X\Vert_{2}^2+\lambda \Vert \psi\Vert_{p}^p+\beta\vert DX-\psi\vert^2,
  \label{eq:ca4}
  \end{equation}
which can also be optimized with an alternating optimization method. Fix $\psi$, $X$ can be optimized by solving the following quadratic problem:
\begin{equation}
\bar{X}=\arg\min_{X}\dfrac{1}{2}\Vert X_0 -X\Vert_{2}^2+\beta\vert DX-\psi\vert^2.
  \label{eq:ca5}
\end{equation}
Fix $X$, $\psi$ can be optimized by solving the following subproblem:
  \begin{equation}
  \bar{\psi}=\arg\min_{\psi}\lambda \Vert \psi\Vert_{p}^p+\beta\vert DX-\psi\vert^2.
  \label{eq:ca6}
  \end{equation}
 The above subproblem can be efficiently solved by using the generalized shrinkage/thresholding (GST) method~\cite{35}.
 Solution to each $(\psi)_{i}$ can be written as:
\begin{equation}
(\psi)_{i}=T_{p}^{GST}((DX)_{i};\frac{\lambda}{2\beta}),
\end{equation}
where $T_{p}^{GST}$ is the generalized shrinkage/thresholding operator \cite{35}. When the penalty factor $\beta\to\infty$, the solution to Eq. \eqref{eq:ca4} converges to that of Eq. \eqref{eq:ca3}. In practice, we adopt the continuation technique by initializing $\beta$ with a small value and gradually increasing it until convergence.

\section{Experiments}
\label{section5}
 We implemented the proposed DCV method using C++ with OpenGL, CGAL and TAUCS library. The predicted images are estimated using projective texture mapping, and OpenGL is adopted to generate the horizon and terminator of triangular surface. CGAL is used to manipulate the triangular mesh and TAUCS is used to manipulate the sparse matrix. We quantitatively and qualitatively  evaluate the performance of DCV on multiple datasets, including the Middlebury benchmark and several public datasets with indoor and outdoor scenes. These public datasets have camera calibration parameters available. We also provide three real datasets, i.e., {\it Buddha}, {\it Totoro} and {\it bell}, taken from mobile phone or digital camera. For these three datasets, cameras are calibrated using the Bundler software \cite{41}. In all experiments, the window size for calculating similarity measure is set as 7$\times 7$ pixels.
\subsection{Initialization and Implementation Details}
In our experiments, we consider two initialization methods: visual hull and PMVS+PSR (Possion Surface Reconstruction). The visual hull is the intersection of the visual cones associated with all image silhouettes, and can provide a good initialization for most indoor scenes where the interested object is easy to be segmented from background. The PMVS is an open source software designed by Furukawa and Ponce~\cite{31a}. A set of dense patches are generated from PMVS with its default parameters and then a triangular surface mesh is estimated by using PSR\cite{40} with octree depth fixed to 8. PMVS+PSR is mainly used to initialize the scene where the background is not easy to be segmented from foreground, including some outdoor and indoor scenes. For the {\it temple} dataset, because some small protruding structures tend to be over-smoothed when large concave region is recovered in the back of temple, we also use the PMVS+PSR to generate an initial mesh. The statistics of all the datasets used in our experiments are listed in Table~\ref{table2_DeCoS}, including the number of images, image resolution, initial points and running times (CPU i7, 2.4Ghz).

Two issues, non-convexity and topology adaptivity, are considered in our implementation. Since the $L_p$-sparsity ($0 \leq p\leq 1$) is used, the objective functional of DCV is non-convex, making the algorithm sensitive to local minimum. To alleviate this, we adopt a multi-resolution scheme. We first minimize the energy on low resolution mesh and downsample images accordingly, and then optimize it on  high-resolution mesh and full-size images. The Gaussian pyramid is used to downsample the image. The  Qslim algorithm~\cite{42} is used to simplify the mesh, and $\sqrt{3}$-subdivision scheme~\cite{43} is used to subdivide the mesh to a higher resolution. Another issue is the topology adaptivity of mesh-based methods, which needs an initial surface of the model with an approximately consistent topology. We use two initialization methods, visual hull and PMVS+PSR, in the experiments. Other initialization methods can also be deployed, e.g., those methods based on features, fusion of depth maps, and volumetric optimization.

\input{table2_DeCoS_double}

\subsection{Middlebury Datasets}
We first evaluate the effectiveness of DCV on the Middlebury benchmark \cite{1} by using two performance indicators: accuracy and completeness. The accuracy is measured by the distance $d$ such that the distance between 90\% of the reconstructed surface and the ground truth surface is less than $d$. Completeness is measured by the percentage $f$ such that the distance between percentage $f$ of the ground truth surface and the reconstructed surface is less than $1.25$ mm.

\begin{figure}[bthp] %
 \centering
   \subfigure { 
 \includegraphics[height=50mm]{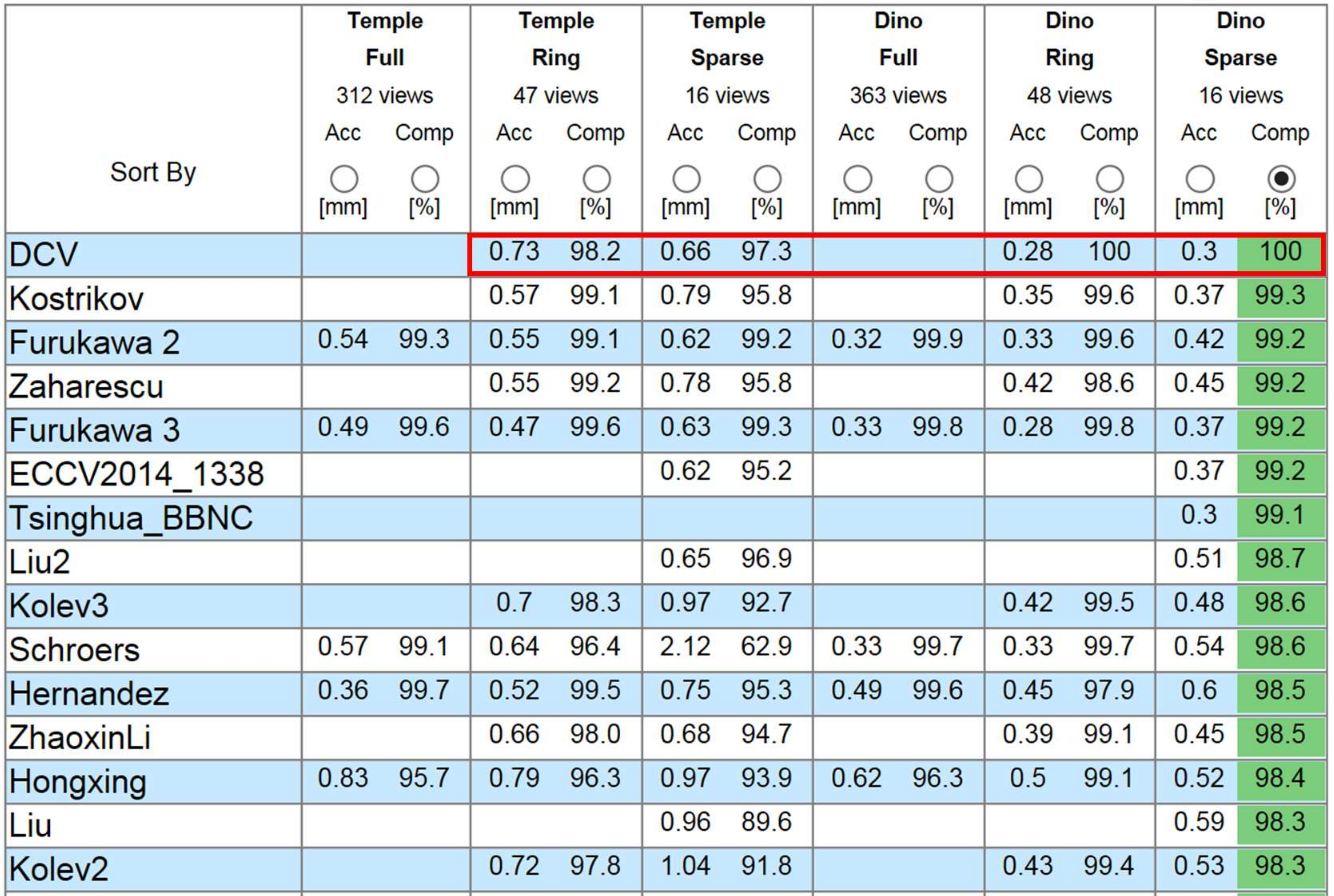}}
 \caption{Evaluation results of DCV on the Middlebury Benchmark.}
 \label{fig:rank}
  \vspace{-1mm}
 \end{figure}

\begin{figure*}[tbp]
      \centering
   \subfigure[]{
   \label{fig:subfig:templering_view1_vu}
   \includegraphics[height=26mm]{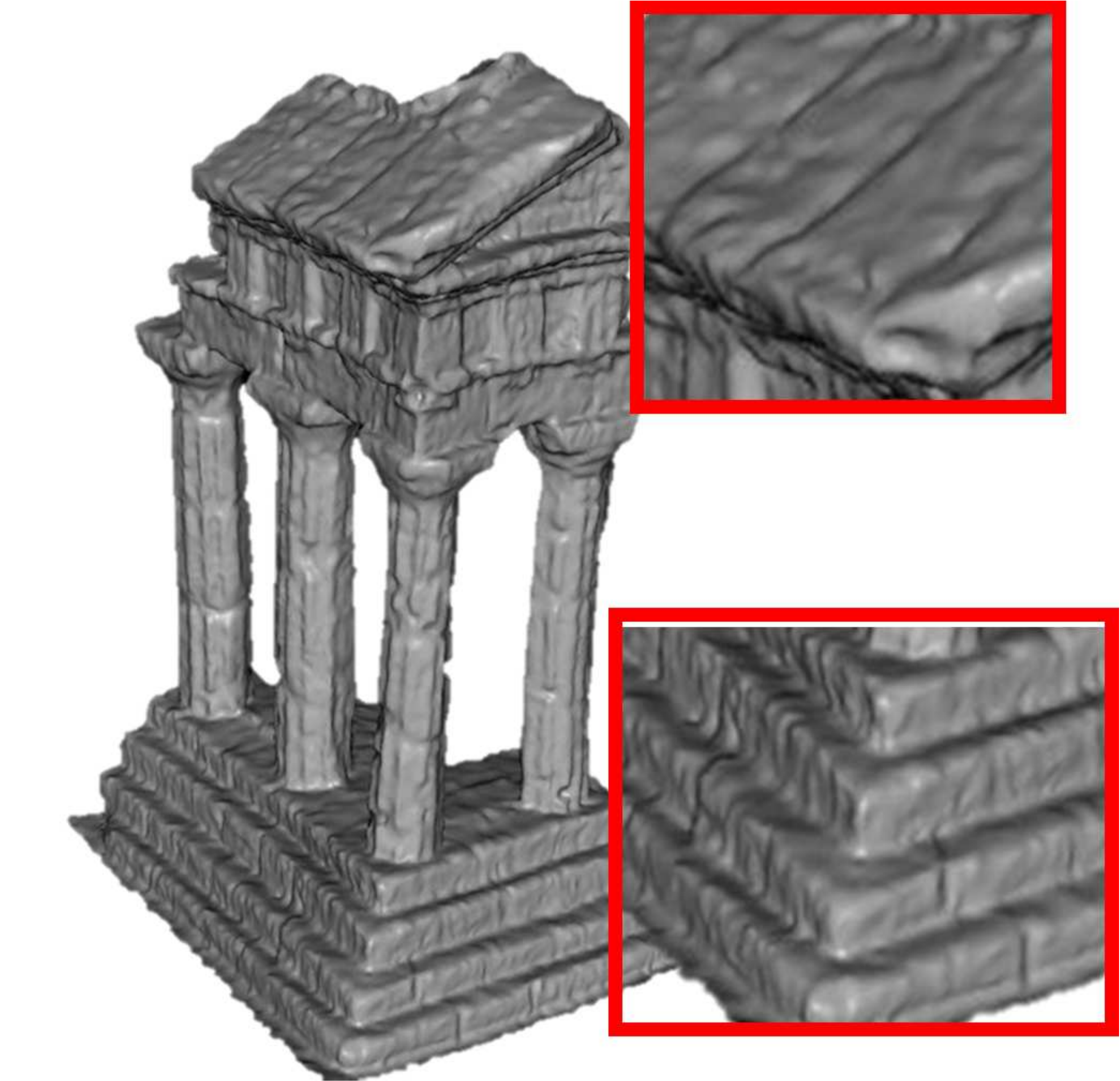}}
   \subfigure[]{
   \label{fig:subfig:templering_view1_campbell}
   \includegraphics[height=26mm]{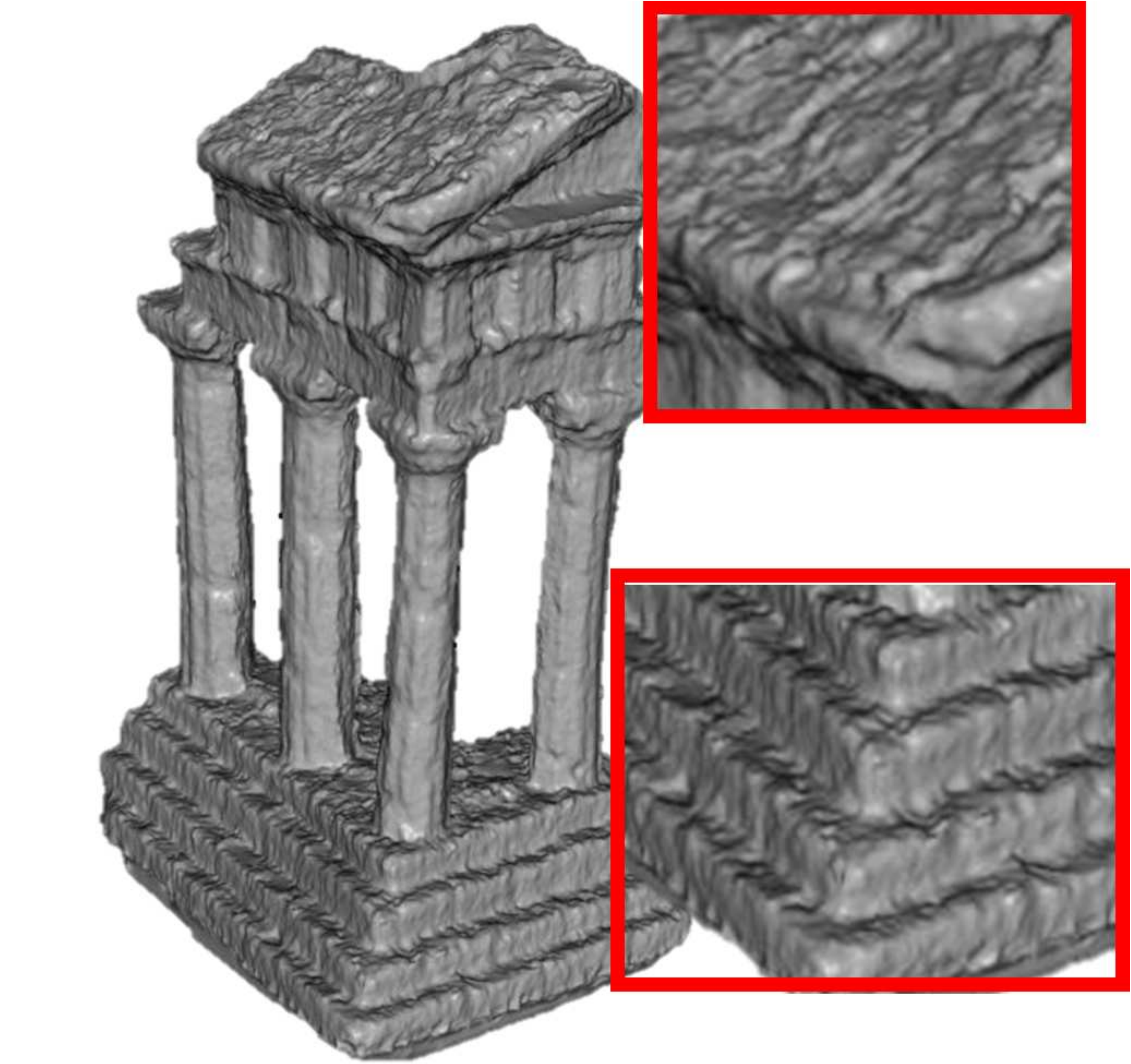}}
   \subfigure[]{
   \label{fig:subfig:templering_view1_furukawa3}
   \includegraphics[height=26mm]{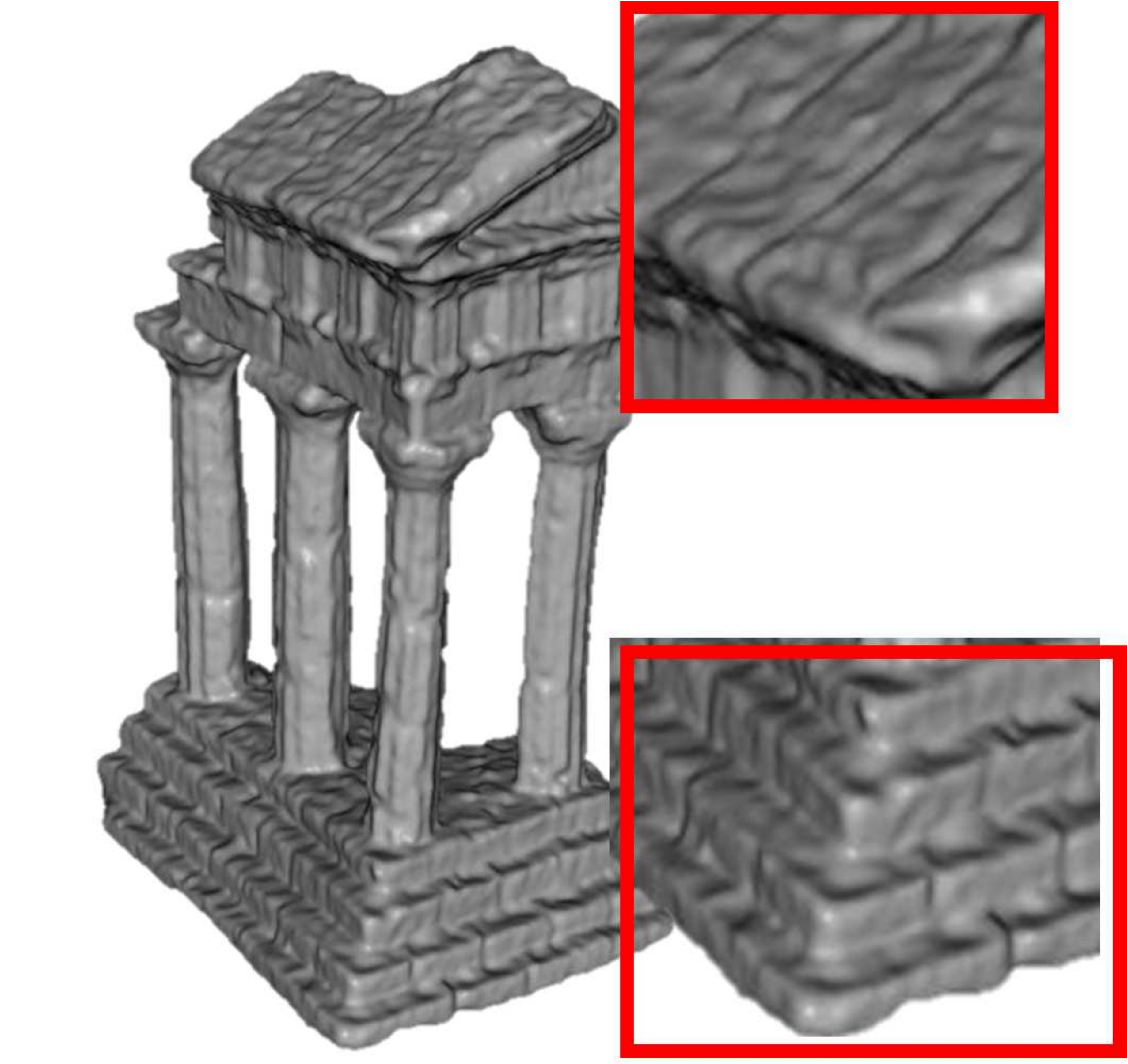}}
   \subfigure[]{
   \label{fig:subfig:templering_view1_hernandez}
   \includegraphics[height=26mm]{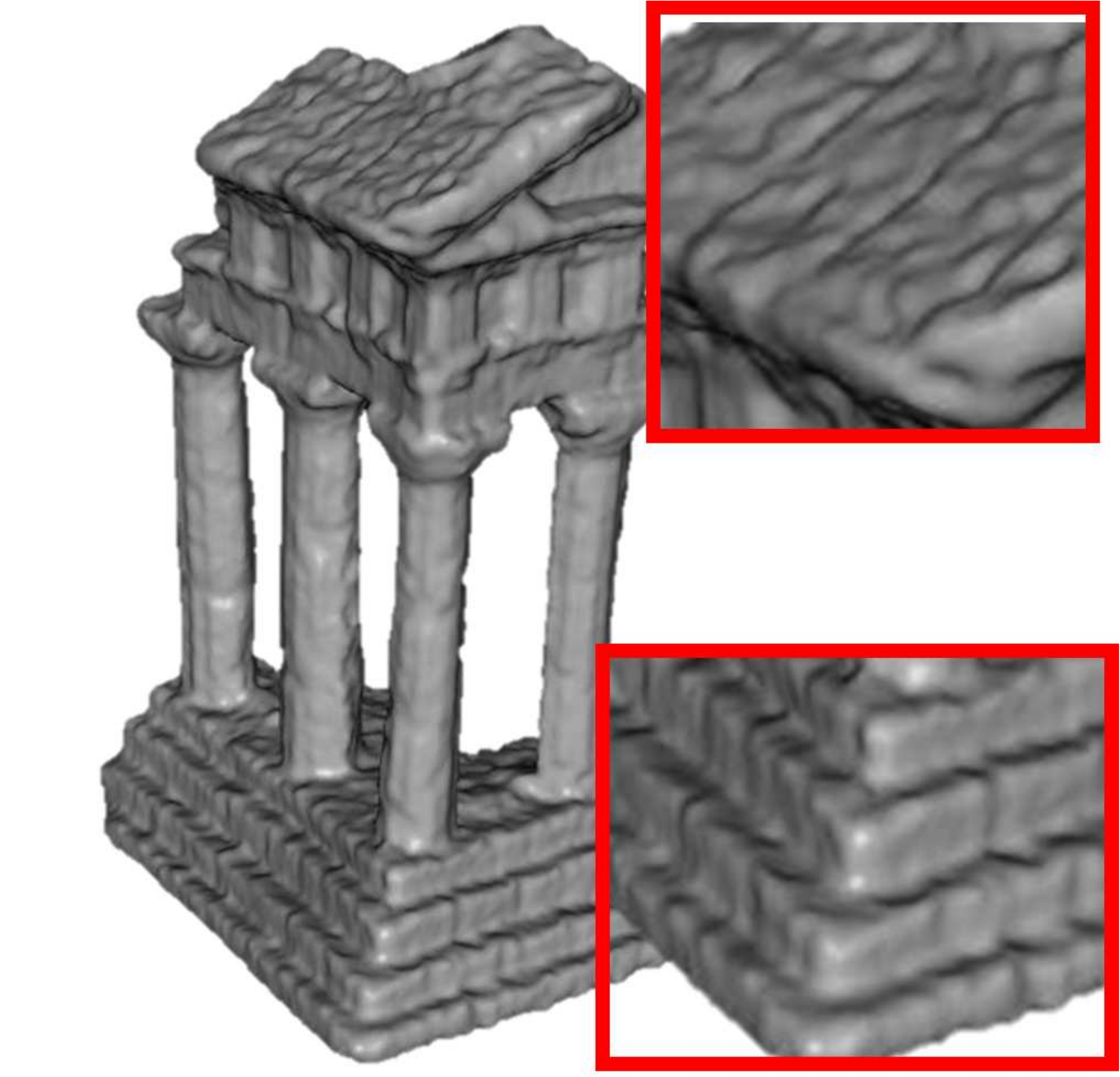}}
   \subfigure[]{
   \label{fig:subfig:templering_view1_decos}
   \includegraphics[height=26mm]{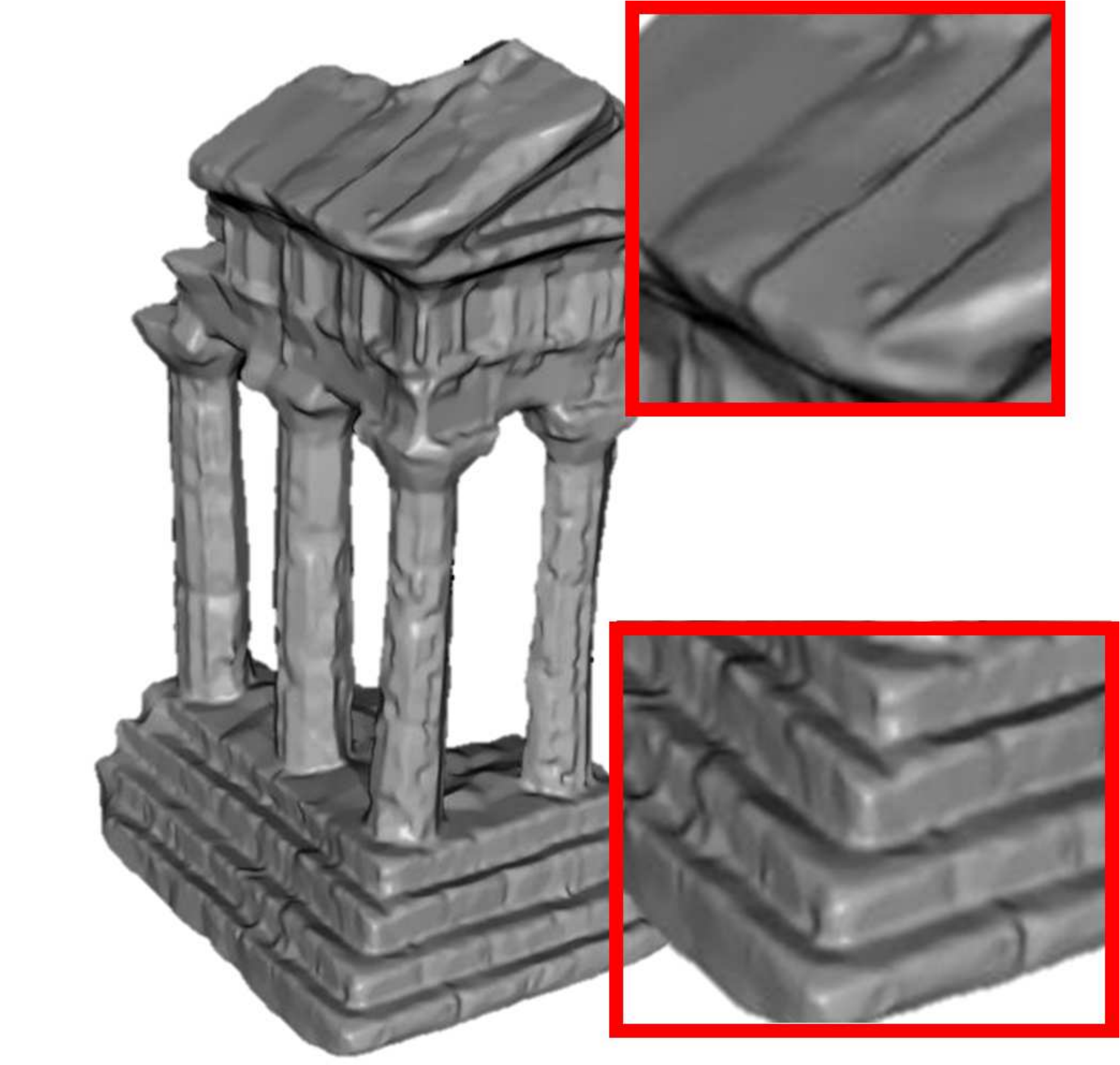}}
   \subfigure[]{
   \label{fig:subfig:templering_view1_ground}
   \includegraphics[height=26mm]{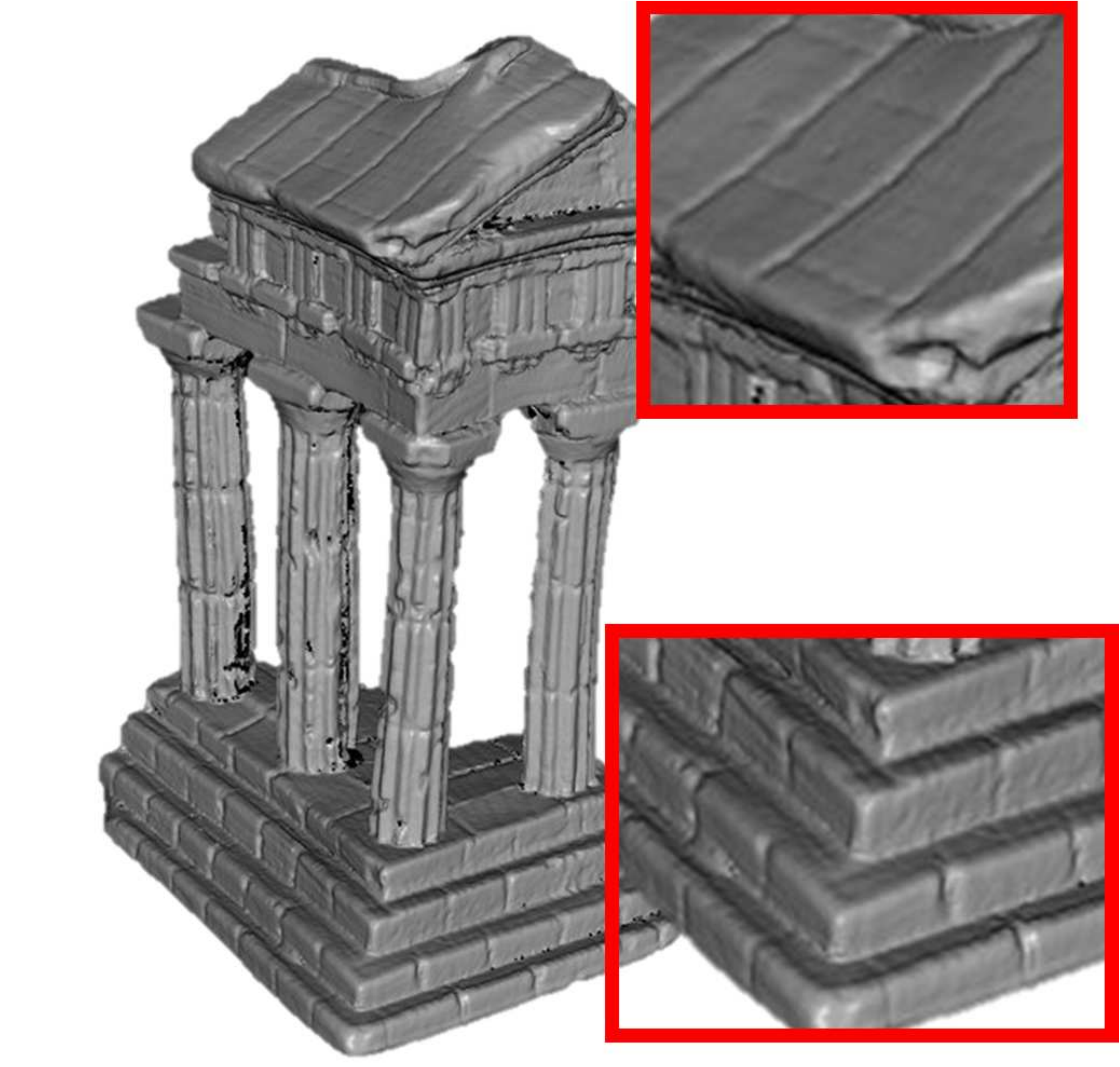}}
      \vspace{-3mm}
      \caption{Comparison of reconstruction results of DCV on the Middlebury \textit{temple ring} dataset. The names of the comparison methods follow the entries in the Middlebury evaluation website. (a) Vu \cite{20}, Acc. 0.45, Comp. 99.8\%. (b) Campbell\cite{49}, Acc. 0.48, Comp. 99.4\%. (c) Furukawa3\cite{31a}, Acc. 0.47, Comp. 99.6\%. (d) Hernandez\cite{12}, Acc. 0.52, Comp. 99.5\%. (e) th proposed DCV method, Acc. 0.73, Comp. 98.2\%. (f) groundtruth.}
      \label{fig:templeringcomp}
      \vspace{-3mm}
      \end{figure*}

\begin{figure*}[tbp]
      \centering
   \subfigure[]{
   \label{fig:subfig:dinosparse}
   \includegraphics[height=25mm]{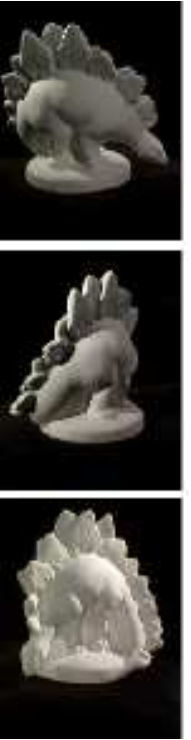}}
   \subfigure[]{
   \label{fig:subfig:dinosparsevh_view1}
   \includegraphics[height=25mm]{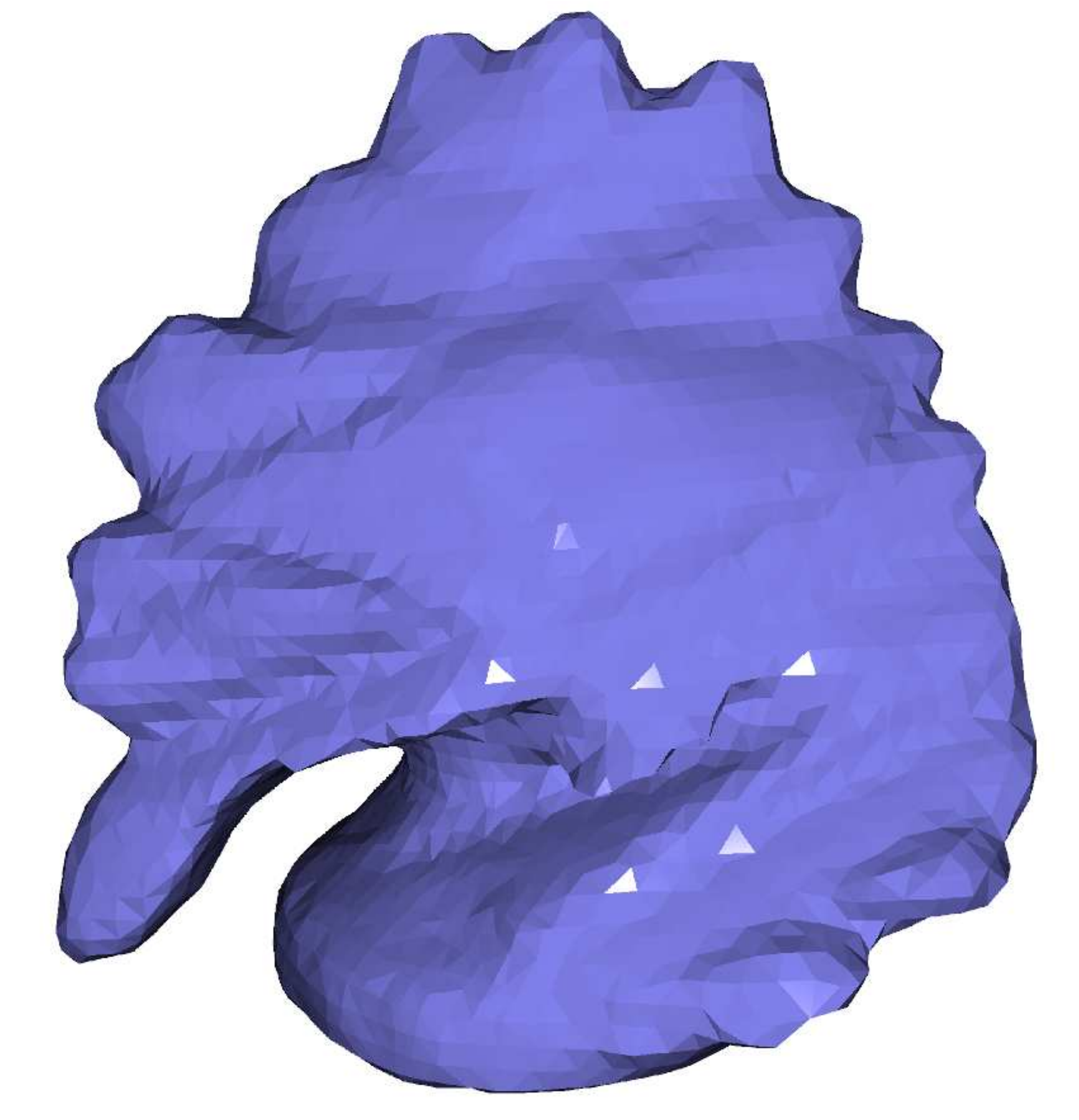}}
   \subfigure[]{
   \label{fig:subfig:dinosparsedecos_view1}
   \includegraphics[height=25mm]{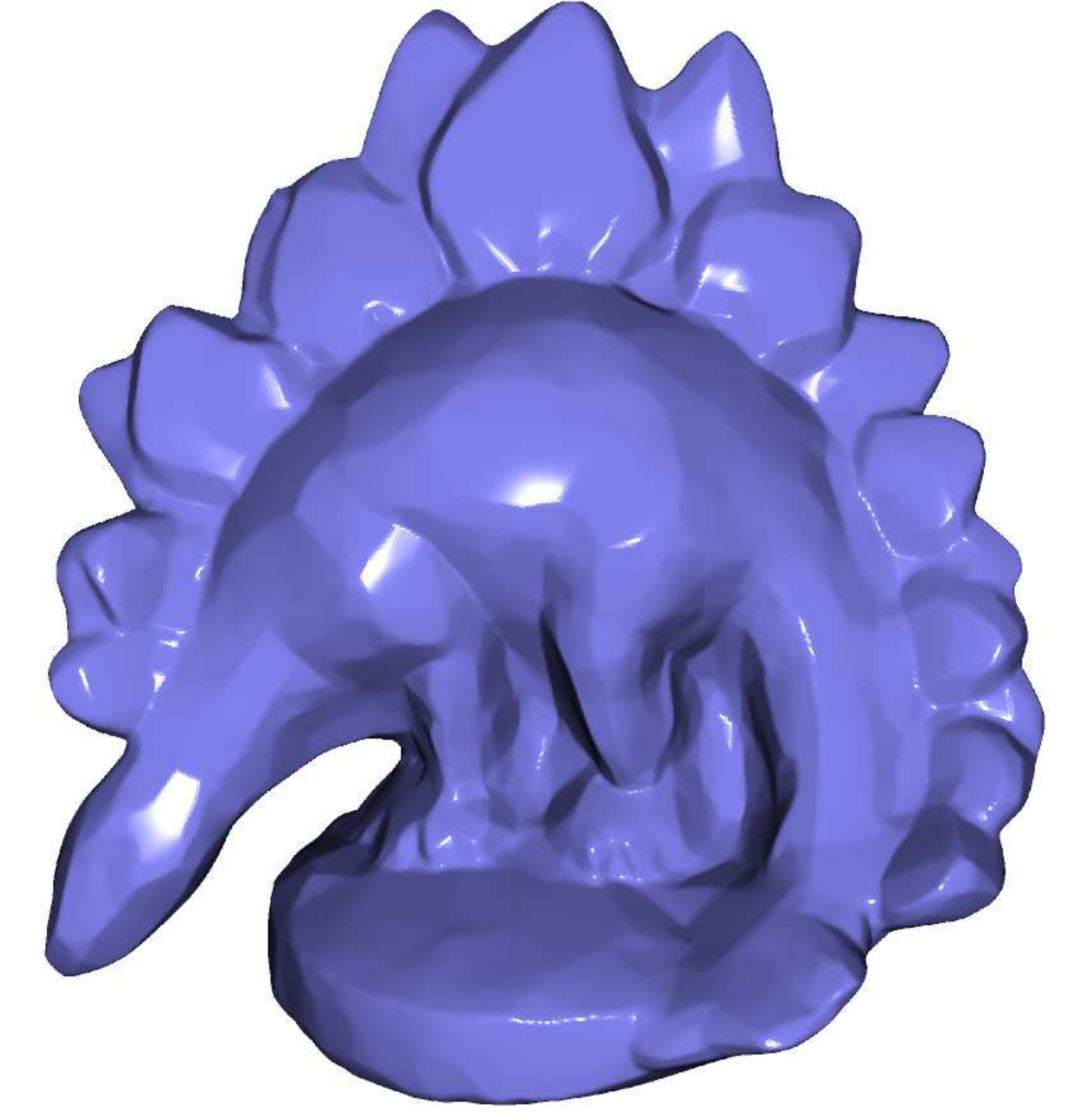}}
   \subfigure[]{
   \label{fig:subfig:dinosparsevh_view2}
   \includegraphics[height=25mm]{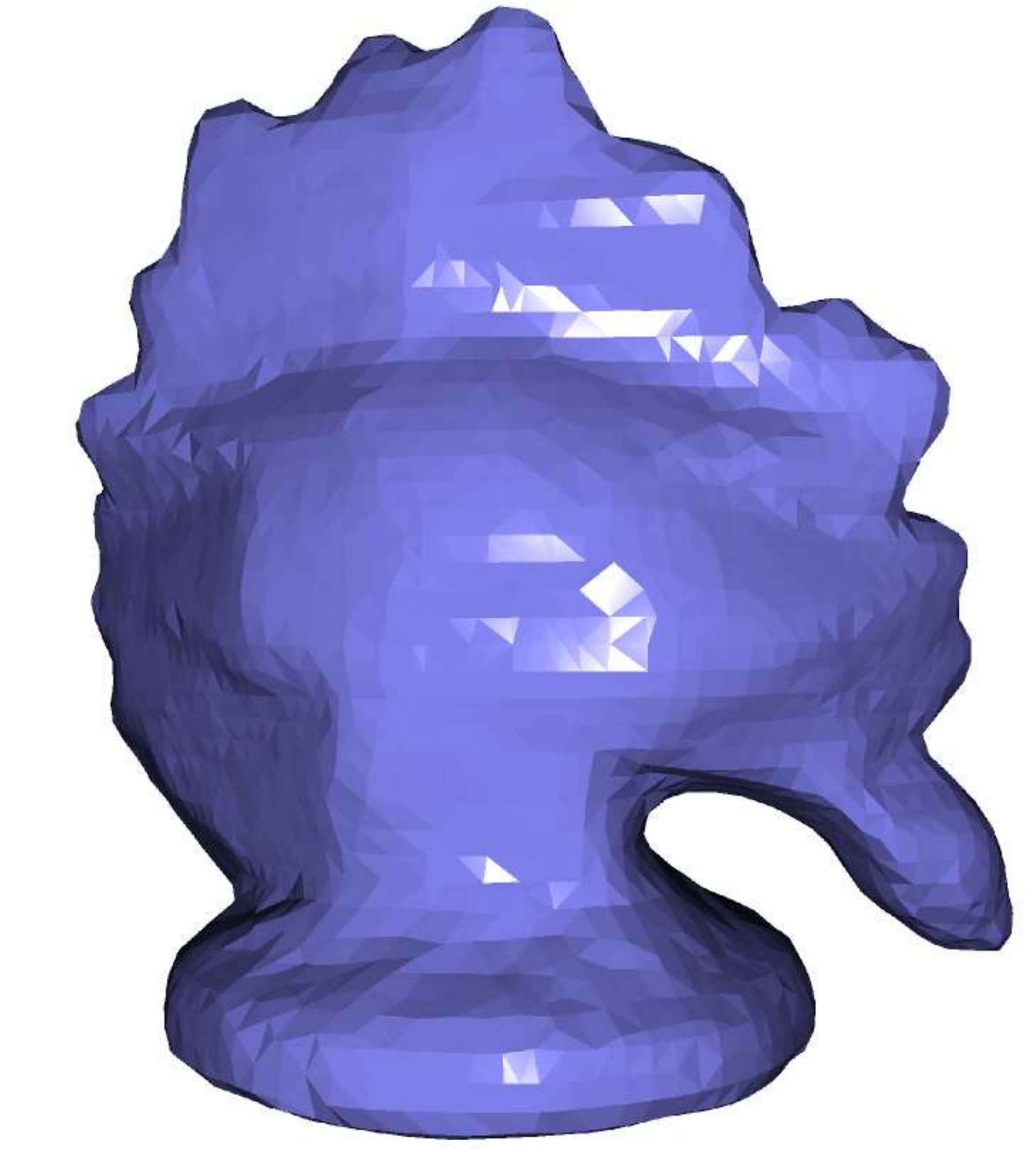}}
   \subfigure[]{
   \label{fig:subfig:dinosparsedecos_view2}
   \includegraphics[height=25mm]{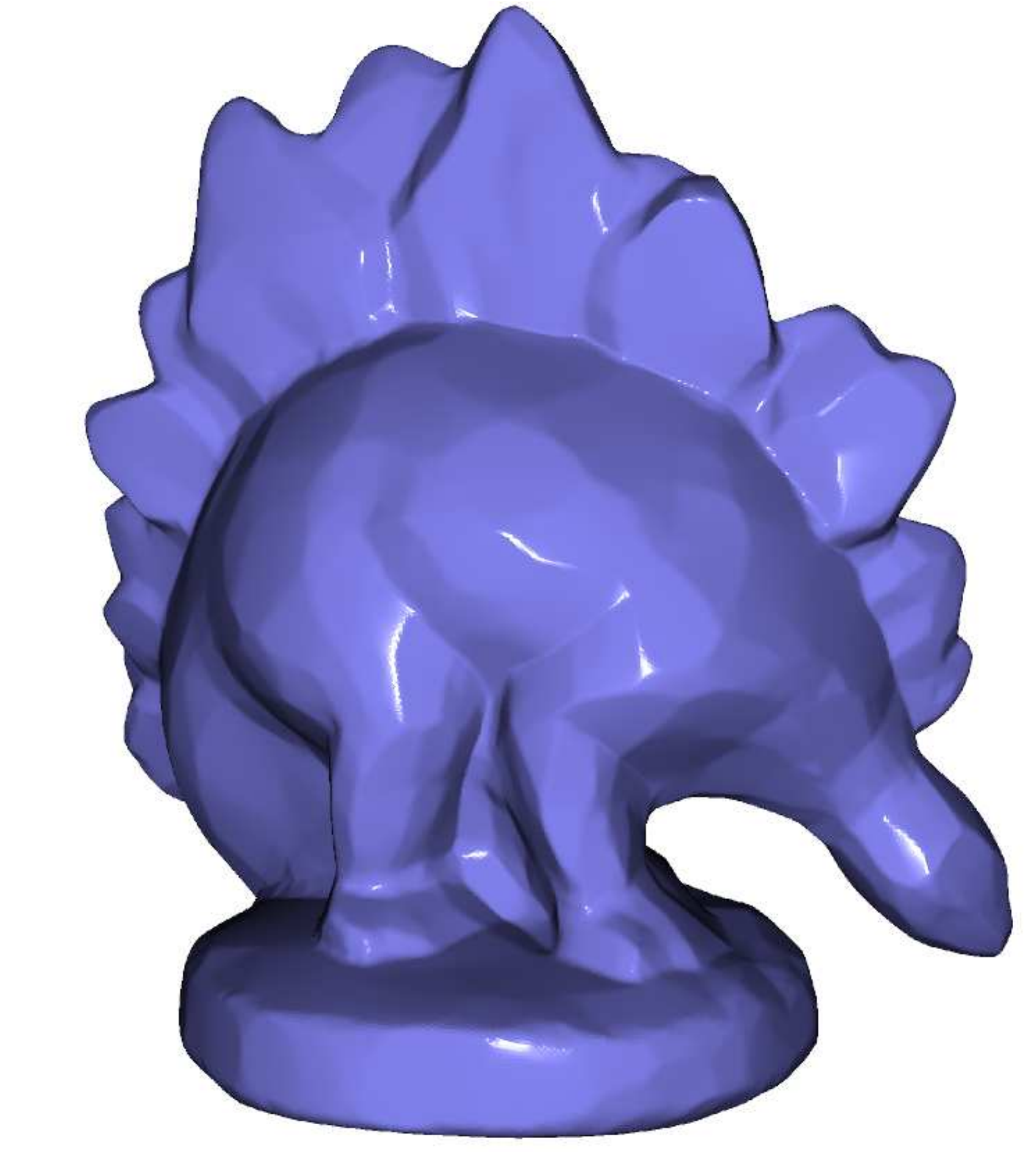}}

   \subfigure[]{
   \label{fig:subfig:templesparse}
   \includegraphics[height=25mm]{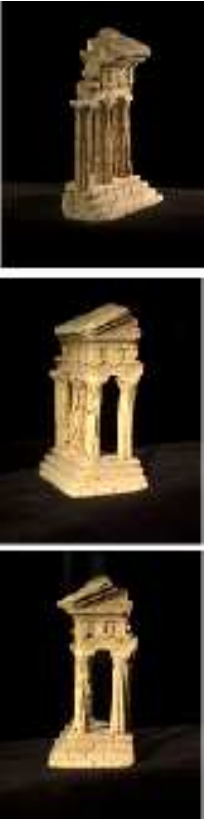}}
   \subfigure[]{
   \label{fig:subfig:templesparsevh_view1}
   \includegraphics[height=25mm]{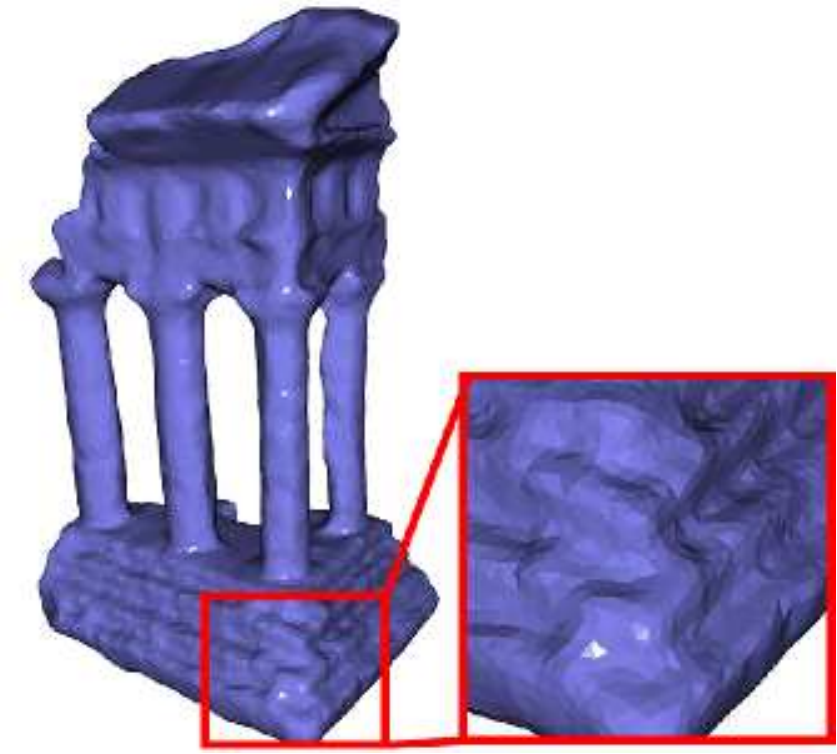}}
   \subfigure[]{
   \label{fig:subfig:templesparsedecos_view1}
   \includegraphics[height=25mm]{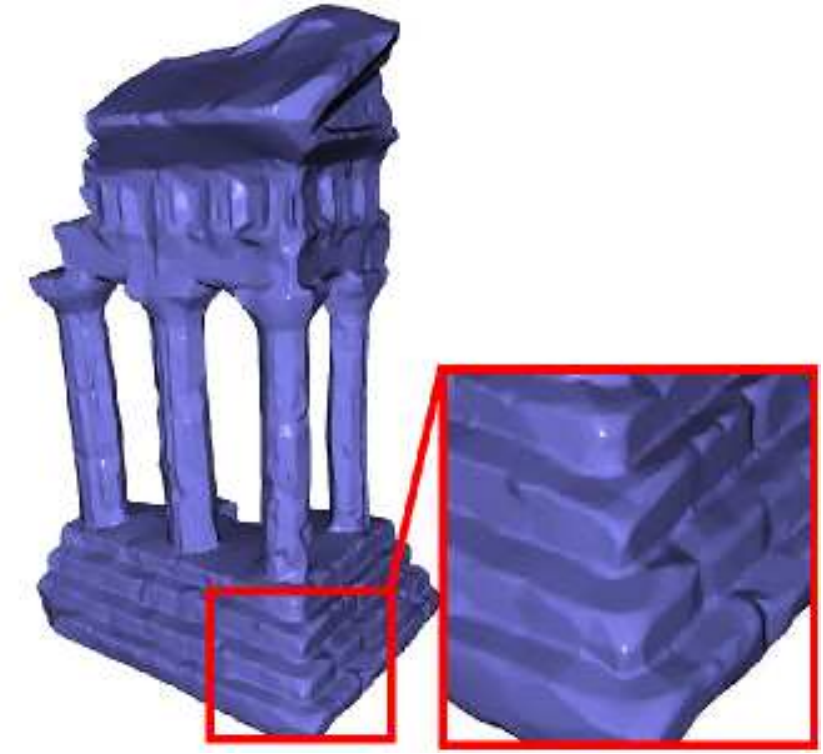}}
   \subfigure[]{
   \label{fig:subfig:templesparsevh_view2}
   \includegraphics[height=25mm]{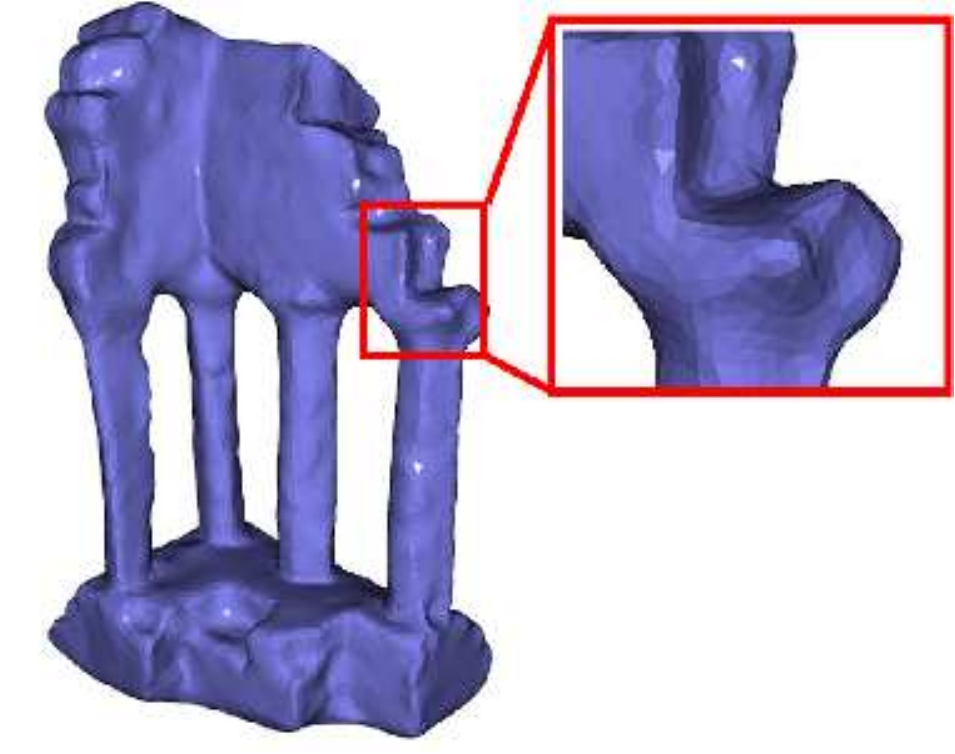}}
   \subfigure[]{
   \label{fig:subfig:templesparsedecos_view2}
   \includegraphics[height=25mm]{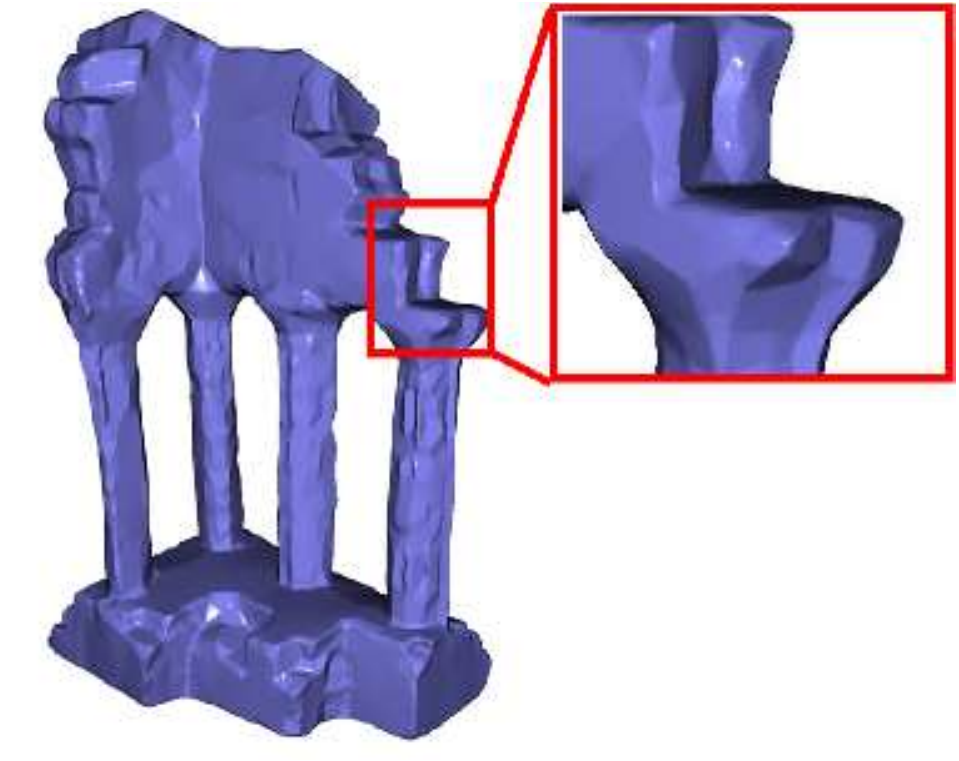}}

     \vspace{-3mm}
      \caption{Reconstruction results of  DCV  on {\it dino sparse ring }(first row) and {\it temple sparse ring }(second row). (a) Some samples of {\it dino sparse ring}; (b) and (d) are the visual hulls; (c) and (e) are the reconstruction results of DCV. (f) Some samples of {\it temple sparse ring}; (g) and (i) are the reconstruction results of PMVS+PSR, which are adopted as the initial surfaces to DCV; (h) and (j) are the reconstruction results of DCV.}
      \label{fig:introductionfig}
      \vspace{-3mm}
      \end{figure*}

We test DCV on the {\it dino ring (48-views)}, {\it dino sparse ring (16-views)}, {\it temple ring (47-views)} and {\it temple sparse ring (16-views)}, respectively. The smaller the number of views is, the more difficult and challenging the reconstruction will be.
The accuracy and completeness of DCV on these four datasets are shown in Fig. \ref{fig:rank}. These  results are also publicly available on the Middlebury evaluation page~\cite{37} and can be compared with state-of-the-arts. It is worthy to mention that (at the time that this paper is submitted) our DCV method achieves the best result on the {\it dino ring} and {\it dino sparse ring} in terms of both model completeness and  accuracy. Though our results on the {\it temple ring} and {\it temple sparse ring} datasets are not top ranked, it can be easily seen that the visual quality of our results is better than most of the other top ranked methods. In Fig. \ref{fig:templeringcomp}, we compare the reconstruction results by DCV and several top ranked methods on the {\it temple ring} datasets. The reconstruction results of {\it dino sparse ring} and {\it temple sparse ring} by DCV are shown in Fig. \ref{fig:introductionfig}, including the comparison with their coarse initializations, which indicates that the proposed method is not sensitive to initialization.

\begin{figure*}[tbp]
 \centering
\subfigure{
\includegraphics[height=25mm]{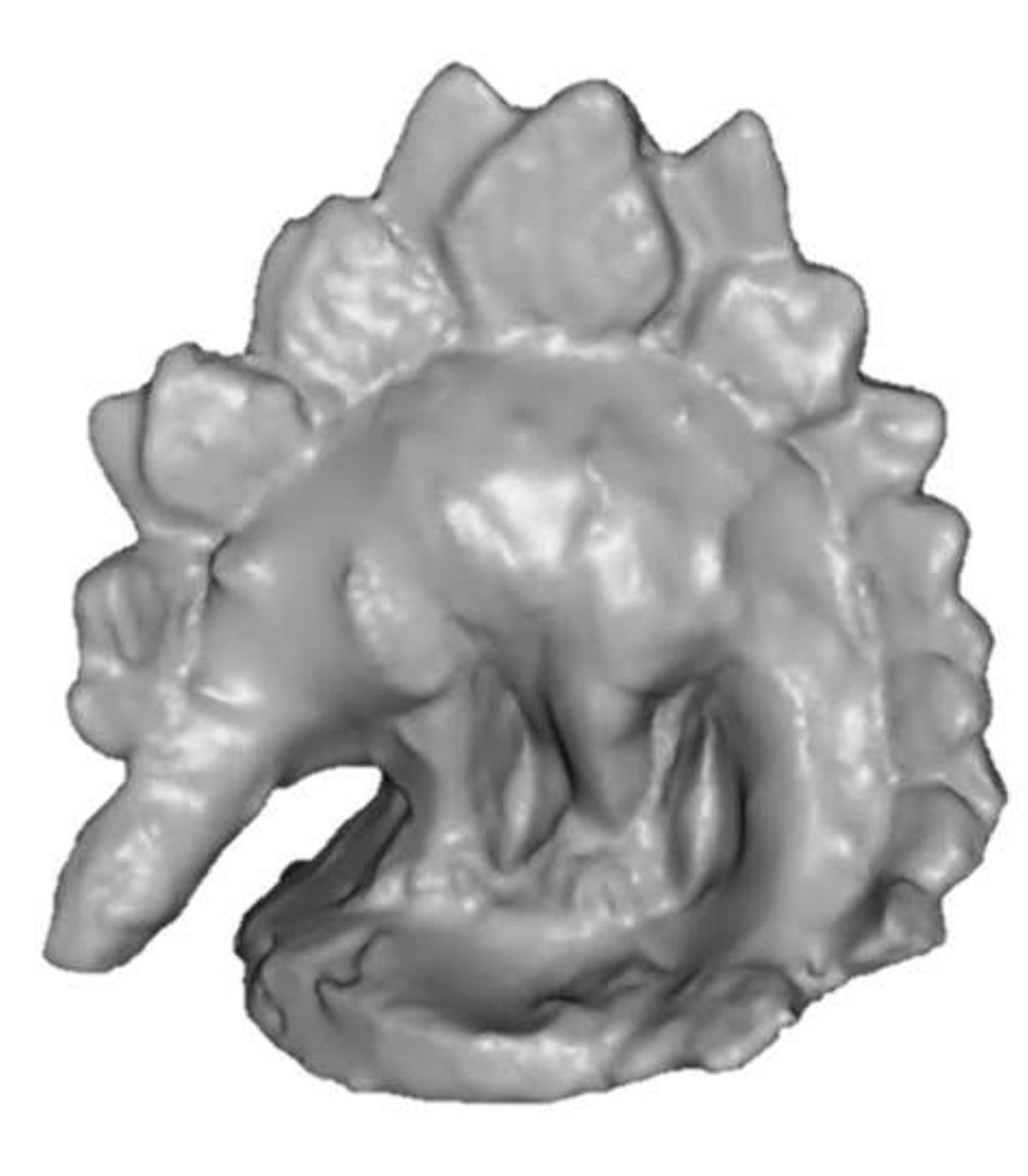}}
 \subfigure{
\includegraphics[height=25mm]{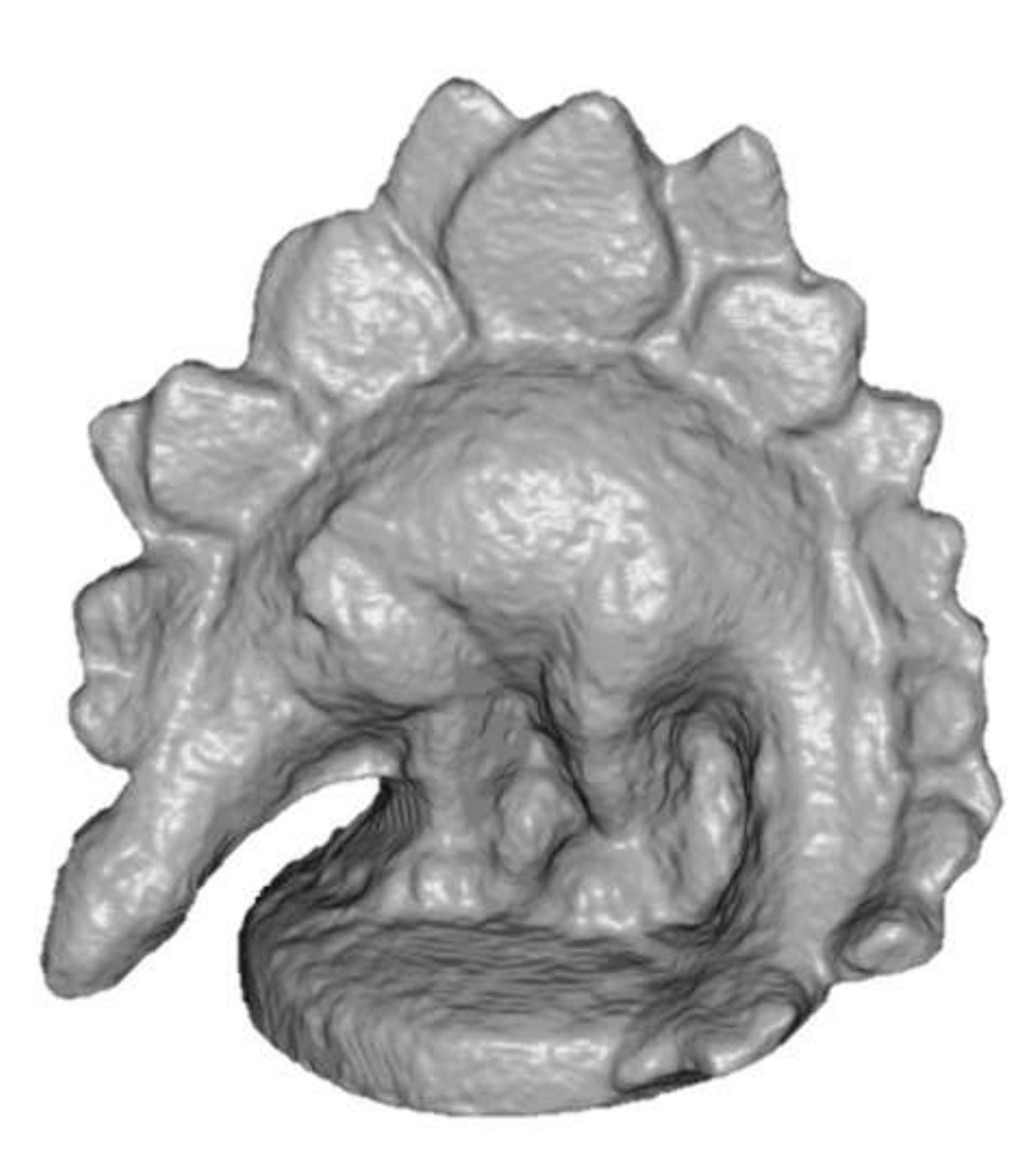}}
 \subfigure{
\includegraphics[height=25mm]{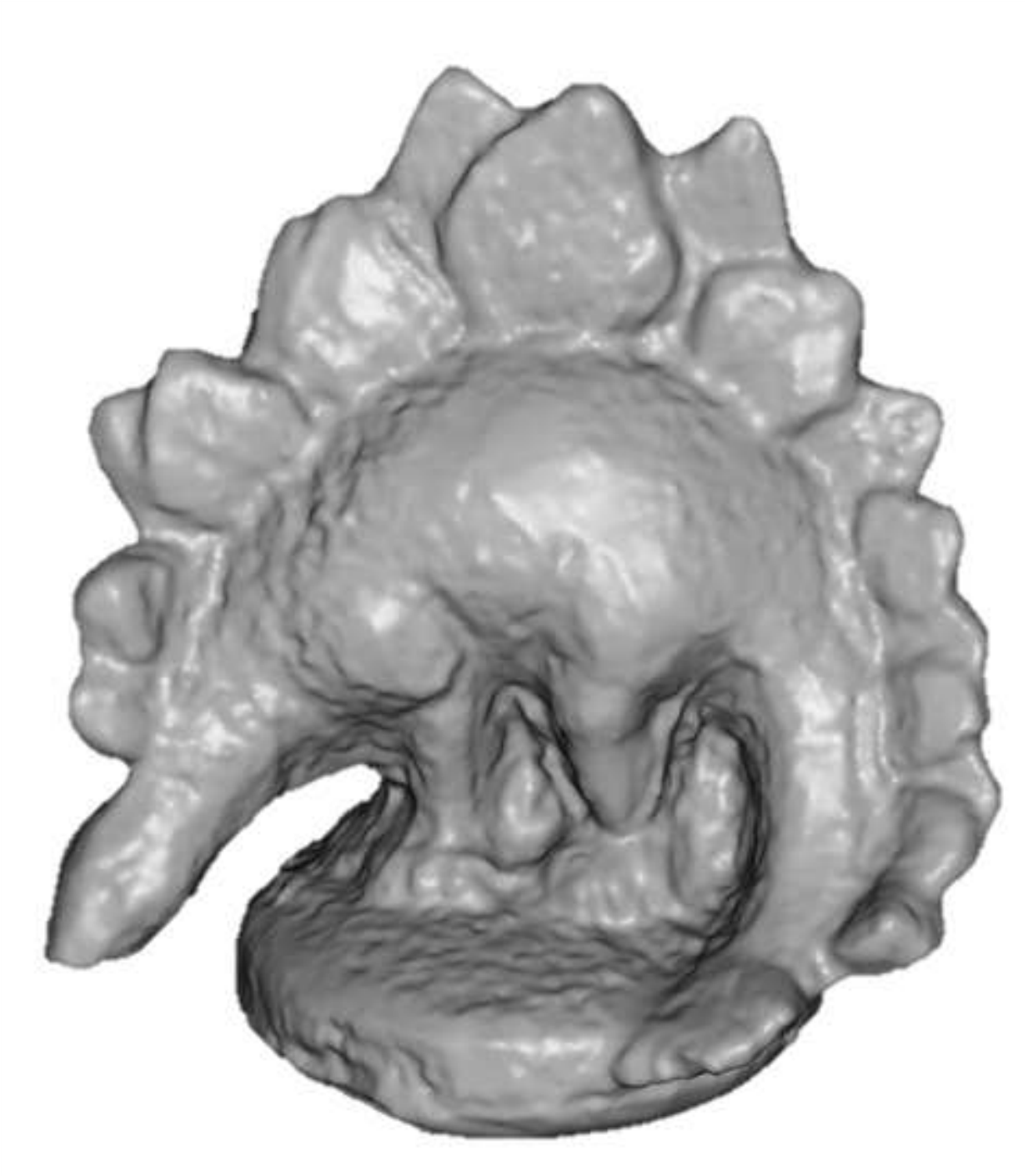}}
\subfigure{
\includegraphics[height=25mm]{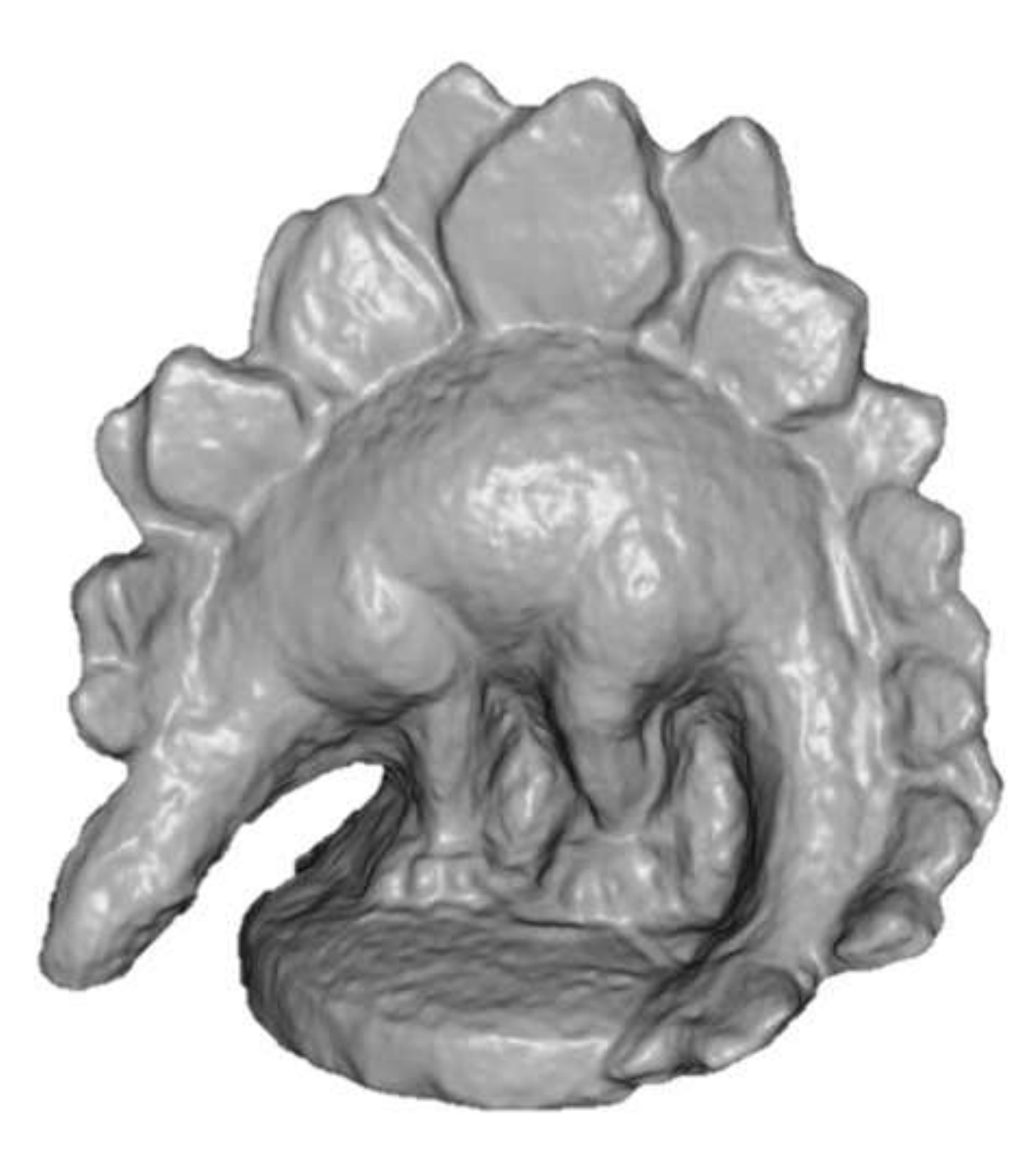}}
\subfigure{
\includegraphics[height=25mm]{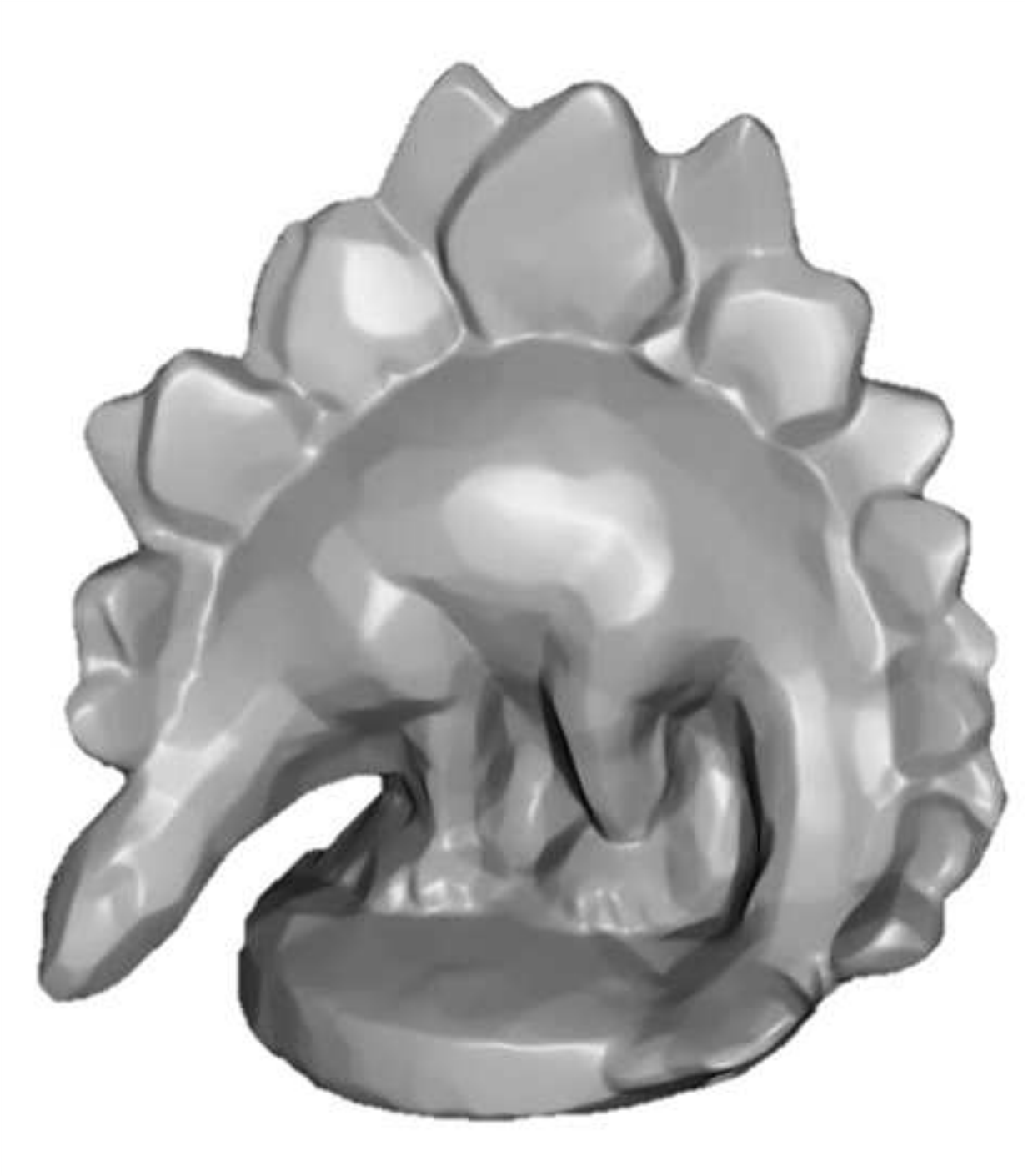}}
 \subfigure{
\includegraphics[height=25mm]{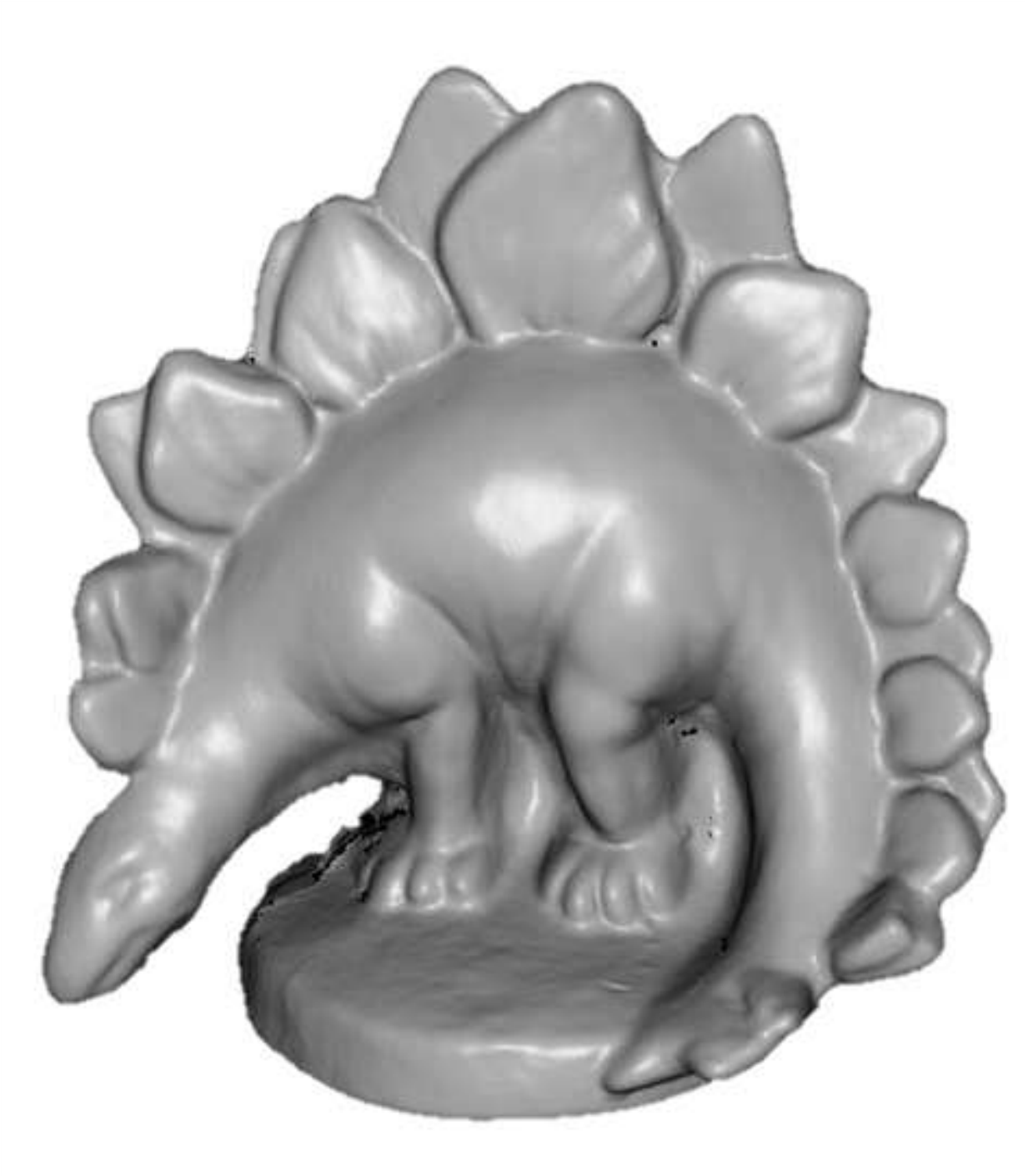}}

\subfigure{
\label{fig:subfig:resultmiddlburyg}
\includegraphics[height=25mm]{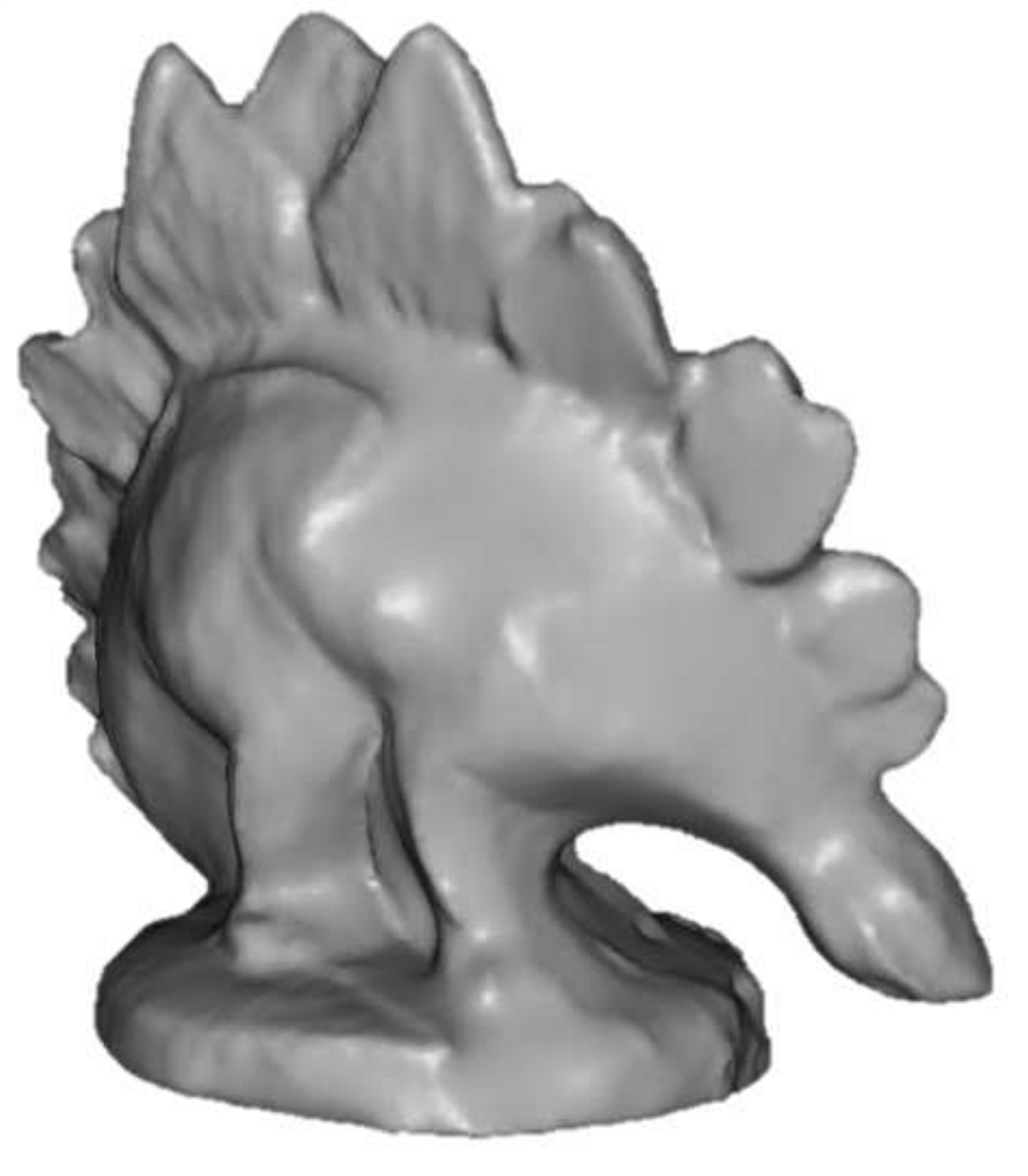}}
 \subfigure{
  \label{fig:subfig:resultmiddlburyh}
\includegraphics[height=25mm]{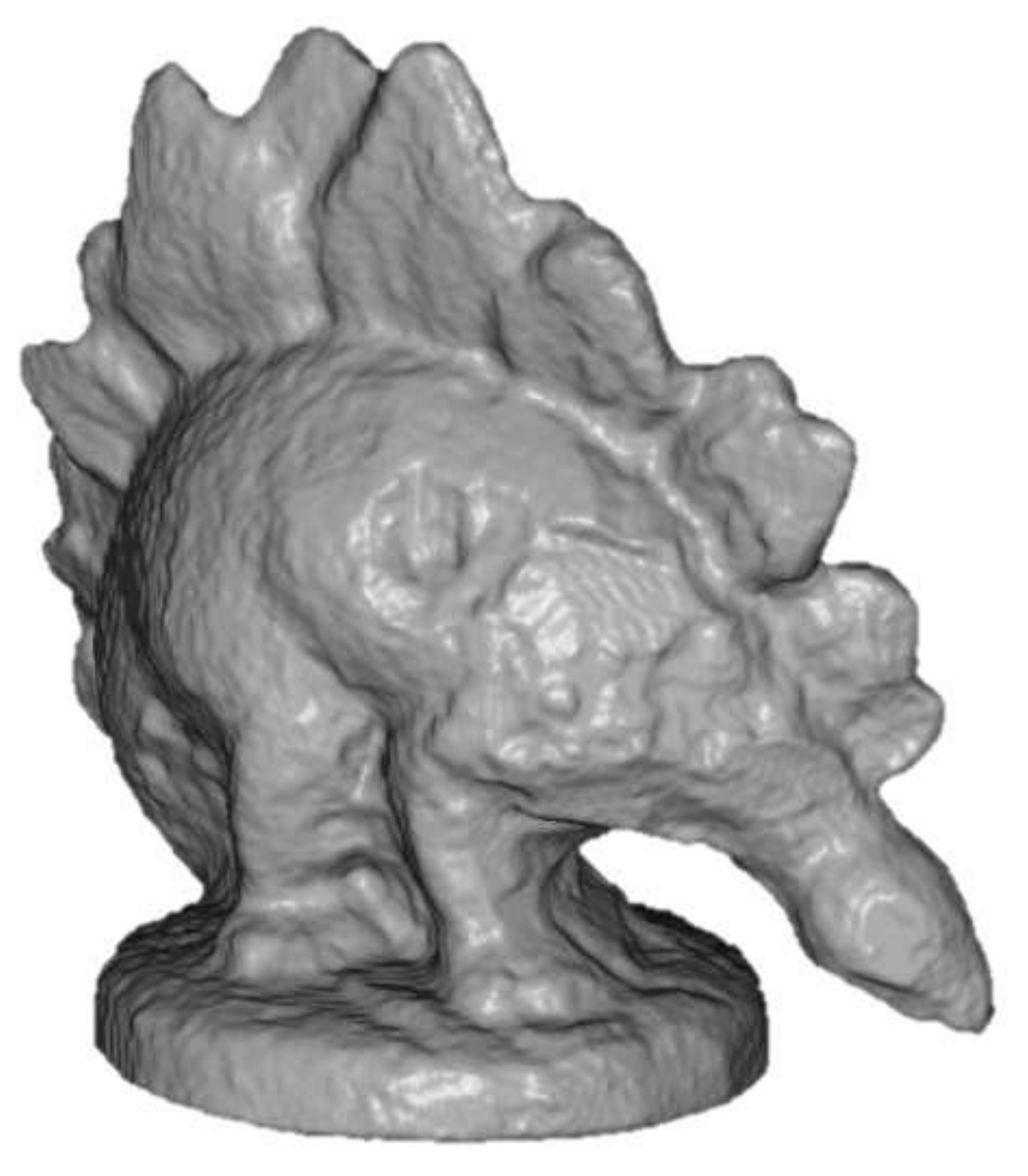}}
 \subfigure{
  \label{fig:subfig:resultmiddlburyi}
\includegraphics[height=25mm]{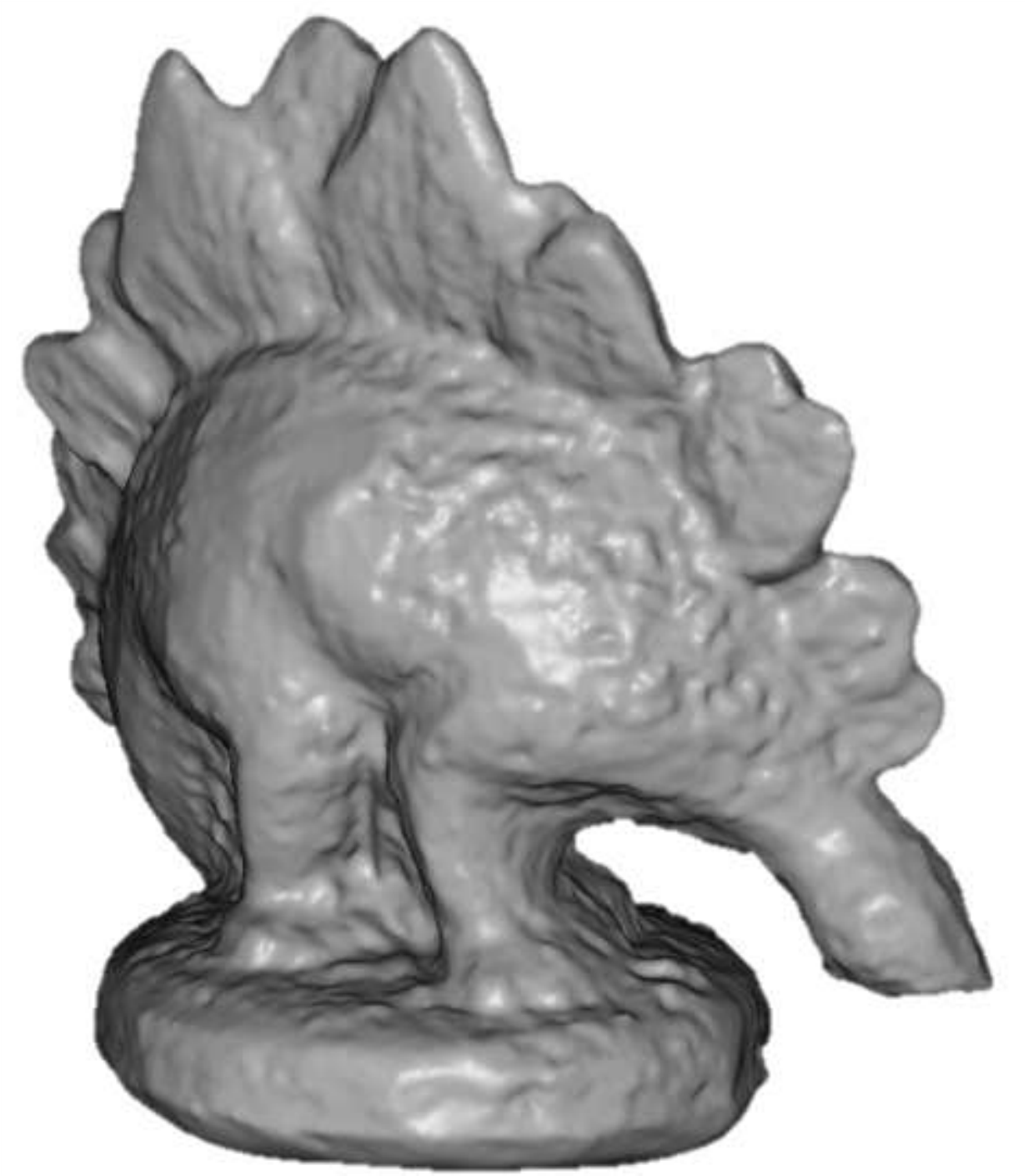}}
\subfigure{
\label{fig:subfig:resultmiddlburyj}
\includegraphics[height=25mm]{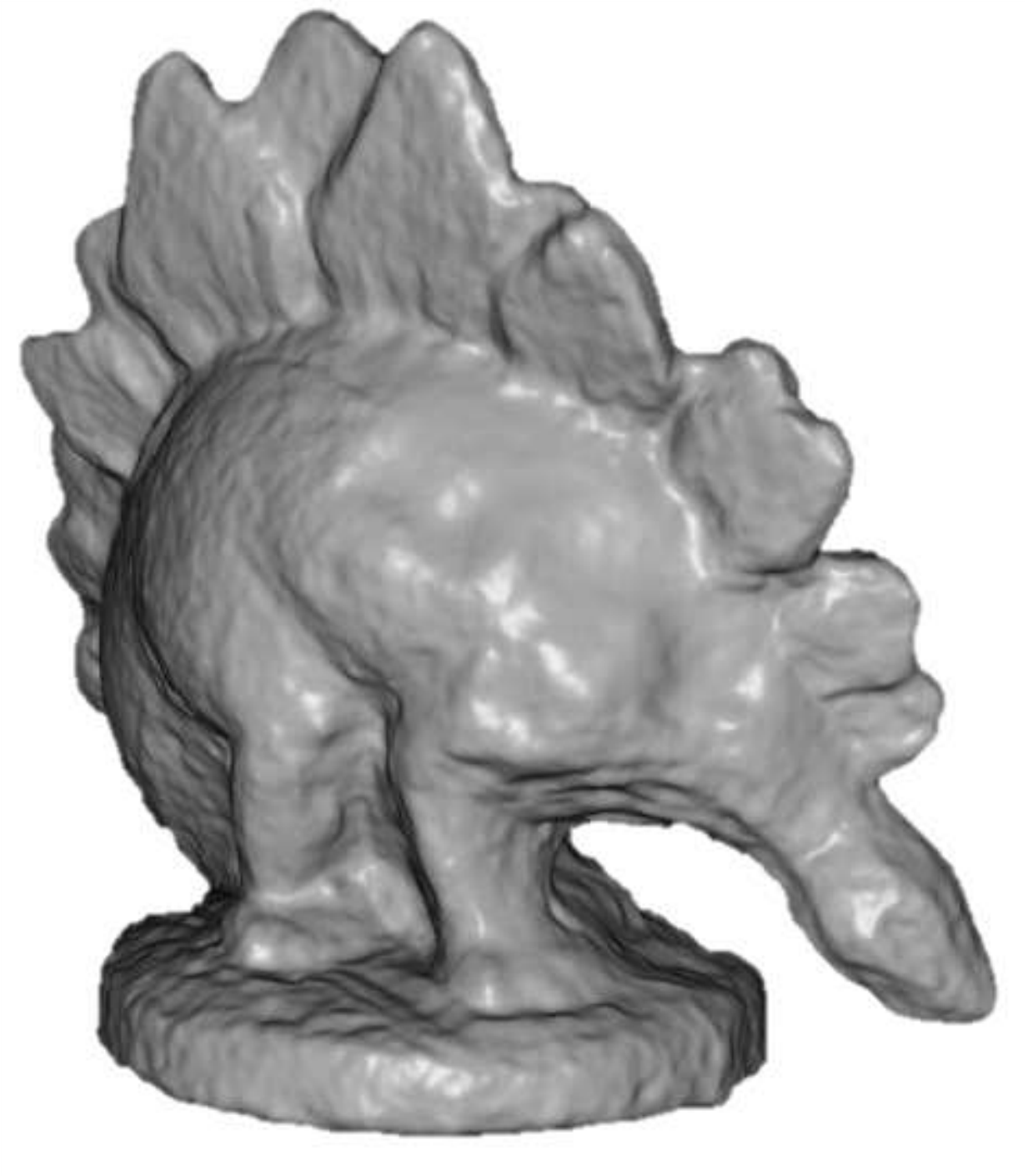}}
 \subfigure{
  \label{fig:subfig:resultmiddlburyk}
\includegraphics[height=25mm]{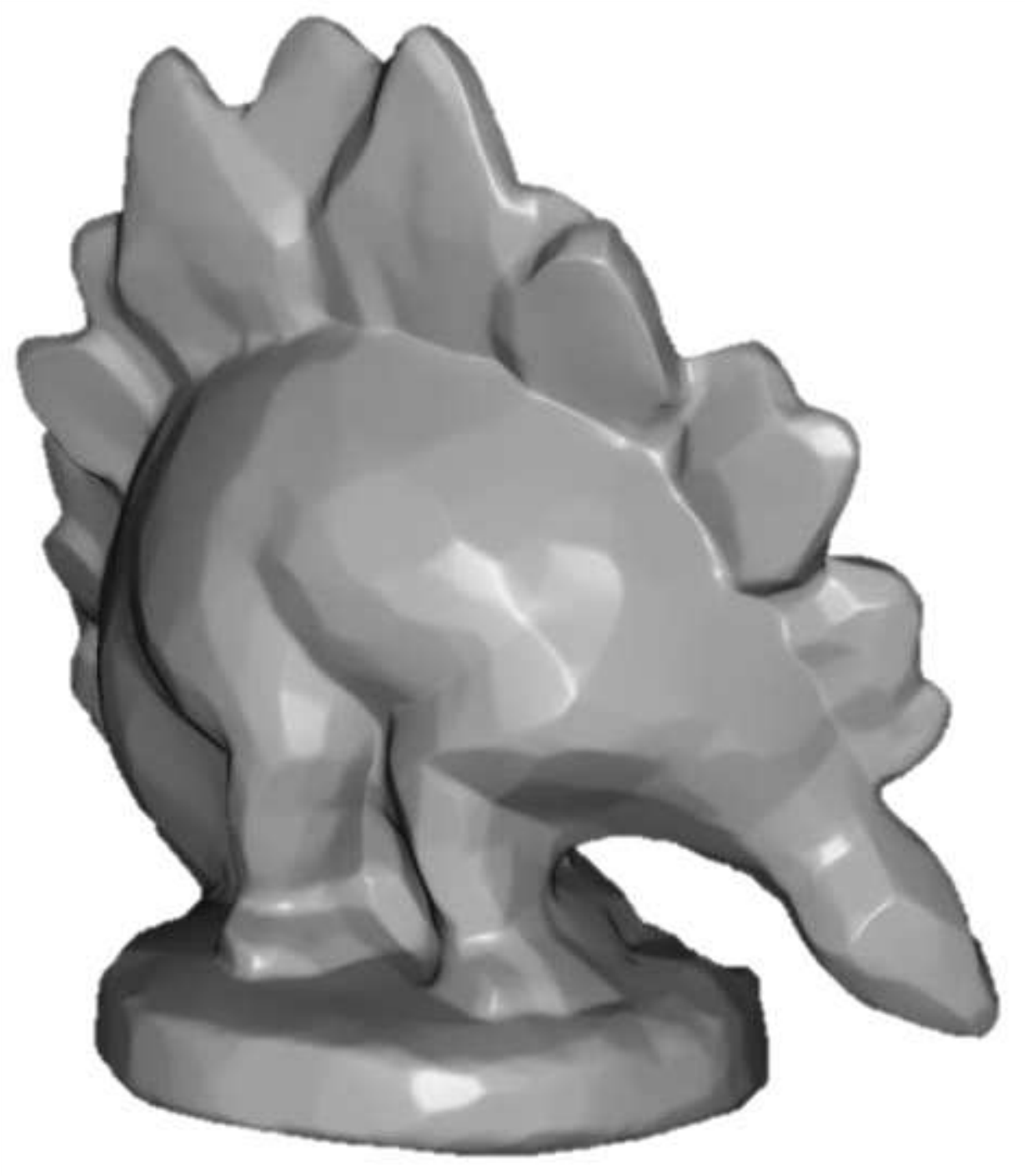}}
 \subfigure{
  \label{fig:subfig:resultmiddlburyl}
\includegraphics[height=25mm]{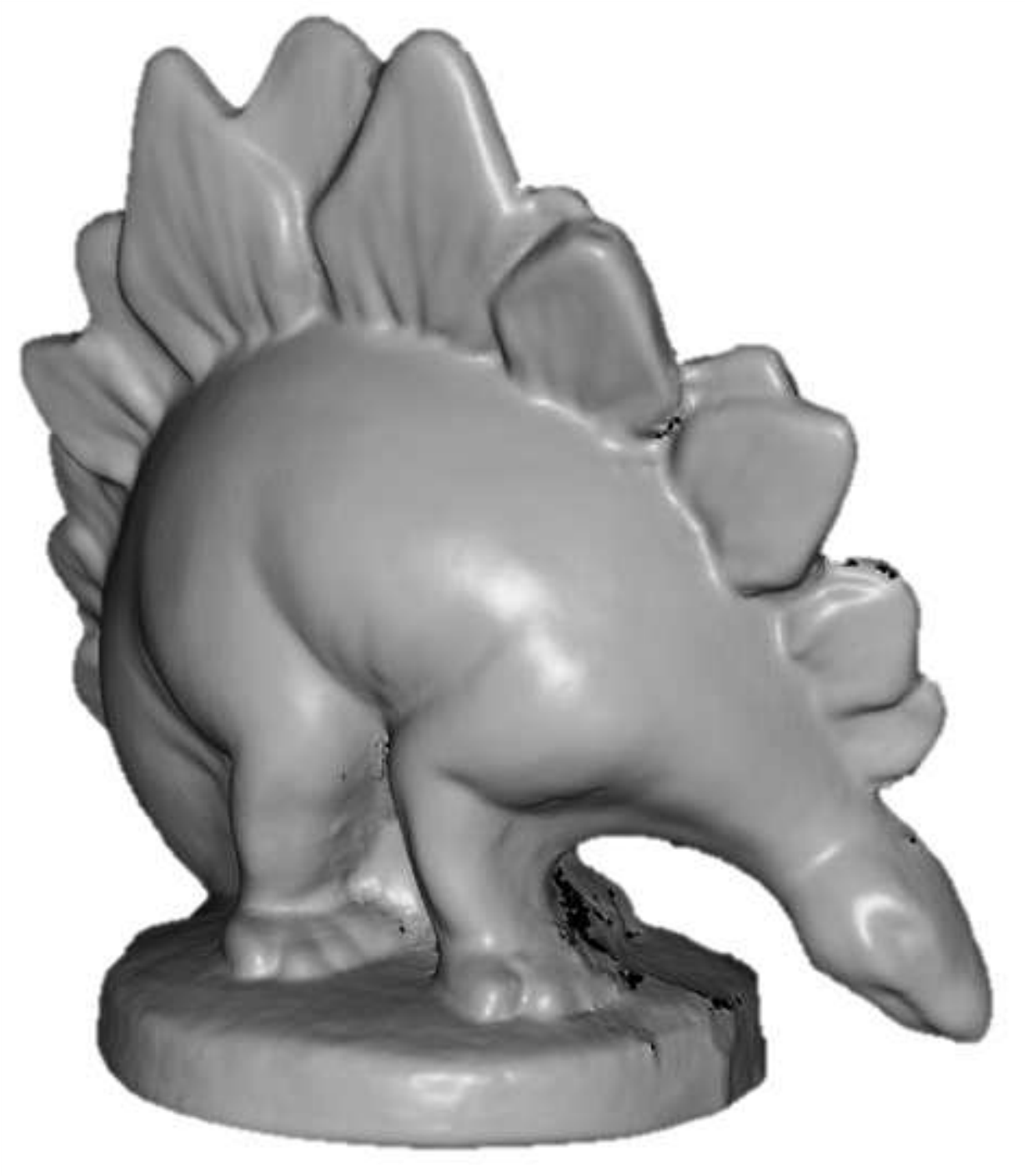}}
\vspace{-3mm}
\caption{Comparison of reconstruction results on the {\it dino sparse} dataset. From left to right: results by ~\cite{24}, ~\cite{16}, ~\cite{18}, ~\cite{31a}, the proposed DCV, and groundtruths, respectively. It is obvious that DCV has better performance in preserving the details and sharp features while filtering the noises.}
\label{fig:middleburyGTComparision}
\vspace{-3mm}
\end{figure*}

 The quantitative comparison results between DCV and several state-of-the-art methods~\cite{2d,18,16a,16,18,23,24,31a} are listed in Table~\ref{table3_DeCoS}. Since some reconstruction results were not reported by authors, we labeled them as '$-$'.  We also conduct a visual comparison with four representative methods~\cite{16,18,24,31a} in Fig. \ref{fig:middleburyGTComparision} on the {\it dino sparse ring} dataset. The methods proposed in \cite{18,24} use the isotropic similarity measure for reprojection error minimization and use isotopic mesh smoothing. The method proposed in~\cite{16} use an anisotropic weighted minimal surface functional. The method\cite{31a} is a combination of patch-based method and isotropic surface refinement.

\input{table3_DeCoS}

\subsection{Results on the Other Datasets} 
We further apply DCV to several other public datasets: the {\it Beethoven} dataset \cite{16a}, the {\it bird} dataset \cite{16}, the {\it fountain-P11} dataset \cite{2}, and the {\it statuegirl} dataset~\cite{acute3d}. We also conduct experiments on two real datasets collected by us: {\it Buddha} and {\it bell}. Since groundtruths of these datasets are not available \footnote{Although there was an evaluation system for quantitatively evaluating the {\it fountain-P11} dataset, the evaluation service is currently unavailable}, qualitative evaluation of the reconstruction results is adopted in the experiment.


\begin{figure*}[tbp]
\centering
\subfigure{
\label{fig:subfig:beethoven_a}
\includegraphics[height=24mm]{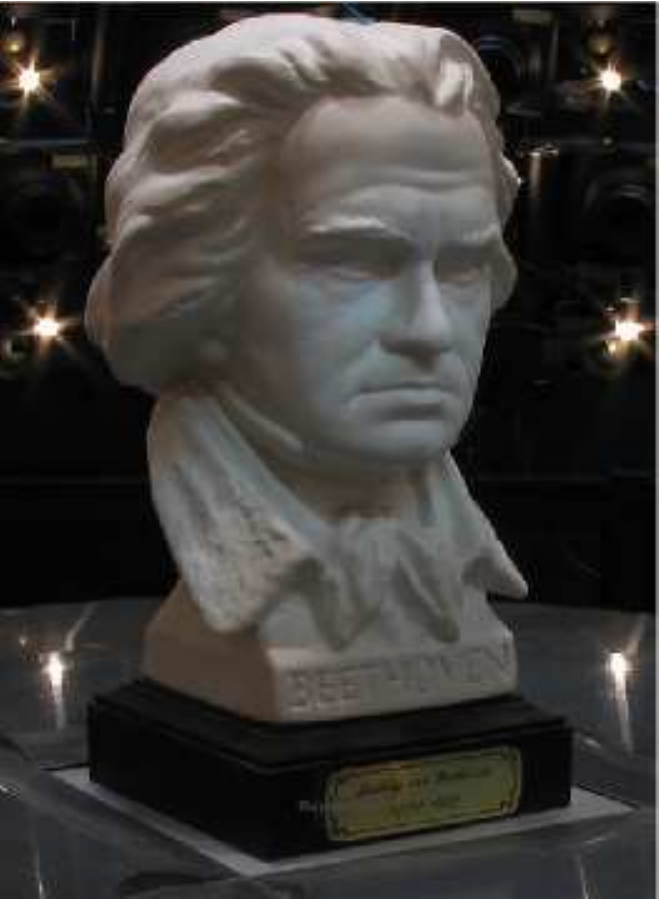}}
 \subfigure{
  \label{fig:subfig:beethoven_b}
\includegraphics[height=24mm]{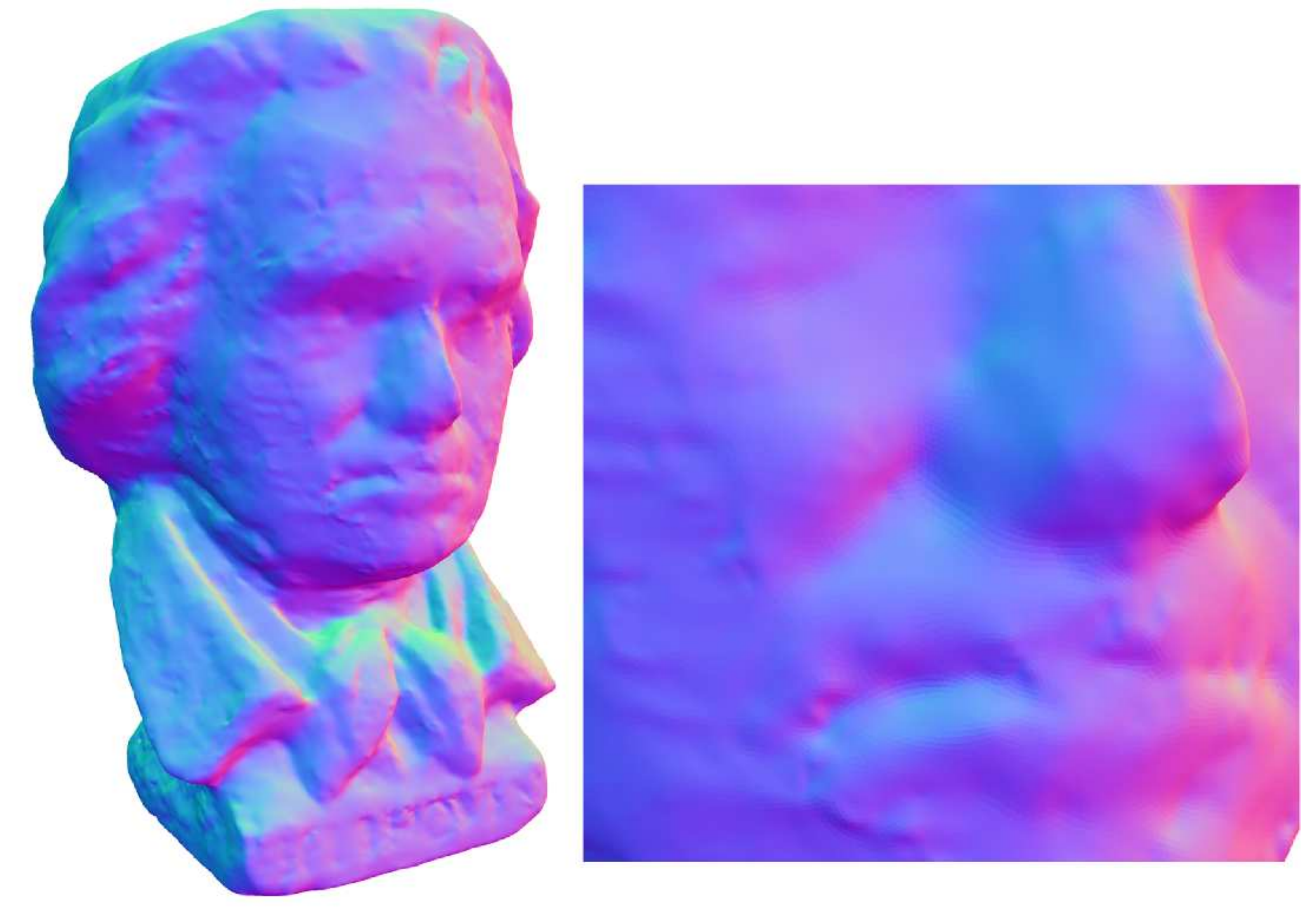}}
 \subfigure{
  \label{fig:subfig:beethoven_c}
\includegraphics[height=24mm]{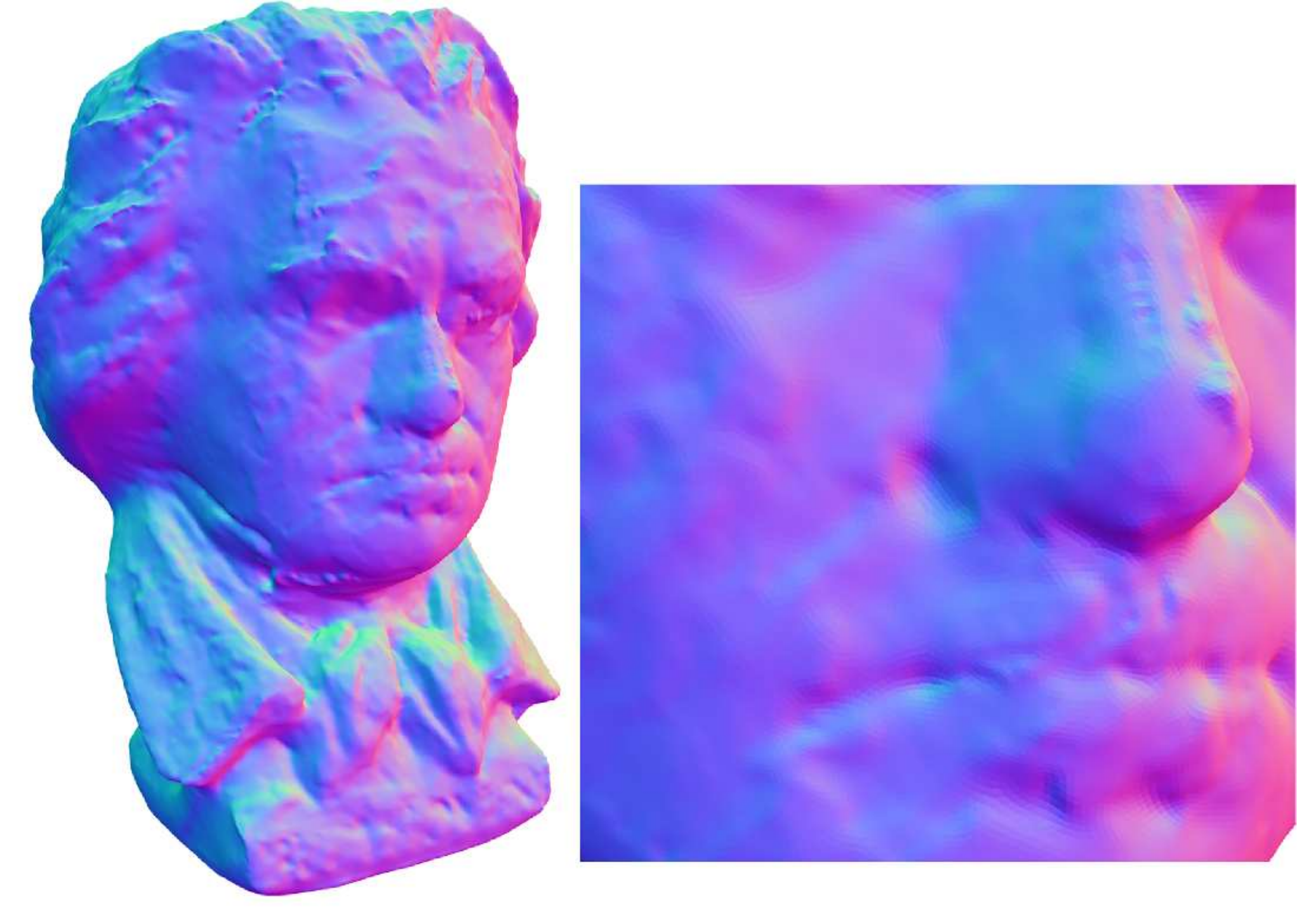}}
\subfigure{
\label{fig:subfig:beethoven_d}
\includegraphics[height=24mm]{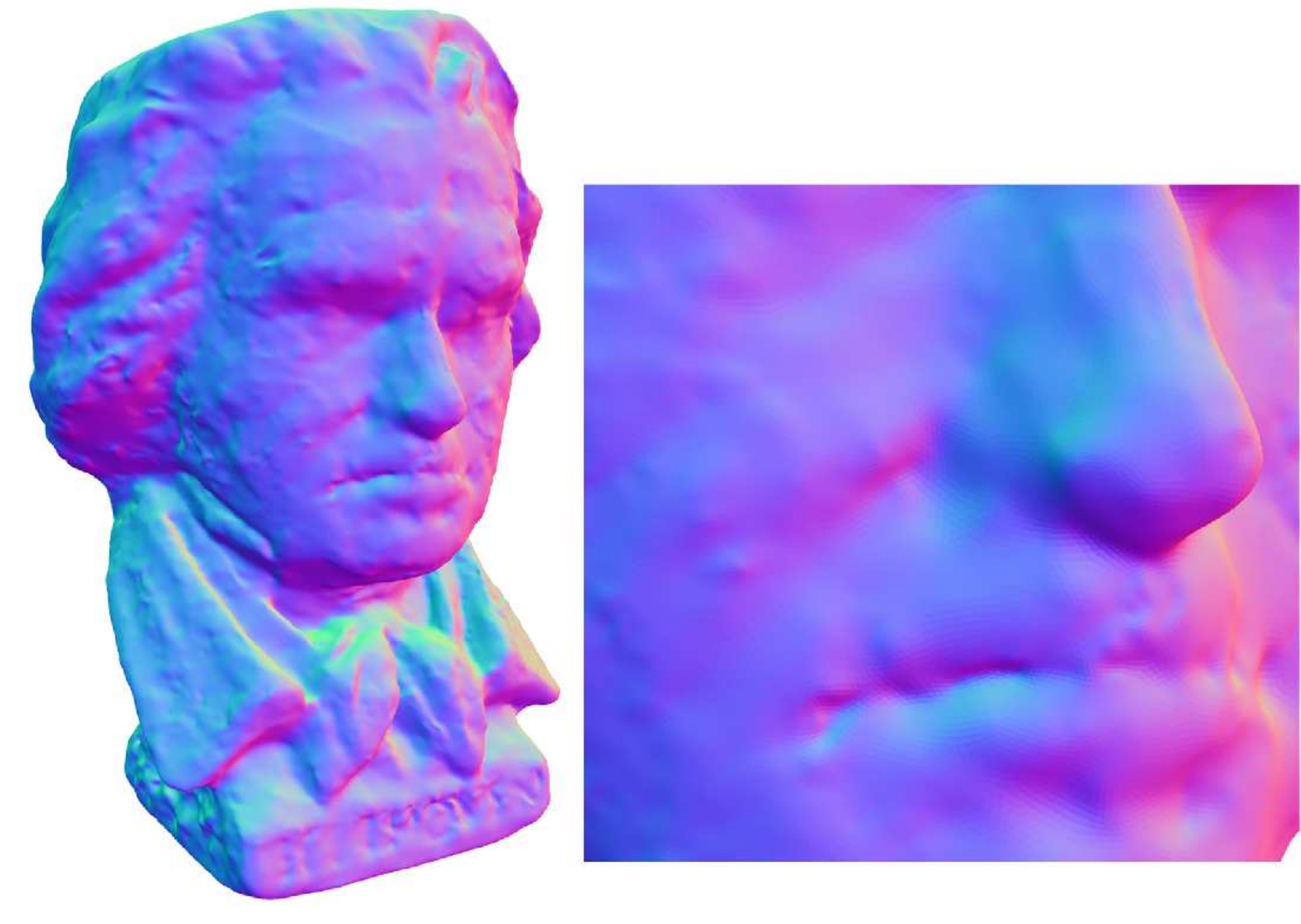}}
\subfigure{
\label{fig:subfig:beethoven_e}
\includegraphics[height=24mm]{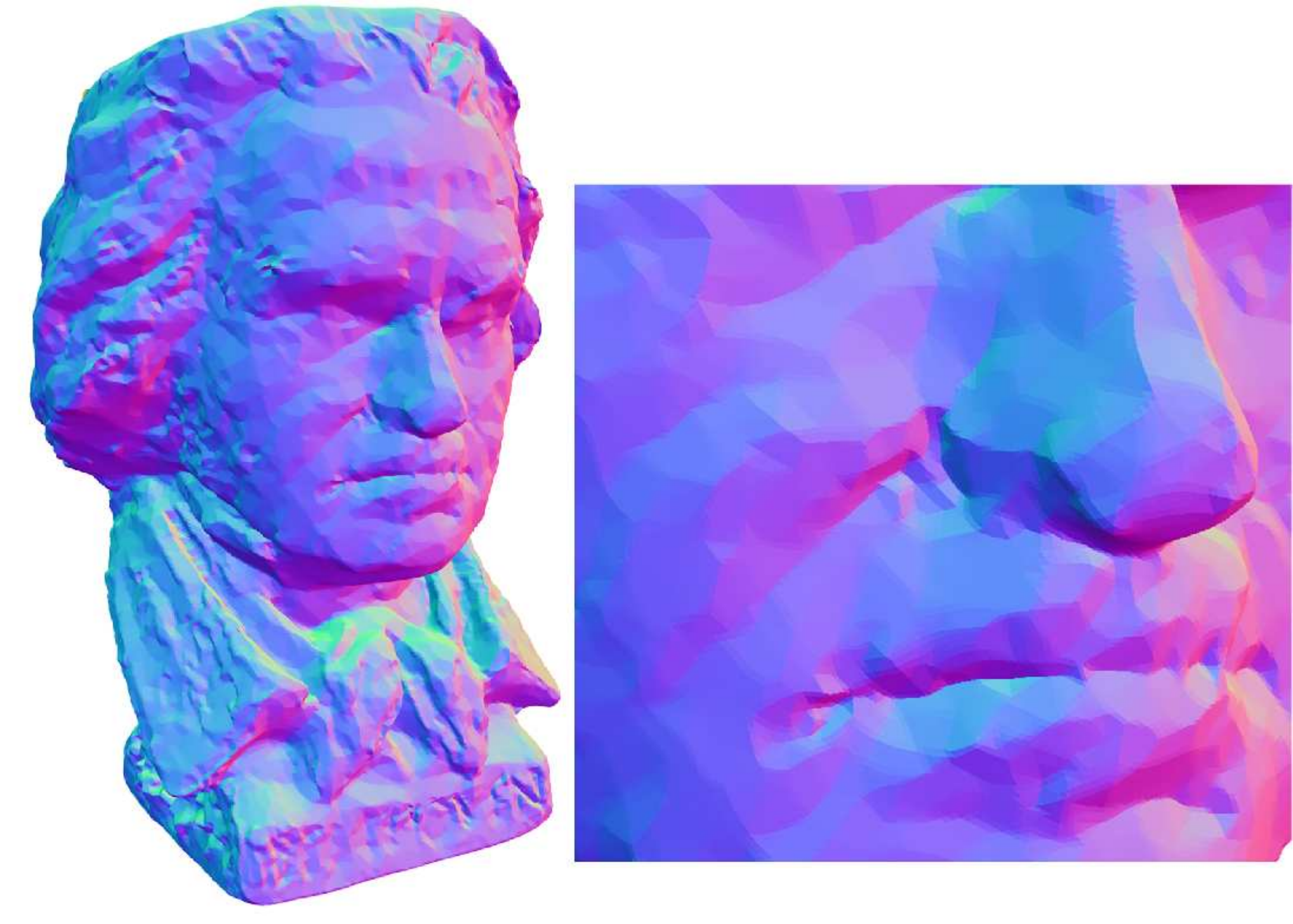}}
\subfigure{
 \label{fig:subfig:beethoven_f}
\includegraphics[height=23mm]{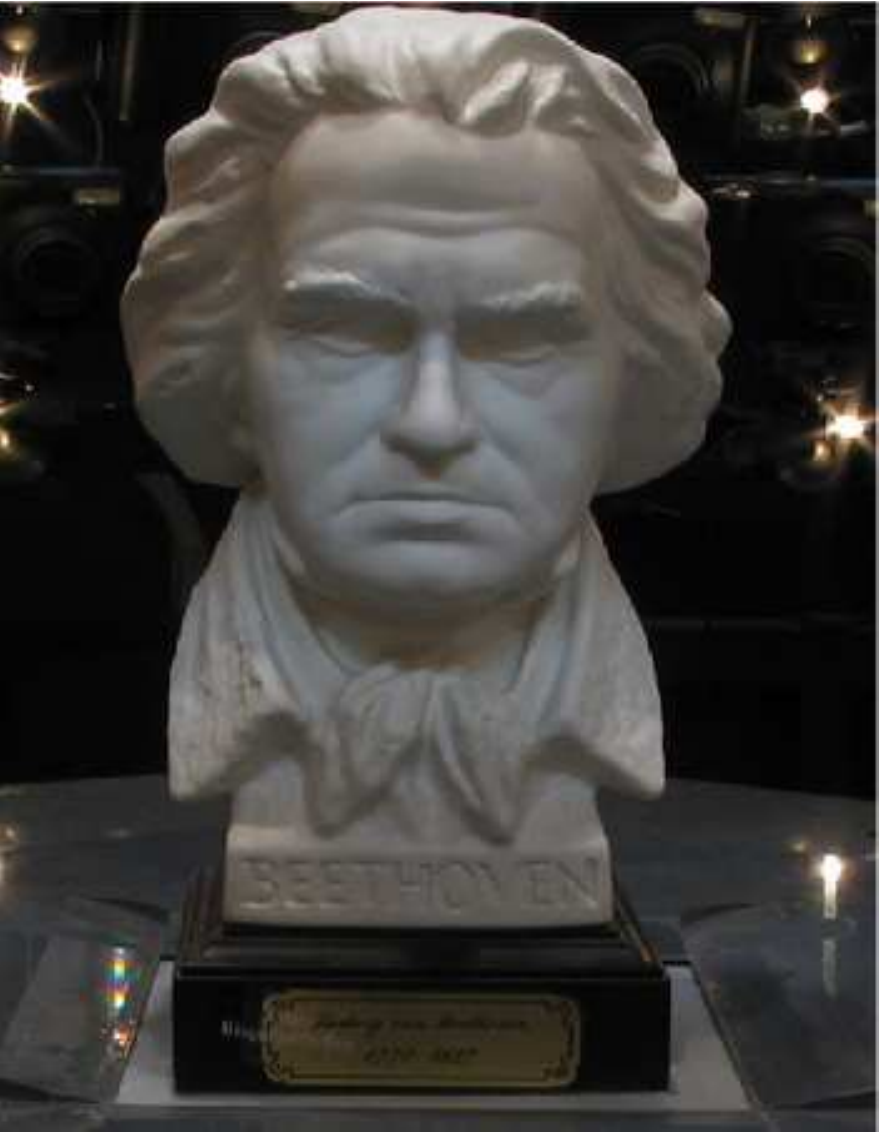}}
 \subfigure{
  \label{fig:subfig:beethoven_g}
\includegraphics[height=23mm]{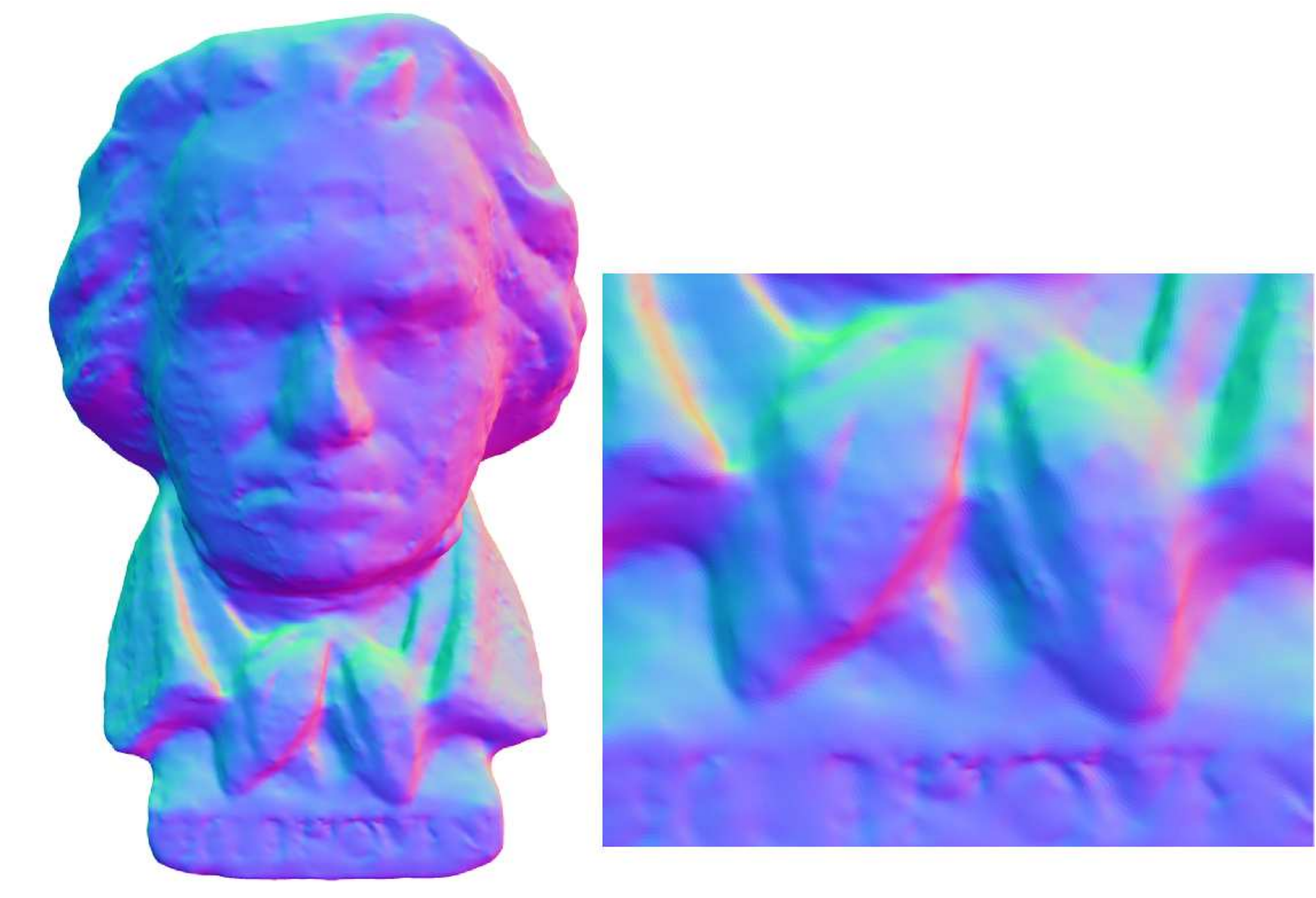}}
\subfigure{
\label{fig:subfig:beethoven_h}
\includegraphics[height=23mm]{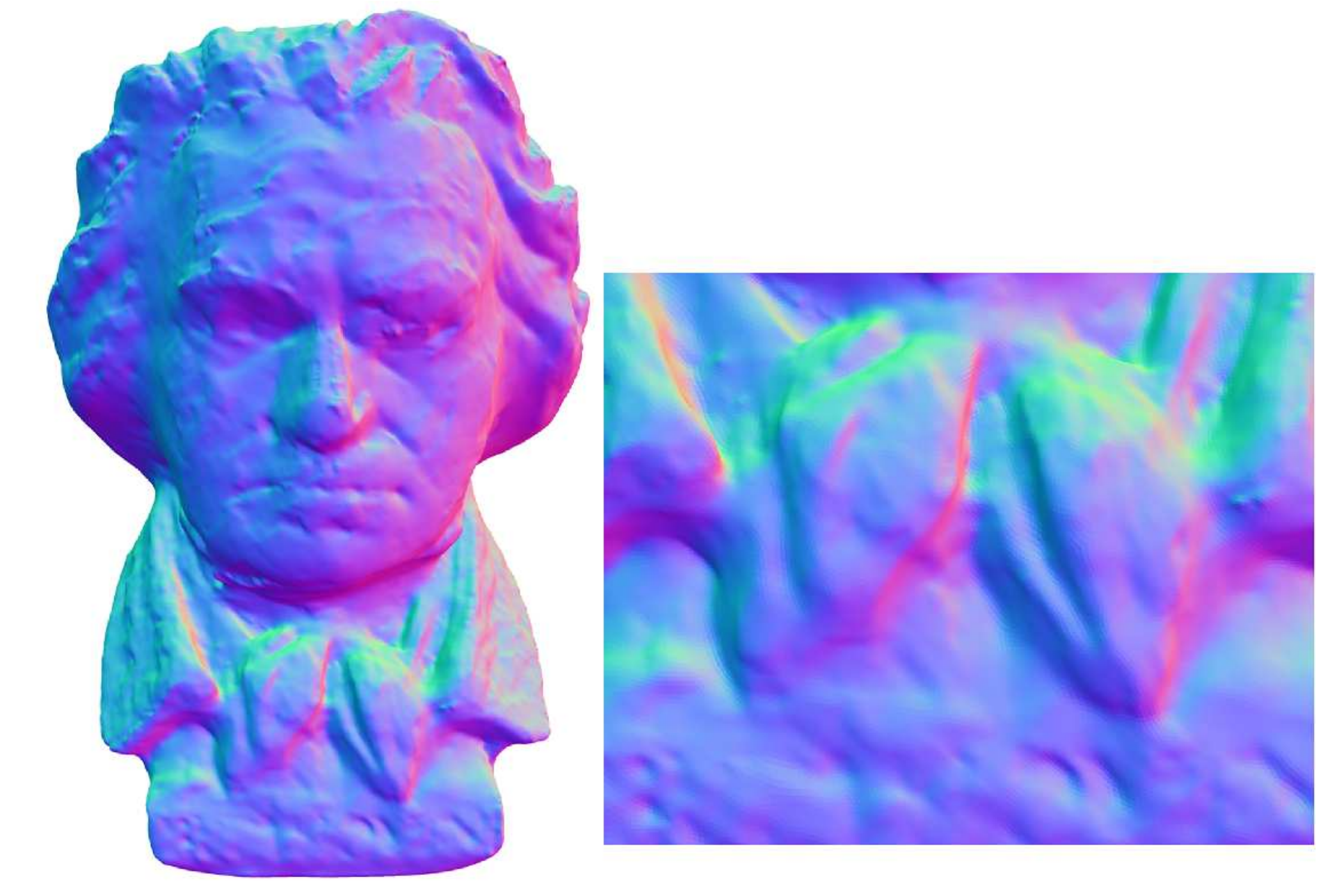}}
 \subfigure{
  \label{fig:subfig:beethoven_i}
\includegraphics[height=23mm]{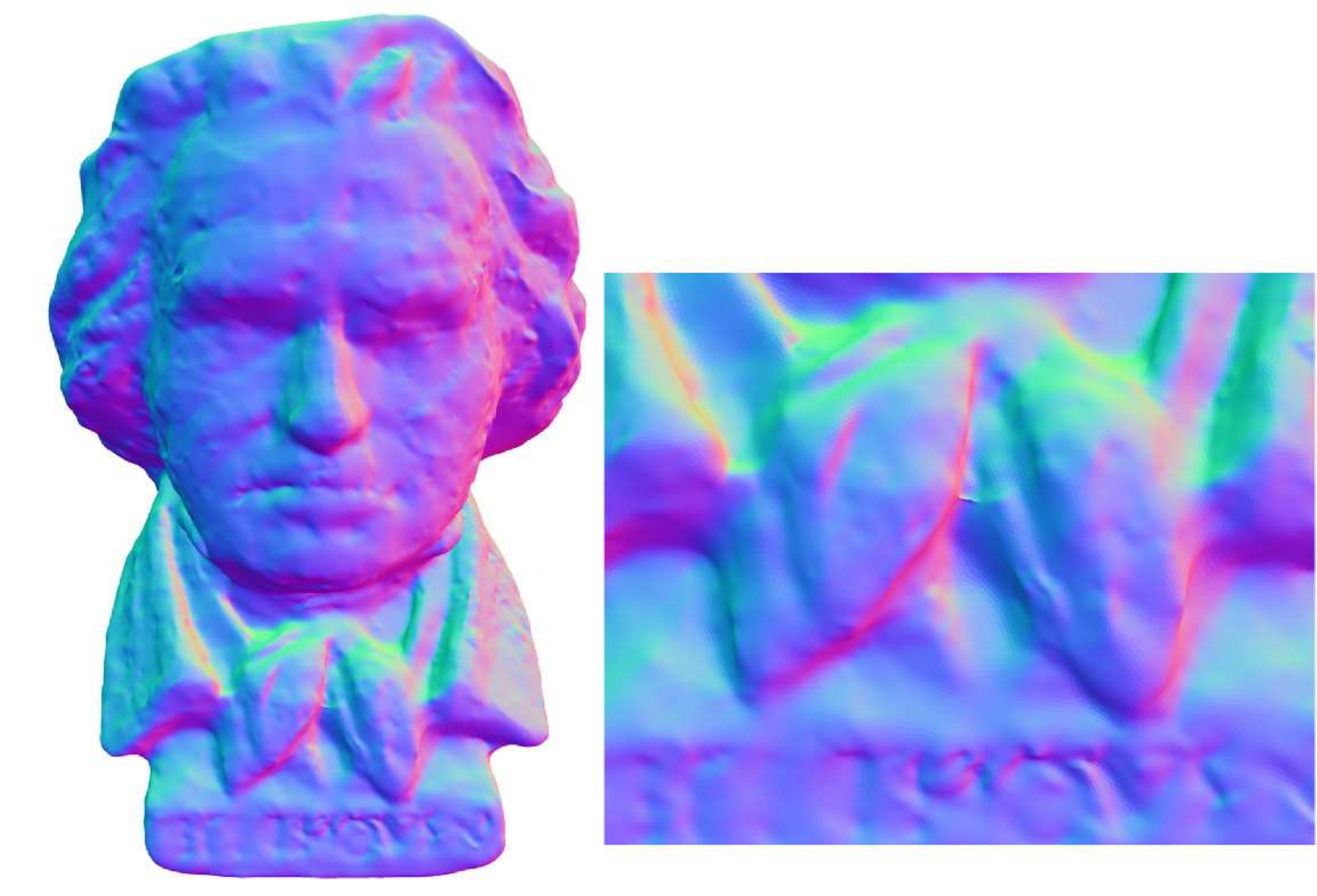}}
 \subfigure{
  \label{fig:subfig:beethoven_j}
\includegraphics[height=23mm]{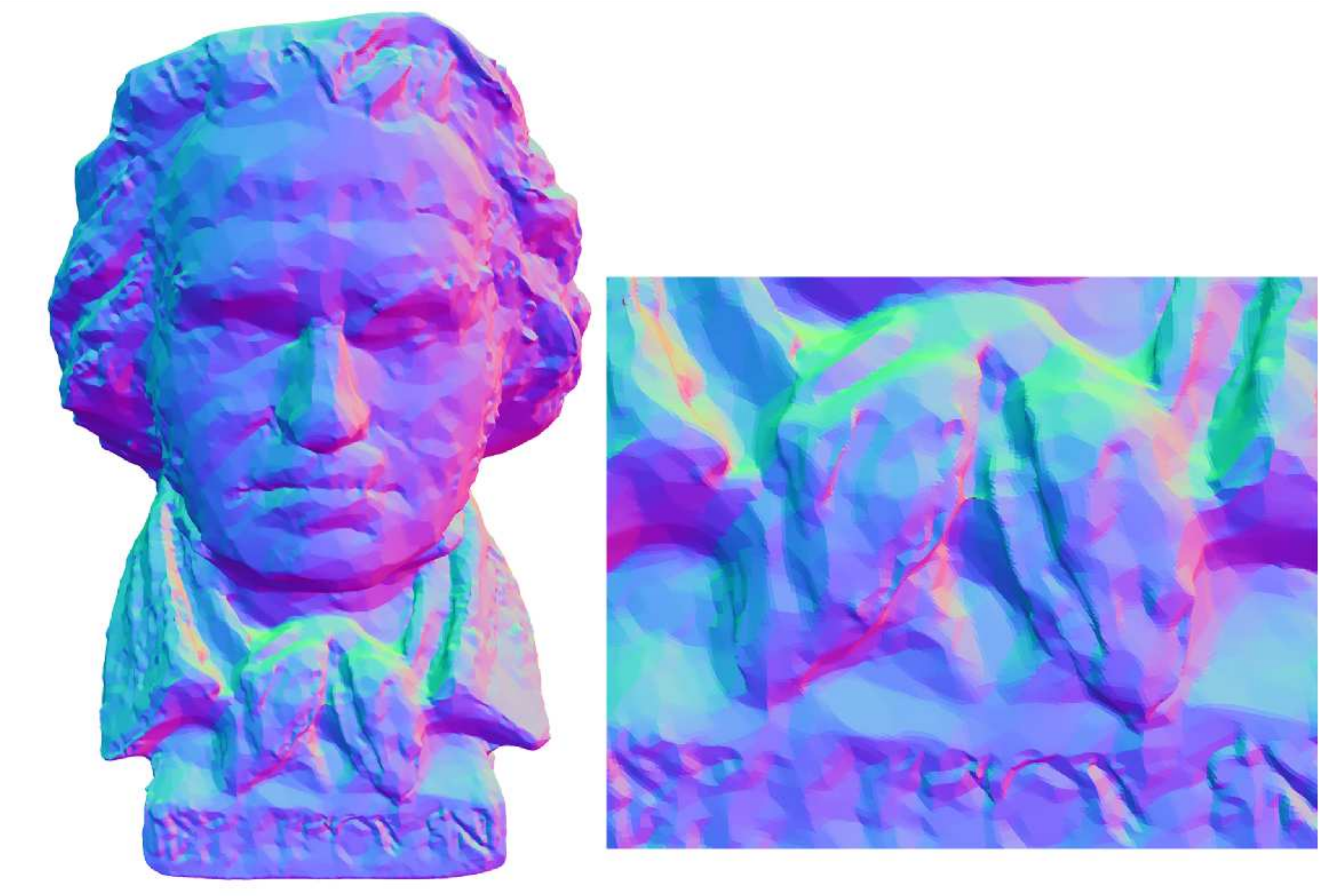}}
\caption{Reconstruction results by several state-of-the-art methods and the proposed DCV on the {\it Beethoven} datasets. From left column to right column: input images, results by ~\cite{18},  ~\cite{24}, ~\cite{20}, and DCV in two views, respectively.}
\label{fig:beethoven_result}
\end{figure*}

\begin{figure*}[tbp]
\centering
\subfigure{
\label{fig:subfig:bird_a}
\includegraphics[height=20mm]{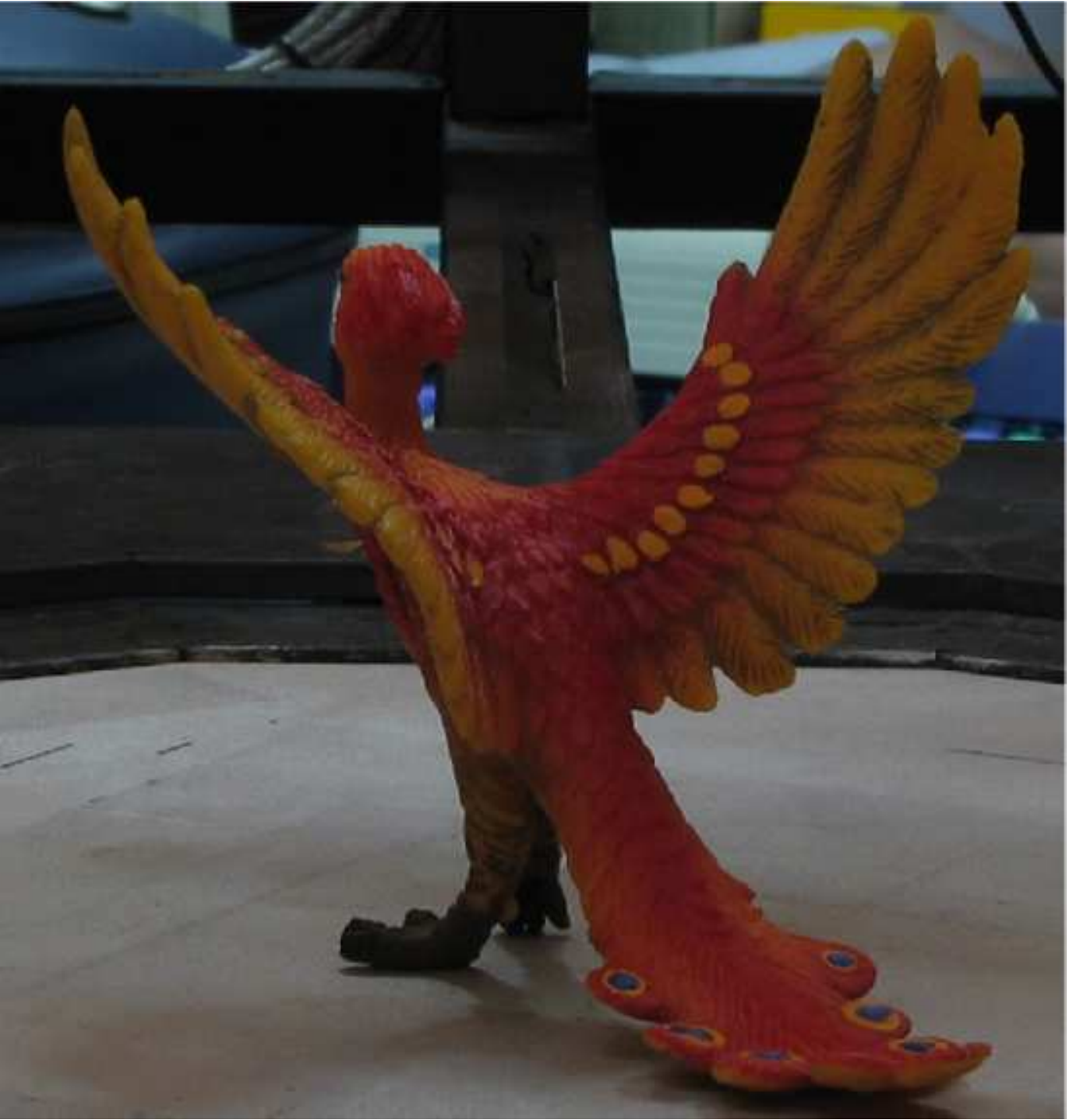}}
 \subfigure{
  \label{fig:subfig:bird_b}
\includegraphics[height=20mm]{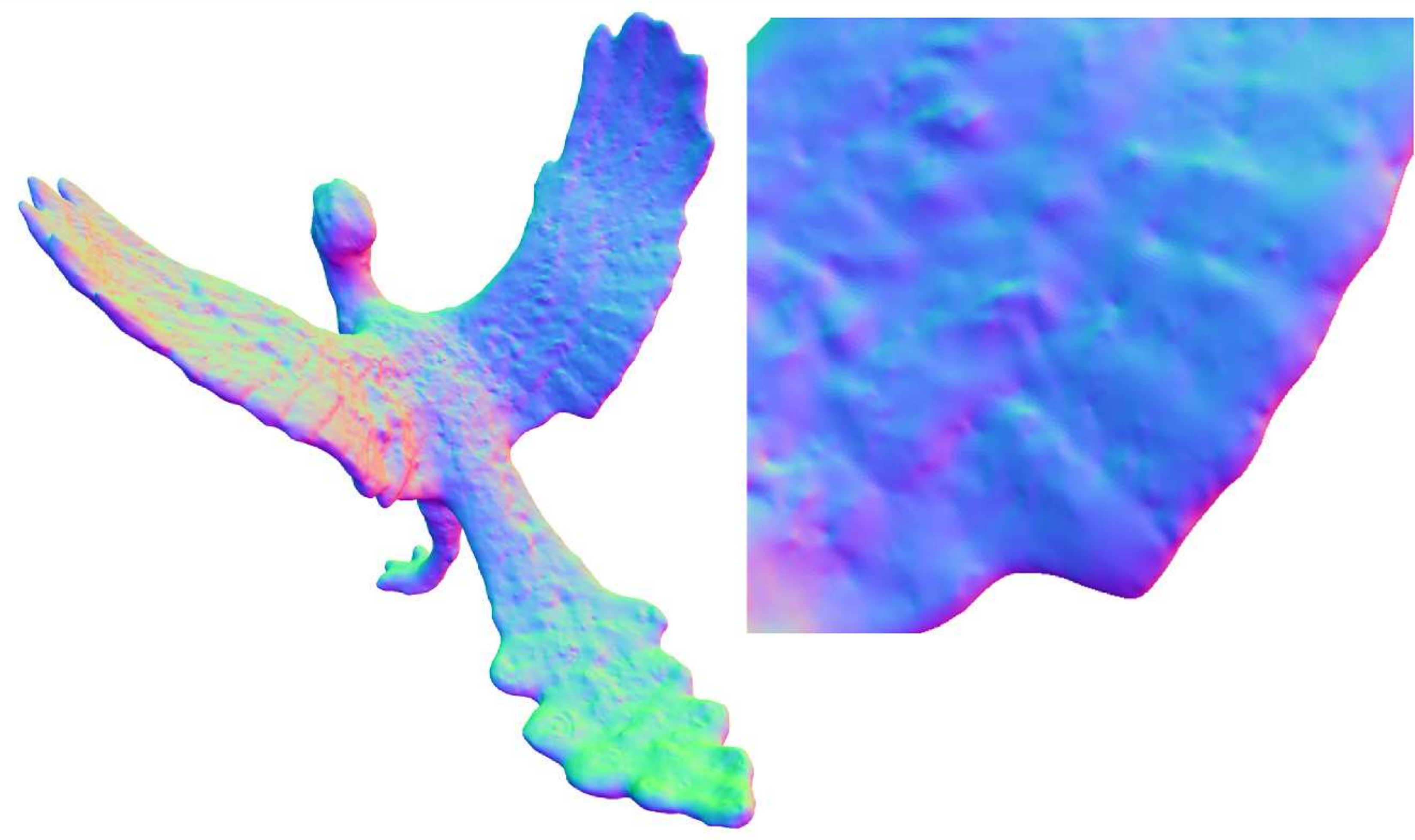}}
 \subfigure{
  \label{fig:subfig:bird_c}
\includegraphics[height=20mm]{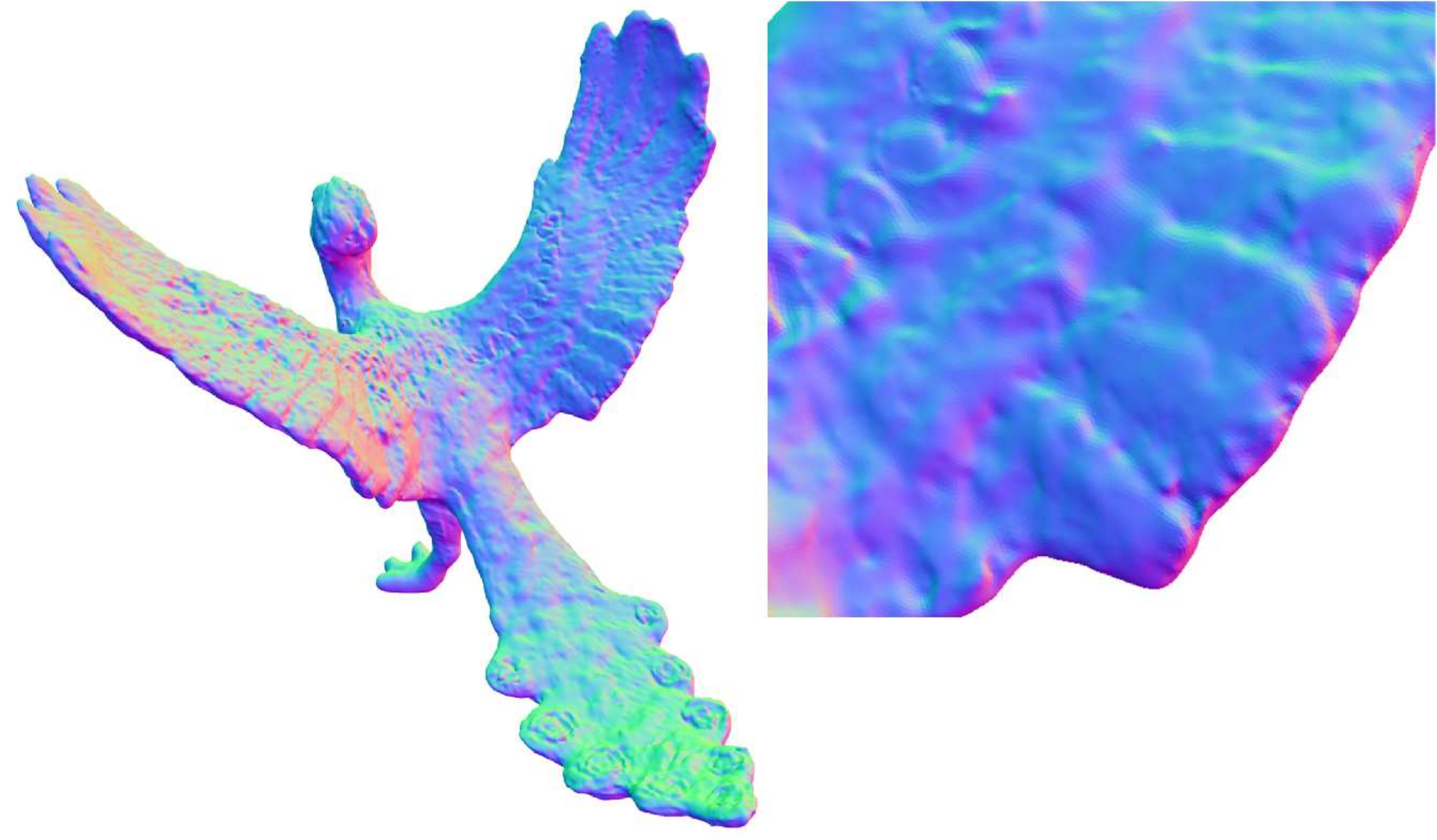}}
\subfigure{
\label{fig:subfig:bird_d}
\includegraphics[height=20mm]{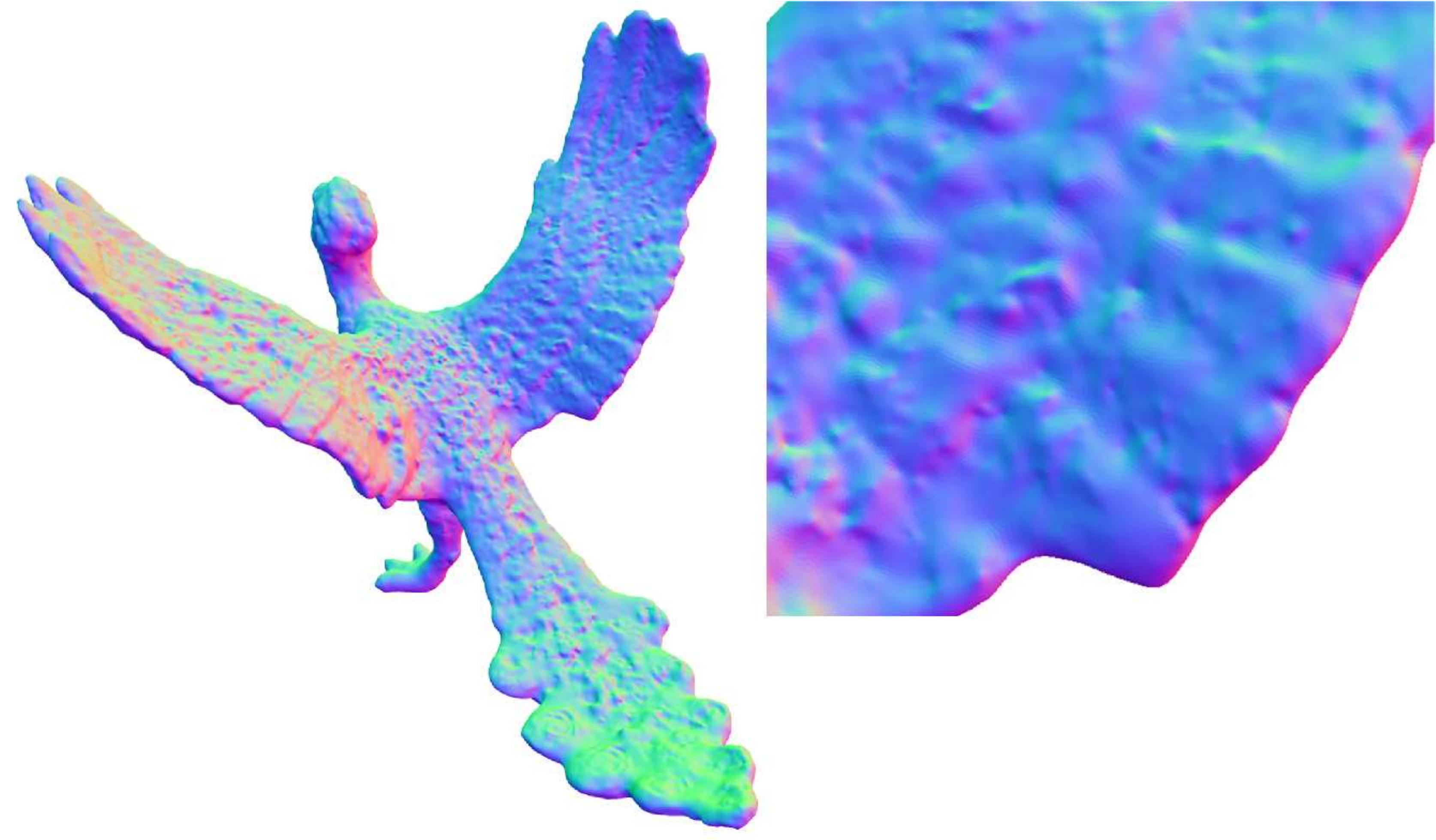}}
\subfigure{
\label{fig:subfig:bird_e}
\includegraphics[height=20mm]{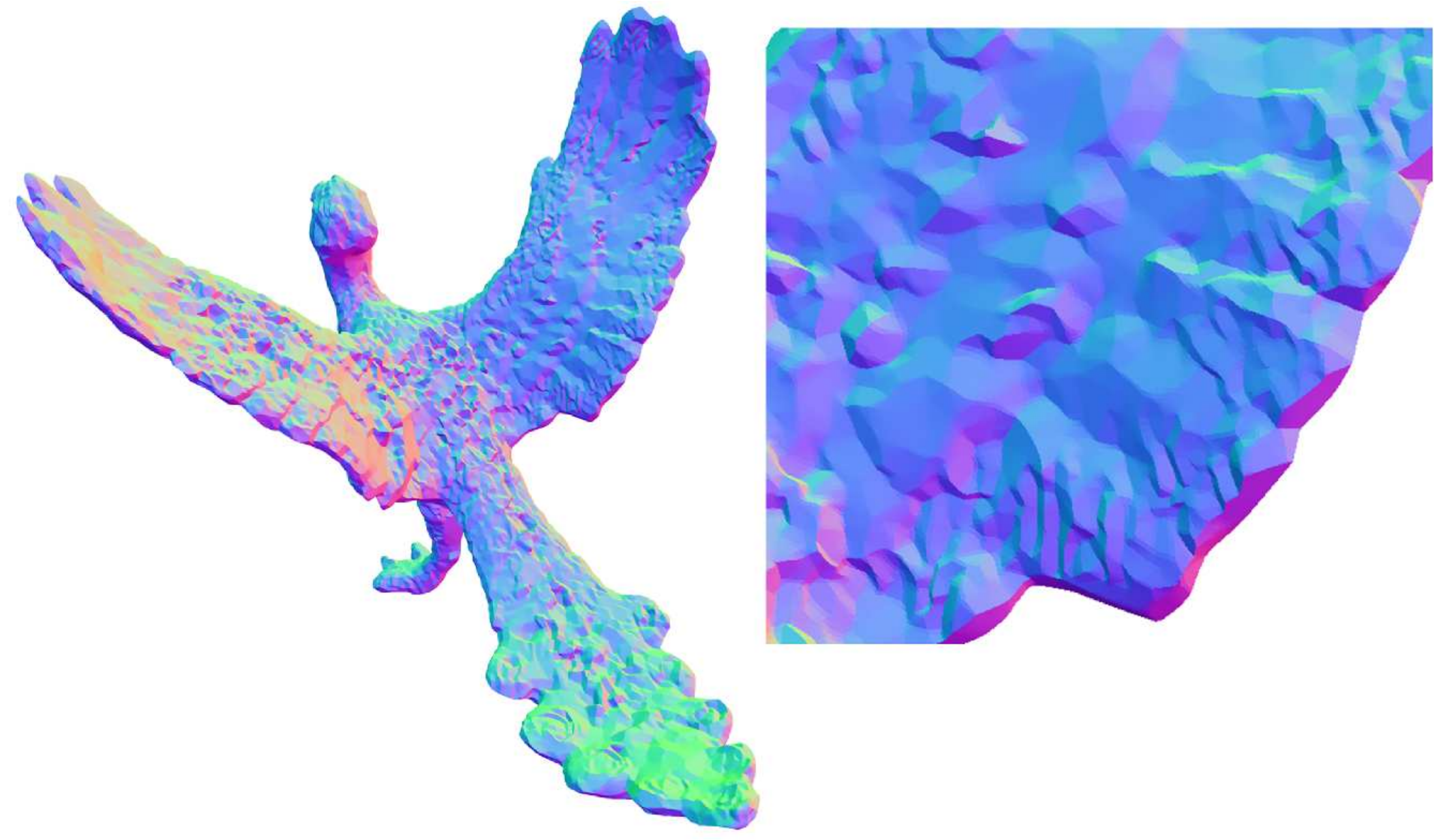}}

\subfigure{
 \label{fig:subfig:bird_f}
\includegraphics[height=20mm]{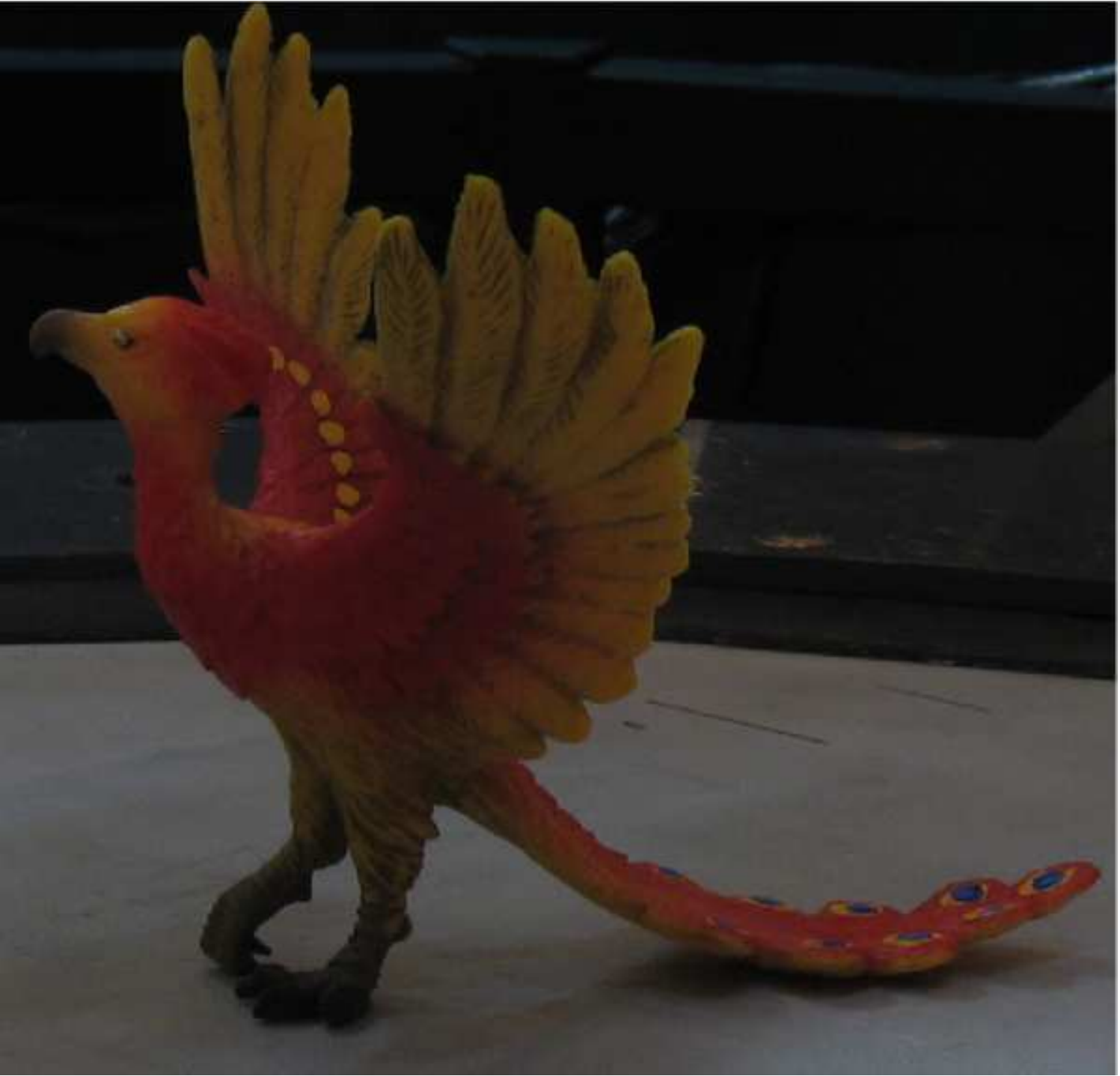}}
 \subfigure{
  \label{fig:subfig:bird_g}
\includegraphics[height=21mm]{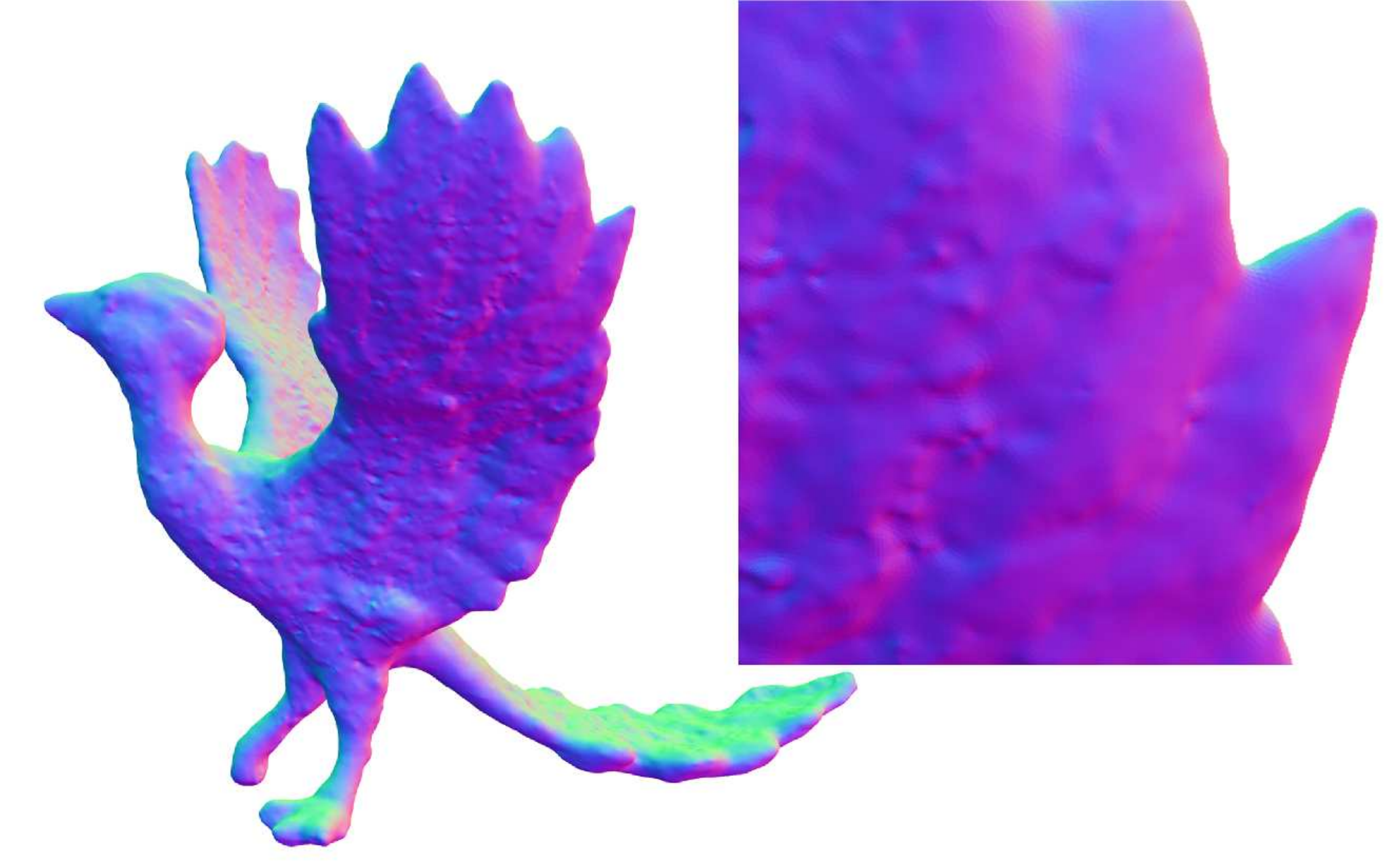}}
\subfigure{
\label{fig:subfig:bird_h}
\includegraphics[height=21mm]{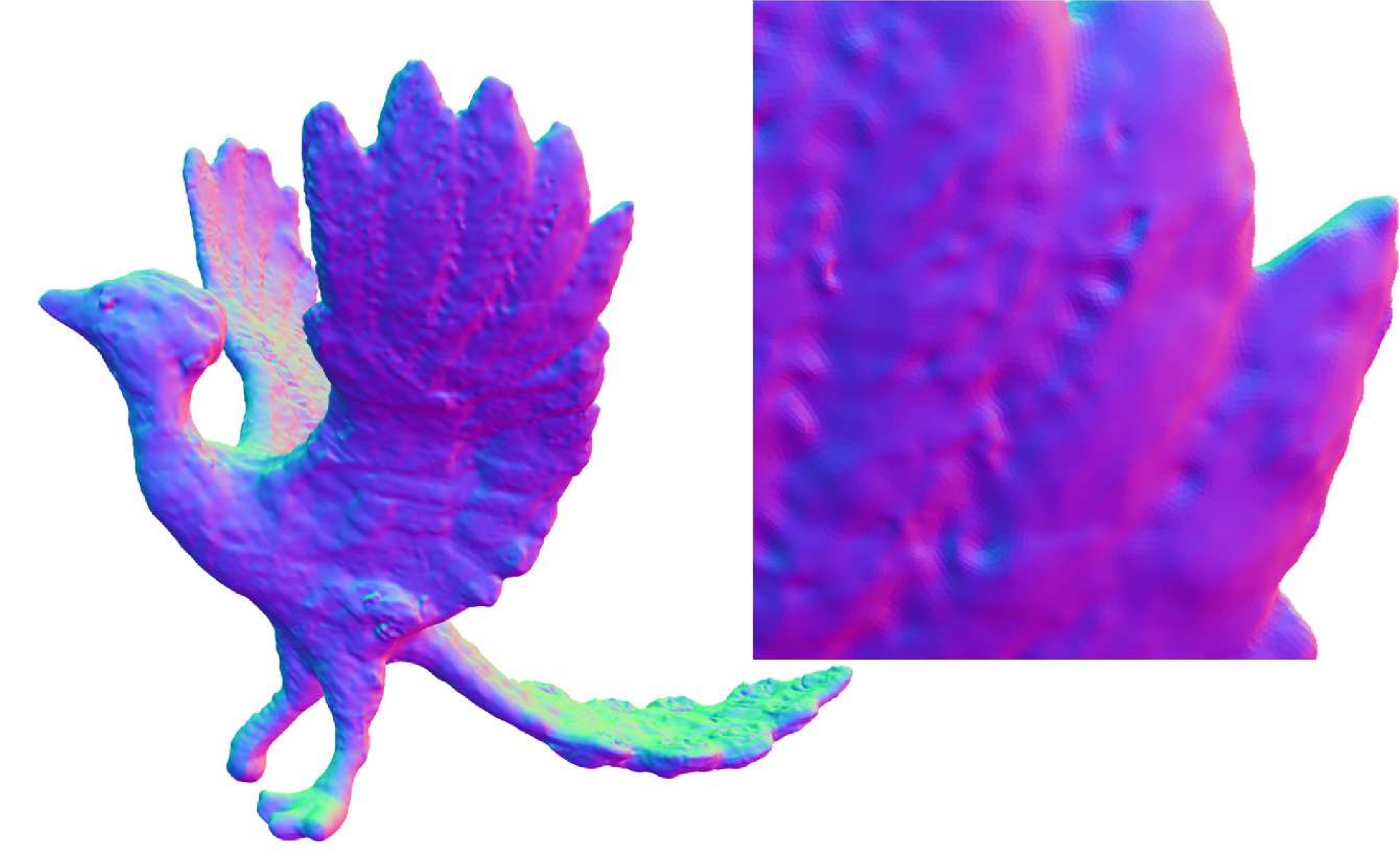}}
 \subfigure{
  \label{fig:subfig:bird_i}
\includegraphics[height=21mm]{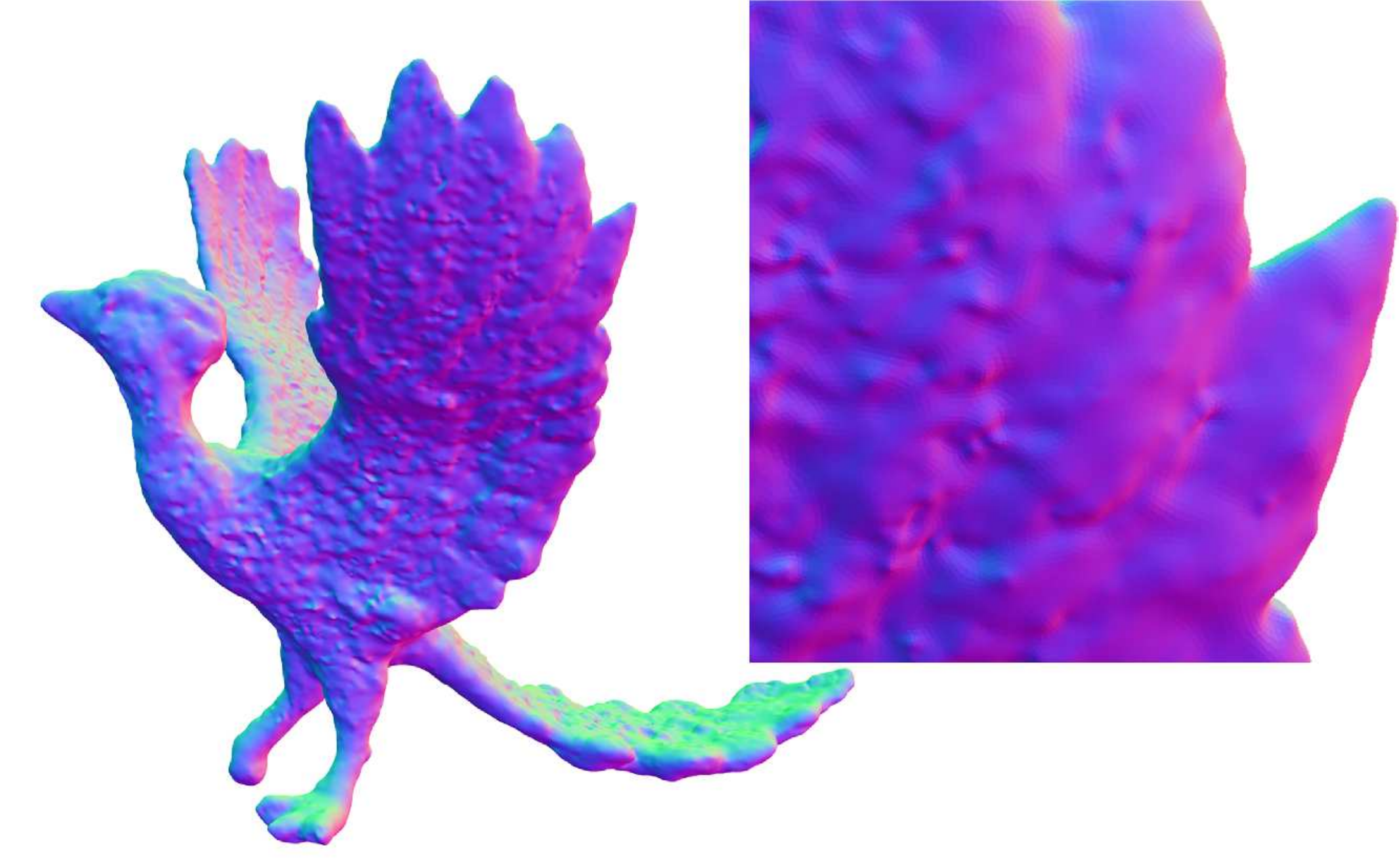}}
 \subfigure{
  \label{fig:subfig:bird_j}
\includegraphics[height=21mm]{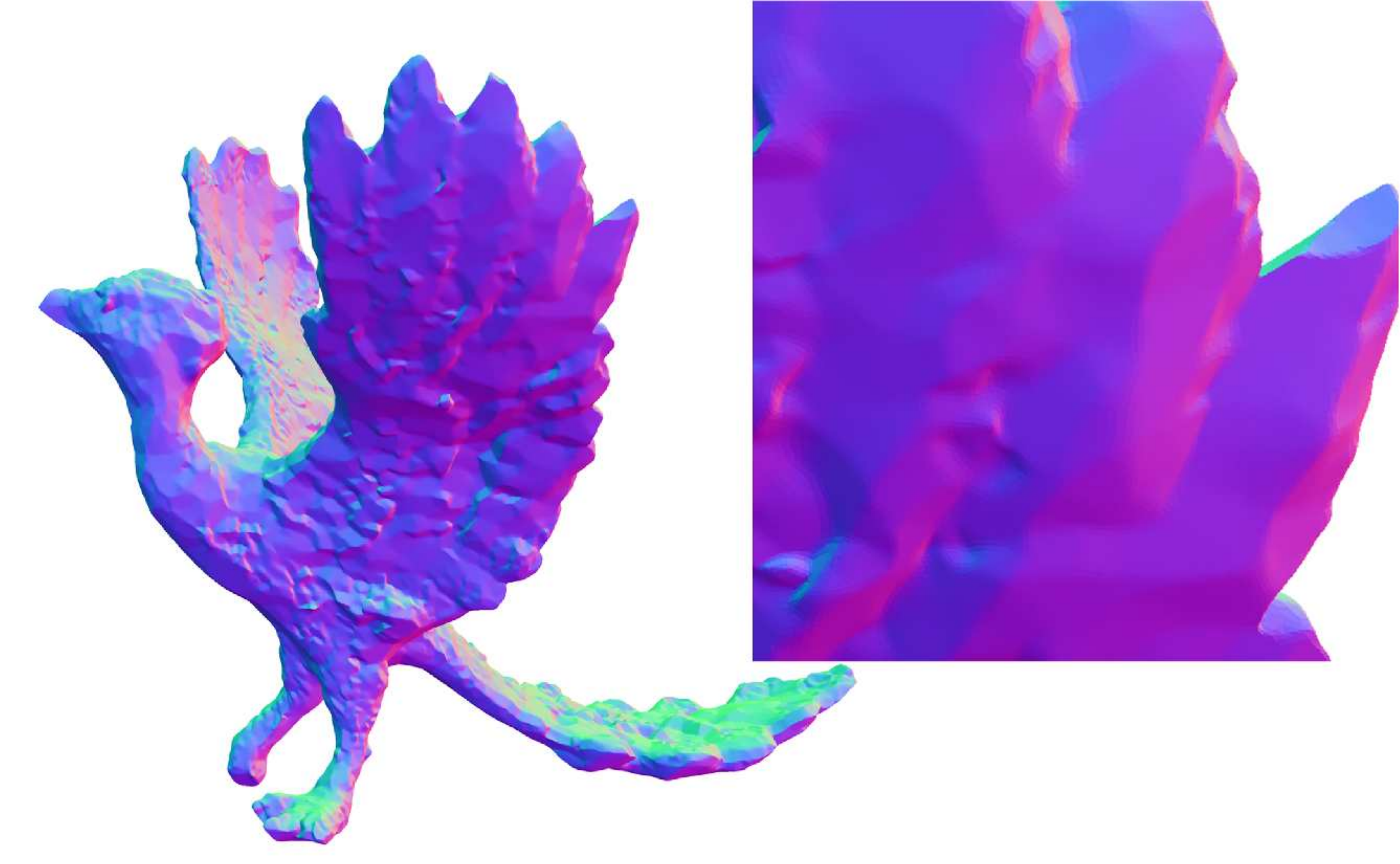}}
\caption{Reconstruction results by state-of-the-art methods and the proposed DCV on the {\it bird} dataset. From left column to right column: input images, results by ~\cite{18}, ~\cite{24}, ~\cite{20} and DCV in two views, respectively.}
\label{fig:bird_result}
\end{figure*}

 The {\it Beethoven} dataset and the {\it bird} dataset contain thirty three $1024\times 768$ images and twenty one $1024\times 768$ images, respectively. They were captured by a set of synchronized cameras. The {\it Beethoven} dataset presents textureless/smooth surface while the {\it bird} dataset presents highly textured surface. The reconstruction results of {\it Beethoven} by several state-of-the-art methods ~\cite{18,20,24} and the proposed DCV are shown in Fig. \ref{fig:beethoven_result}. Thanks to the content-aware $L_{p}$-minimization based denoising algorithm, DCV is able to effectively suppress noise and outliers while keeping the sharp features of surface. The results on the {\it bird} dataset by these methods are shown in Fig. \ref{fig:bird_result}. If the isotropic similarity measure is used, small protrusions such as the wings, details of feathers, claws and head of bird are considered as high-frequency noise and are thus over-smoothed. With the proposed  detail-preserving similarity measure, the fine-scale details are well preserved.

\begin{figure*}[tbp]
\centering
\subfigure[]{
\label{fig:subfig:bell_a}
\includegraphics[height=20mm]{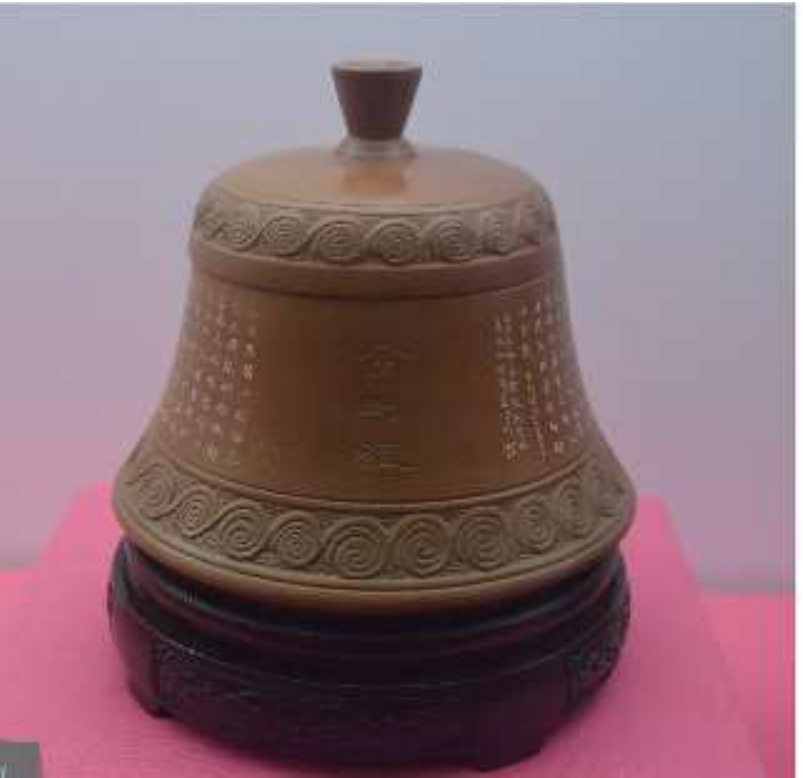}}
 \subfigure[]{
  \label{fig:subfig:bell_b}
\includegraphics[height=22mm]{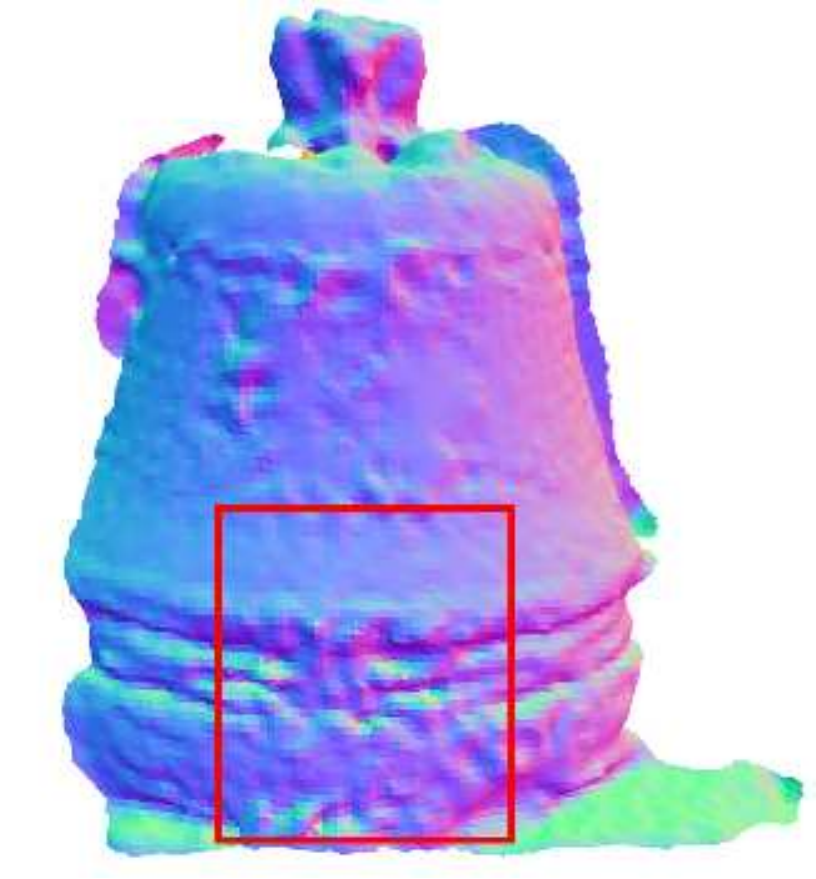}}
 \subfigure[]{
  \label{fig:subfig:bell_c}
\includegraphics[height=22mm]{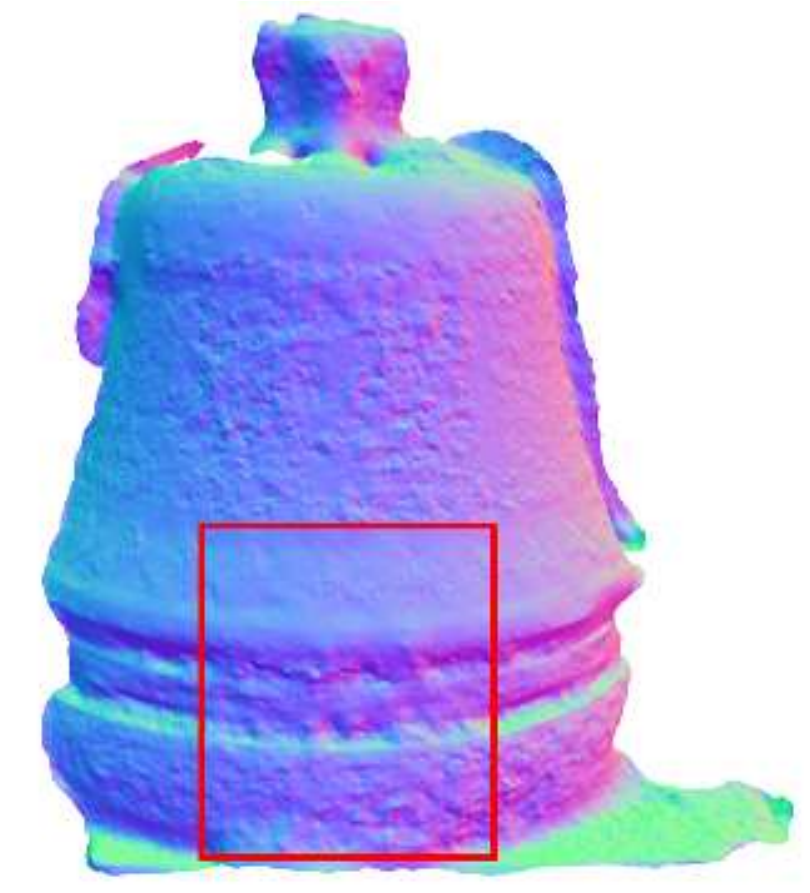}}
\subfigure[]{
\label{fig:subfig:bell_d}
\includegraphics[height=22mm]{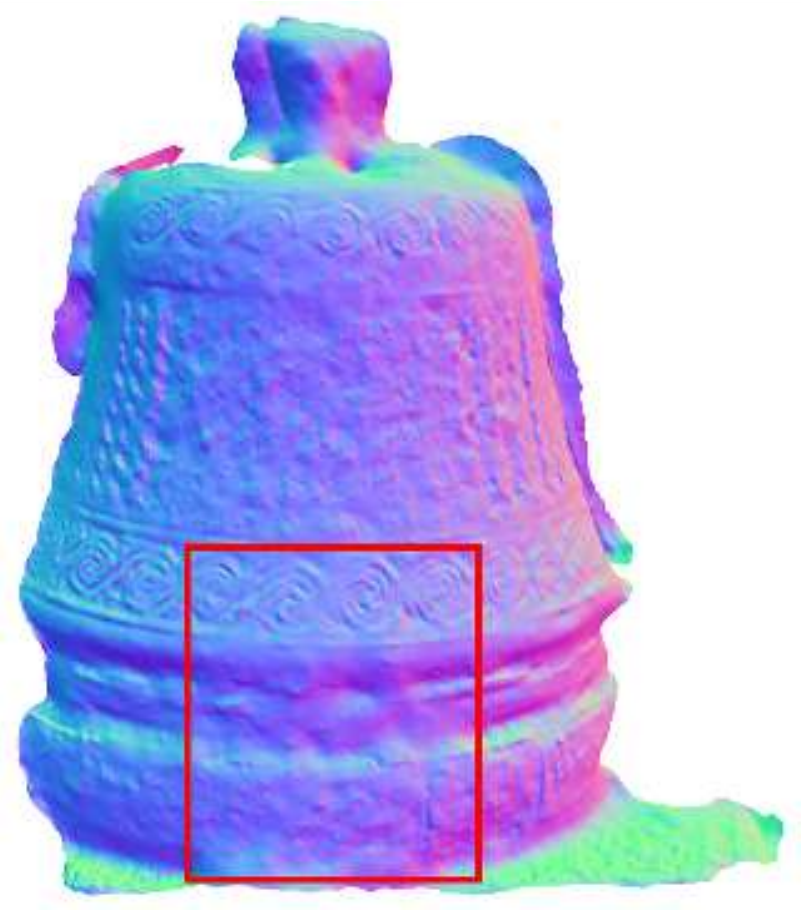}}
\subfigure[]{
\label{fig:subfig:bell_e}
\includegraphics[height=22mm]{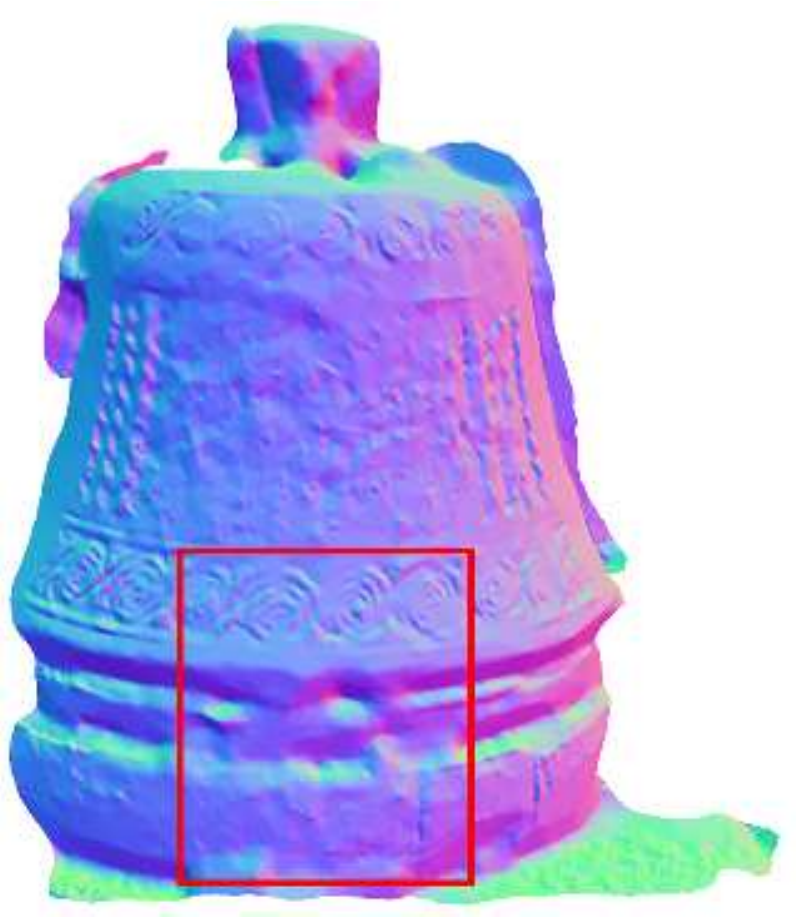}}

\subfigure[]{
 \label{fig:subfig:bell_f}
\includegraphics[height=22mm]{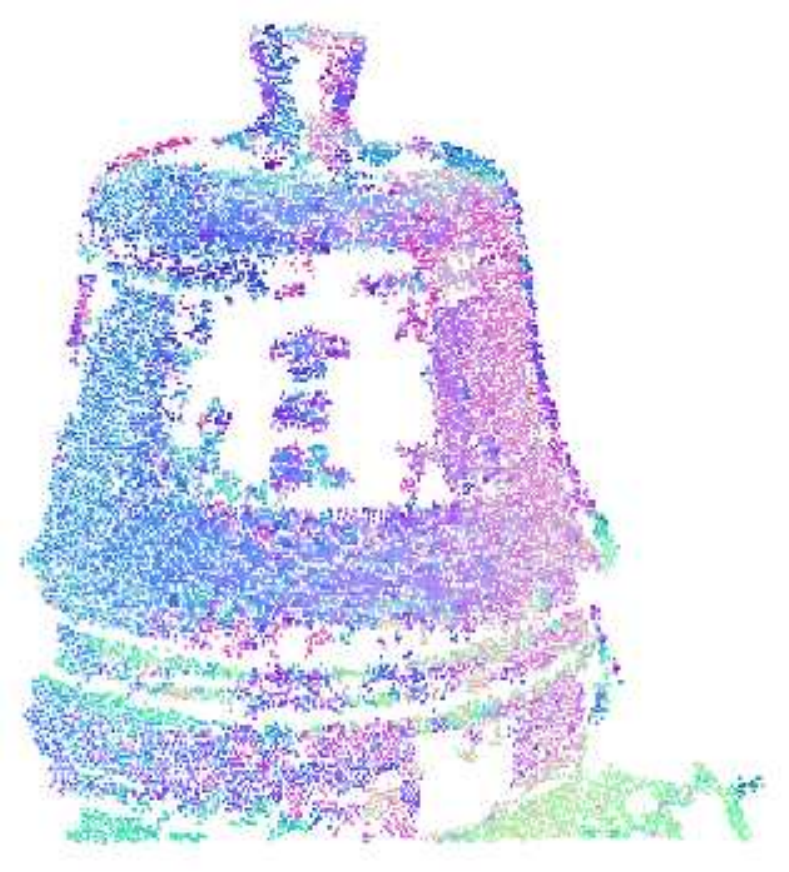}}
 \subfigure[]{
  \label{fig:subfig:bell_g}
\includegraphics[height=22mm]{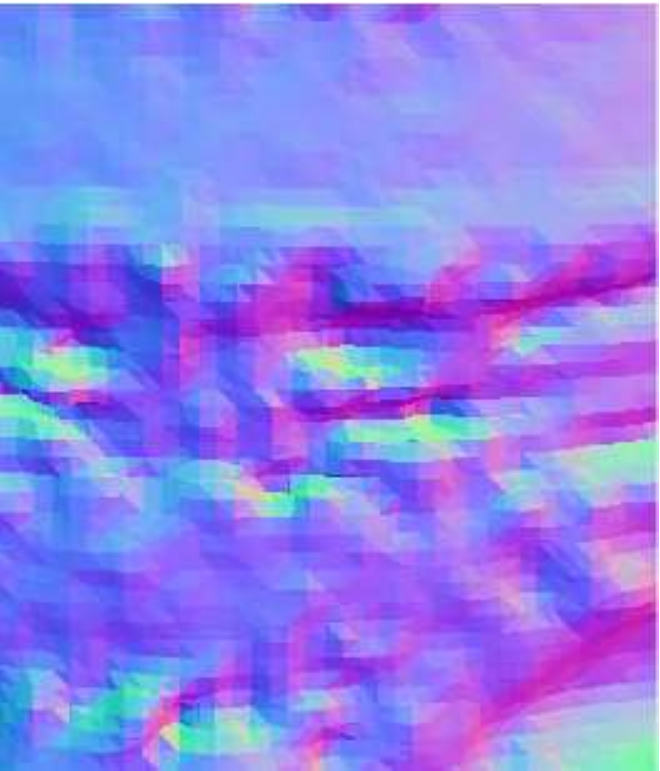}}
\subfigure[]{
\label{fig:subfig:bell_h}
\includegraphics[height=22mm]{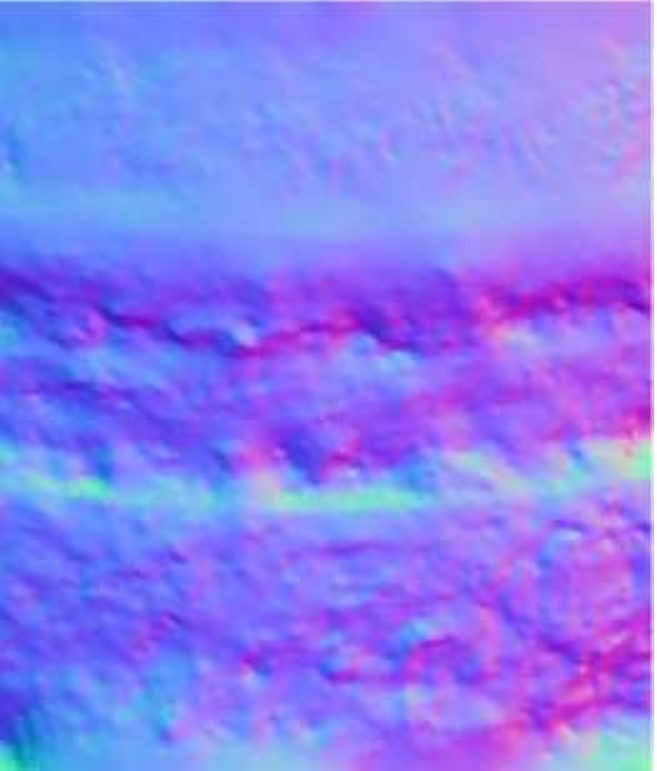}}
 \subfigure[]{
  \label{fig:subfig:bell_i}
\includegraphics[height=22mm]{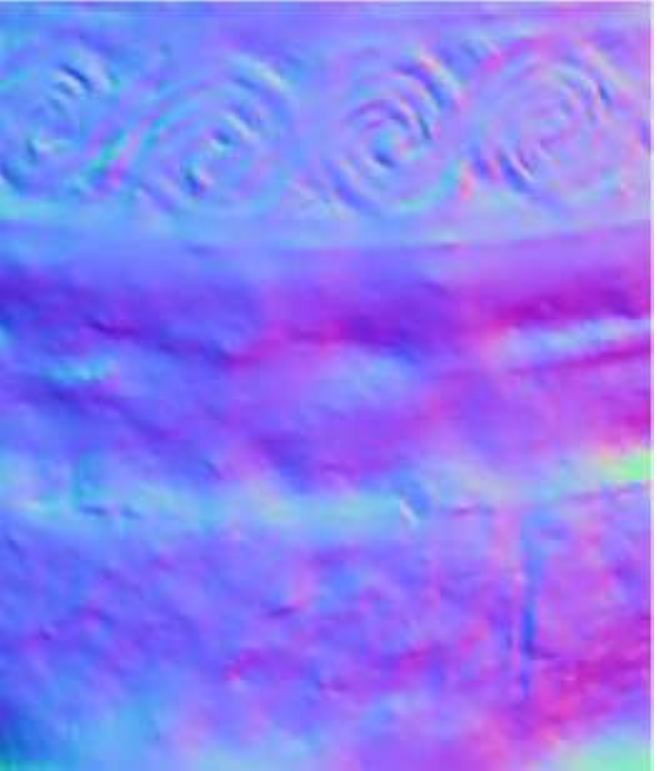}}
 \subfigure[]{
  \label{fig:subfig:bell_j}
\includegraphics[height=22mm]{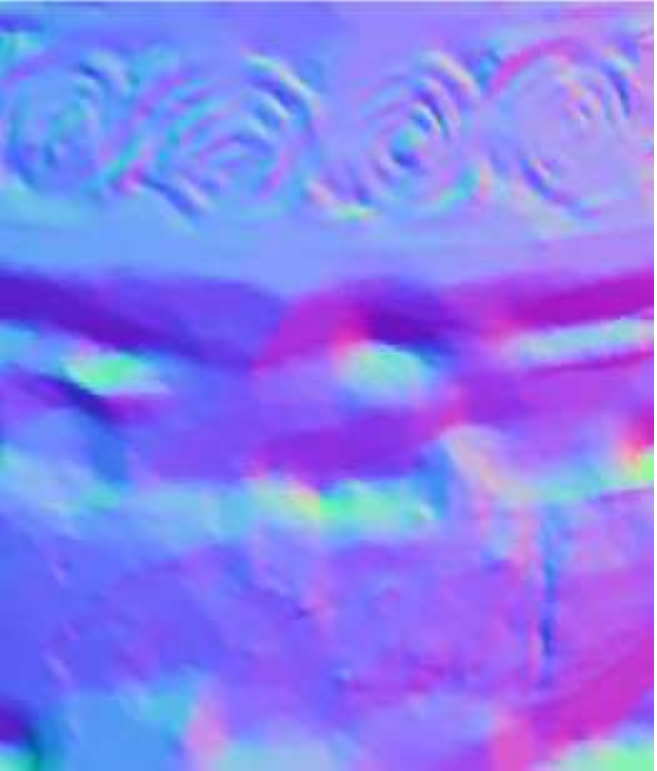}}
\caption{Reconstruction results on the {\it bell} dataset. (a) One of the input images. (b) The initial reconstruction using PMVS+PSR. The point clouds generated by PMVS are shown in (f). (c) Reconstruction result by isotropic similarity measure + isotropic smoothing. (d) Reconstruction result by detail-preserving similarity measure + isotropic smoothing. (e) Reconstruction result by DCV. (g)-(j) are the close-up images corresponding to the red rectangle regions in (b)-(j), respectively. One can see that DCV preserves well the fine details and smooth surface.}
\label{fig:bell_result}
\end{figure*}

The {\it Buddha} dataset consists of five $2400\times 1800$ images, and it has been used to validate the proposed detail-preserving similarity measure in Fig. \ref{fig:detailenhancefig}. Fig. \ref{fig:subfig:detailenhancefigd} is one sample image from the dataset. Fig. \ref{fig:subfig:detailenhancefige} shows reconstruction results by using the isotropic ZNCC similarity measure,  and Fig. \ref{fig:subfig:detailenhancefigf} shows the results by using the proposed detail-preserving similarity measure. Note that the preserved features on the finger (left up close-up image) and wrinkles on the clothes (right down close-up image) of the Buddha statue. The results on  the {\it bell} dataset are shown in Fig.~\ref{fig:bell_result}. The {\it bell} dataset consists of only three $1504\times 1004$ images of a bell in a museum, and thus only a partial surface of bell is reconstructed. The less number of observed images makes the regularization scheme more important. One of observed images is shown in Fig.~\ref{fig:subfig:bell_a}. The initial surface of the bell is estimated by PMVS+PSR. The point clouds generated by PMVS and watertight surface generated by PSR are shown in Fig.~\ref{fig:subfig:bell_f} and Fig.~\ref{fig:subfig:bell_b}, respectively. Fig.~\ref{fig:subfig:bell_c}, Fig.~\ref{fig:subfig:bell_d} and Fig.~\ref{fig:subfig:bell_e} show the reconstruction results of "isotropic similarity+isotropic regularization", "detail-preserving  similarity+isotropic regularization", and DCV, respectively. The choosed isotropic regularization combines the first order and second order Laplacian \cite{31a}. Fig.~\ref{fig:subfig:bell_h} and Fig.~\ref{fig:subfig:bell_j} show the close-up images of the corresponding results in Fig.~\ref{fig:subfig:bell_b} and Fig.~\ref{fig:subfig:bell_e}. It is clear that DCV presents the best results for both fine details and surface smoothness among all competing methods.

\begin{figure}[tbp]
\centering
\subfigure[]{
\label{fig:subfig:resultfountaina}
\includegraphics[height=25mm]{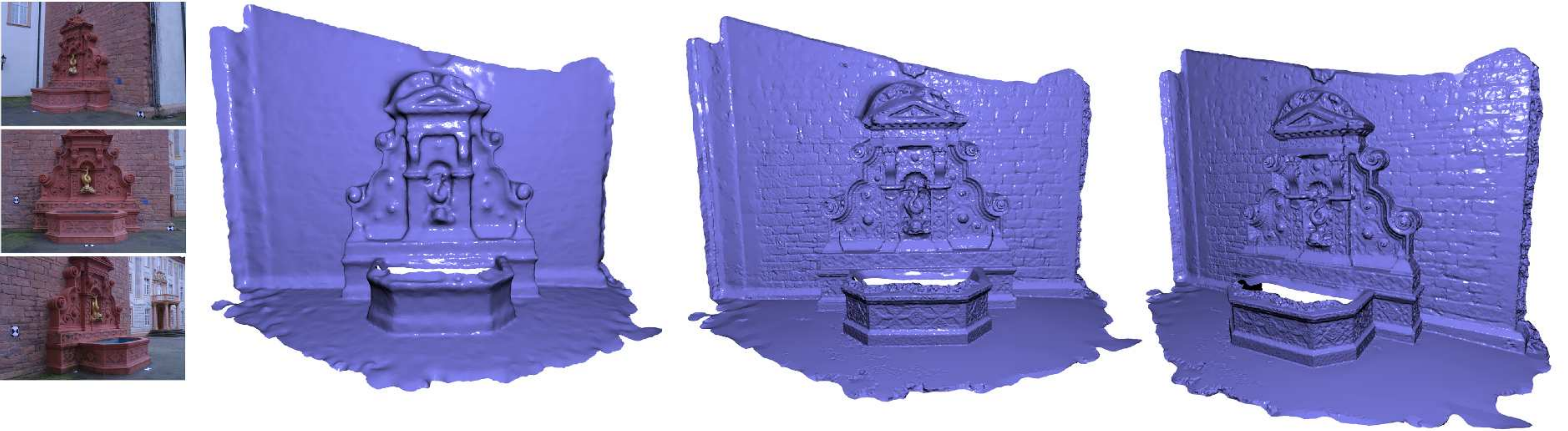}}

 \subfigure[]{
  \label{fig:subfig:resultfountainb}
\includegraphics[height=33mm]{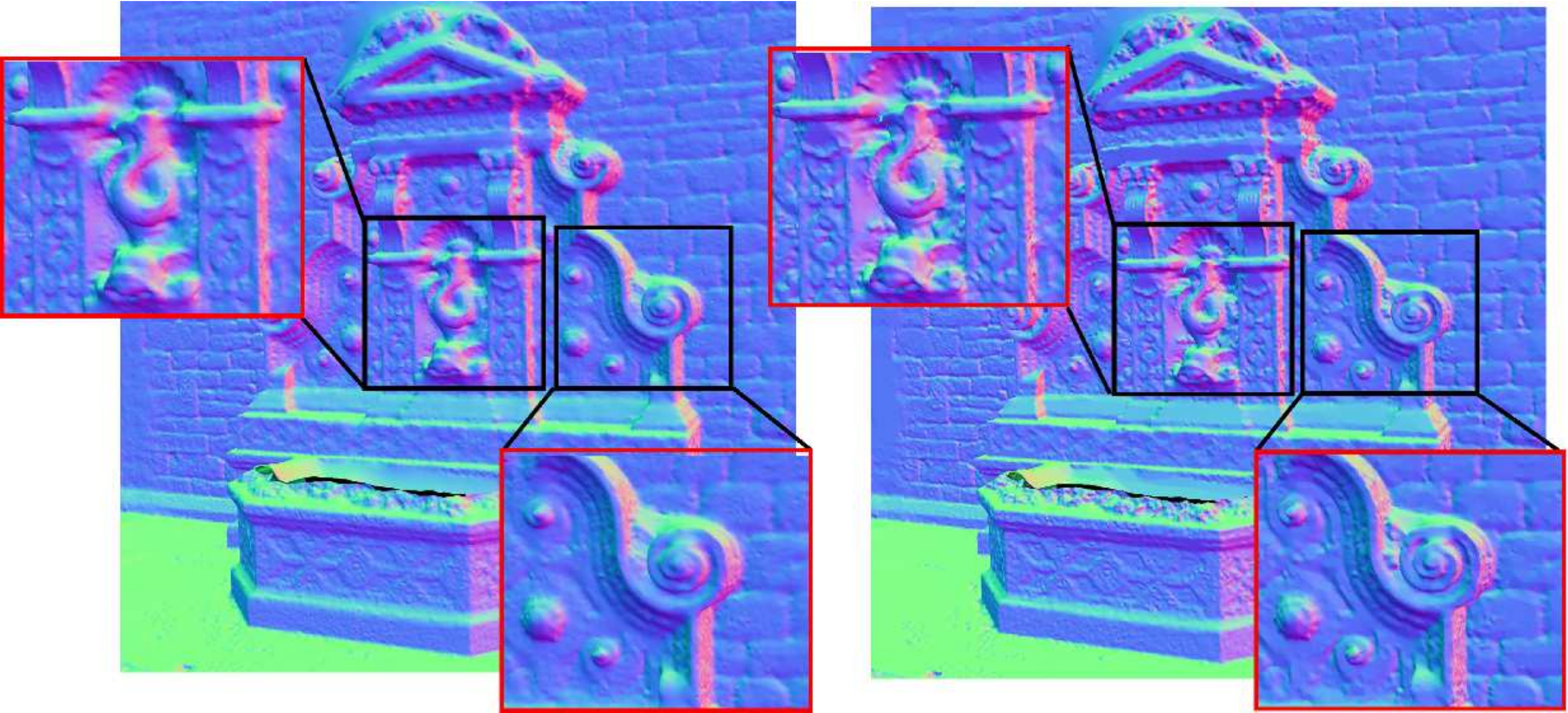}}
\vspace{-2mm}
\caption{(a) Results of DCV on the {\it fountain-P11} dataset. From left to right: several input images, initial surface, results obtained by DCV. (b) The left image shows the result obtained by the method based on isotropic similarity measure and surface regularization, and the right image shows the results obtained by DCV. Obviously, DCV performs better in preserving the small-scale details and sharp features. }
\label{fig:resultfountain}
\vspace{-5mm}
\end{figure}

The {\it fountain-P11} is an outdoor dataset including eleven $3072\times 2048$ images. The results on {\it fountain-P11} dataset are shown in Fig.~\ref{fig:resultfountain}. The input images, initial surface generated by PMVS+PSR and reconstruction results by DCV are shown in Fig.~\ref{fig:subfig:resultfountaina}. The comparison results of DCV with the isotropic method are shown in Fig.~\ref{fig:subfig:resultfountainb}. It can be easily seen that DCV performs better than the isotropic method on preserving the fine-scale details and sharp features.

The {\it statuegirl} is an outdoor dataset including fifty $2592\times 3888$ images. The input images, initial surface generated by PMVS+PSR, and the reconstruction results by DCV and commercial 3D reconstruction software Smart3Dcapture (free edition)~\cite{acute3d} are shown in the first row, respectively. The close-ups images in different surface regions are shown in the second and the third rows.  The images have been down-sampled by half before performing the reconstructions. For the Smart3Dcapture software, a complete and robust reconstruction pipeline has been integrated, including camera calibration, dense reconstruction and visualization. We have used the software's ultra high precision option to recover more details. For DCV, bundler is used for calibration and PMVS+PSR is used for initialization.  The comparison results show that DCV can generally obtain similar results to Smart3Dcapture, and in some part (e.g., toes) it can recover more fine-scale details.

\begin{figure*}[tbp]
\centering
\subfigure{
\label{fig:subfig:resultstatuegirla}
\includegraphics[height=35mm]{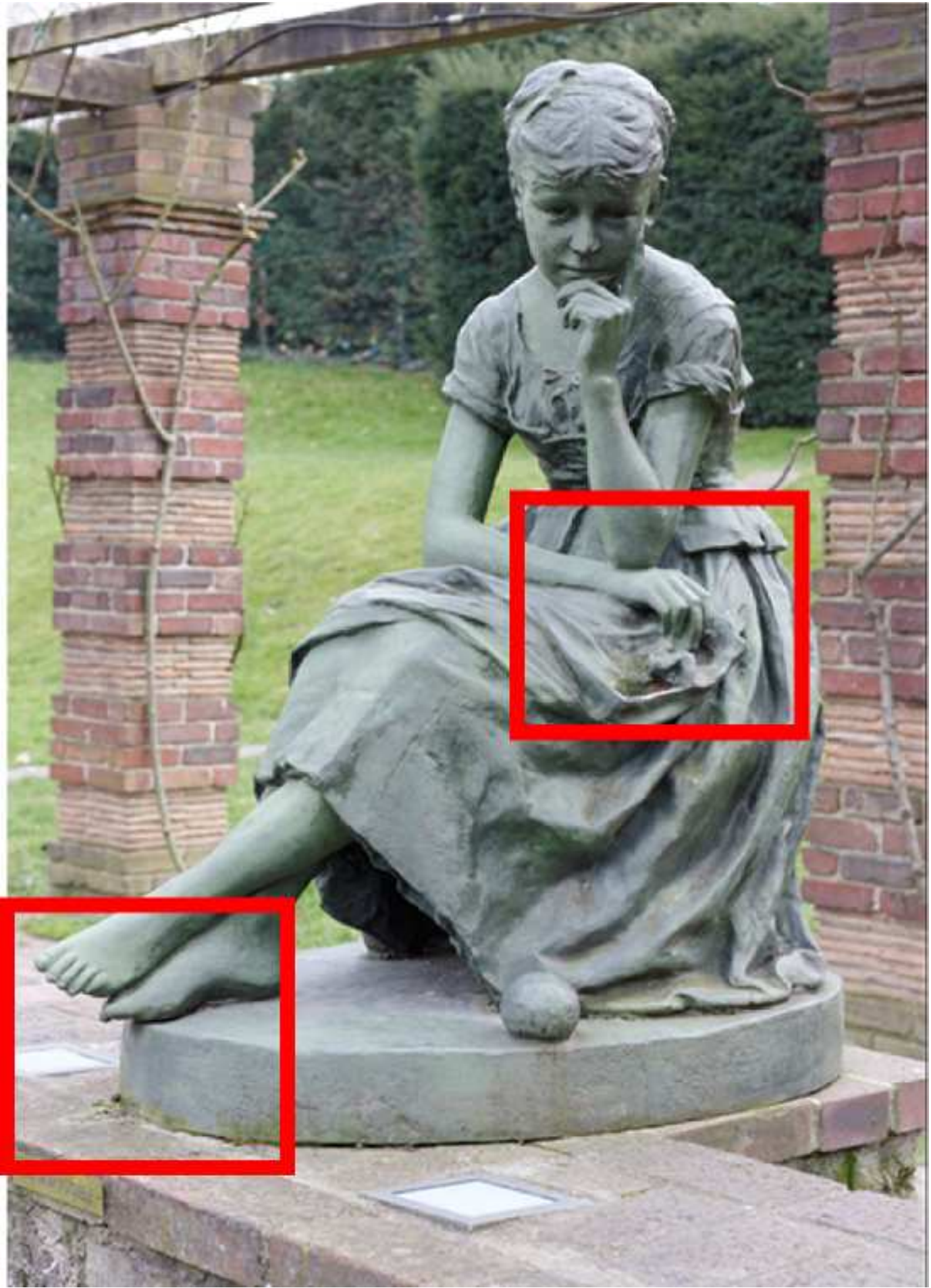}}
\subfigure{
\label{fig:subfig:resultstatuegirlb}
\includegraphics[height=35mm]{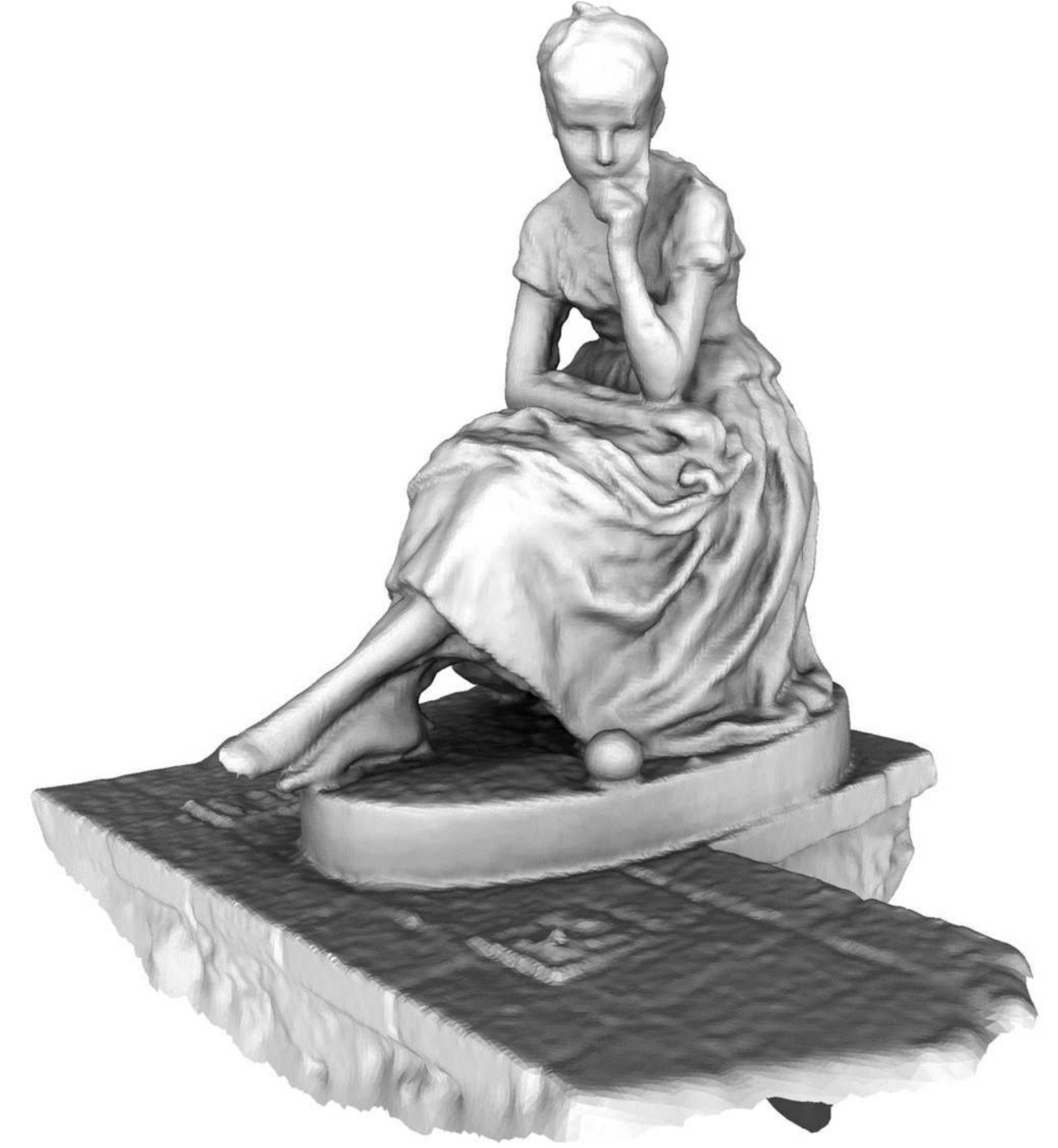}}
\subfigure{
\label{fig:subfig:resultstatuegirlc}
\includegraphics[height=35mm]{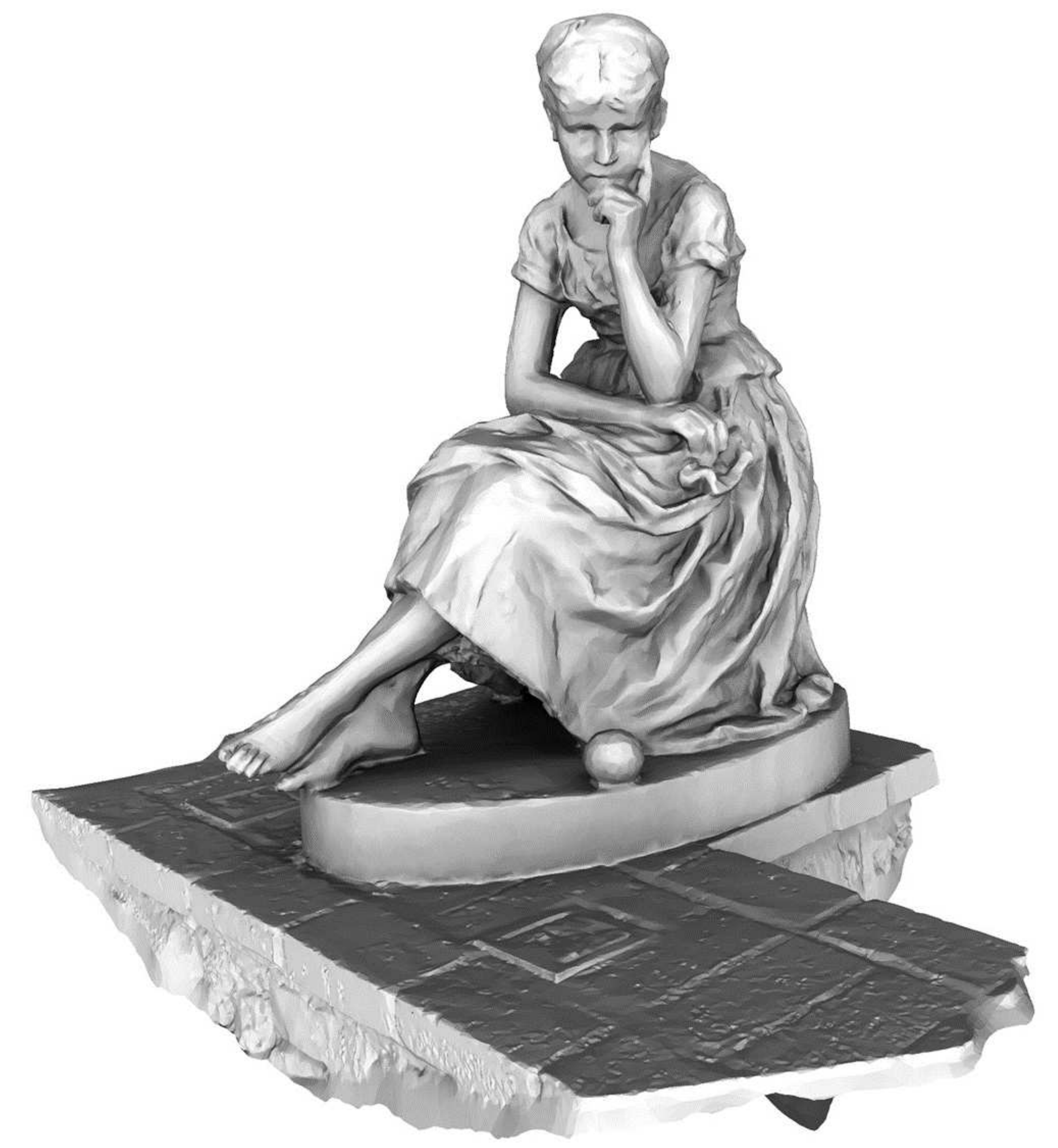}}
 \subfigure{
  \label{fig:subfig:resultstatuegirld}
\includegraphics[height=35mm]{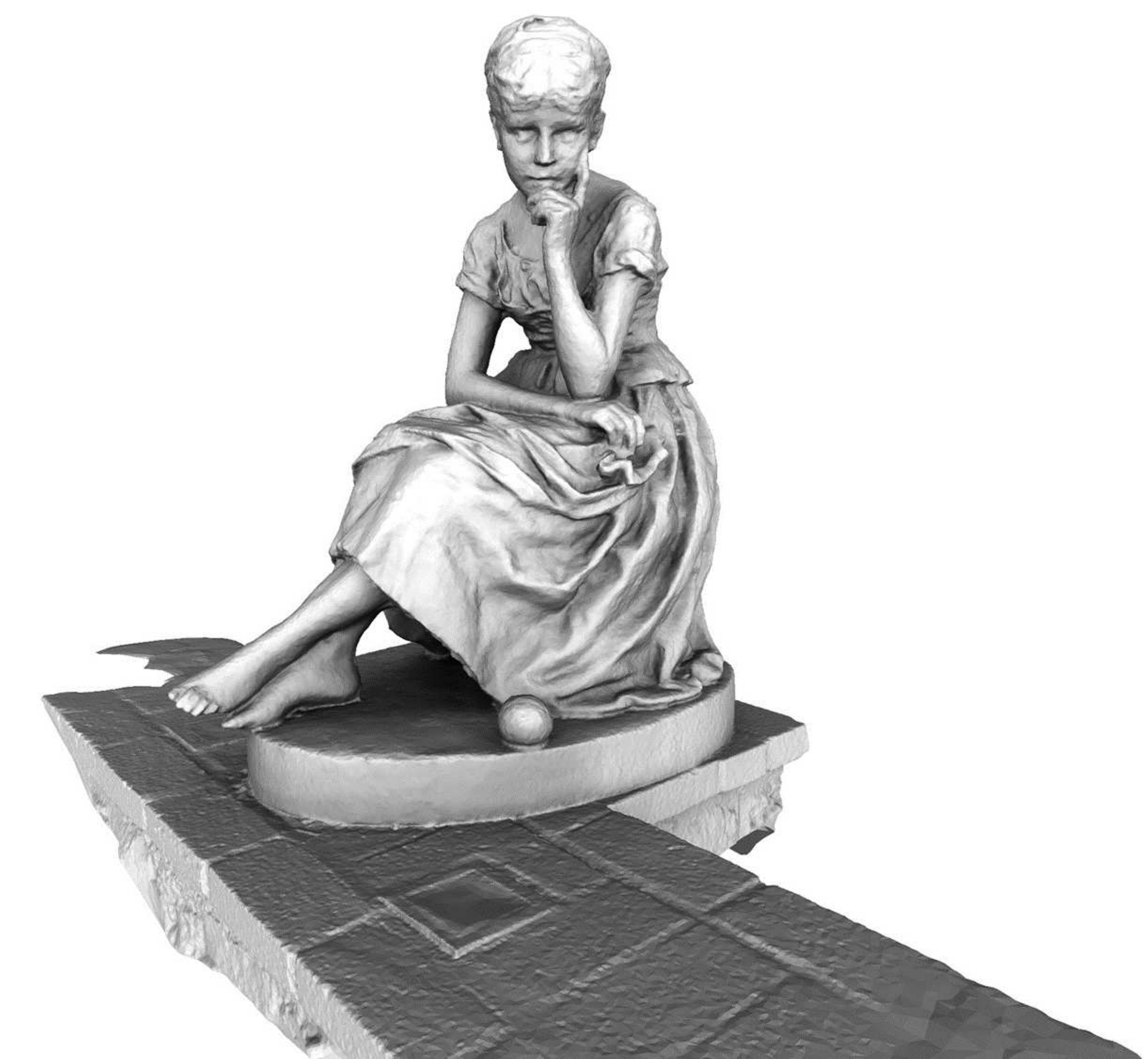}}

 \subfigure{
  \label{fig:subfig:resultstatuegirle}
\includegraphics[height=25mm]{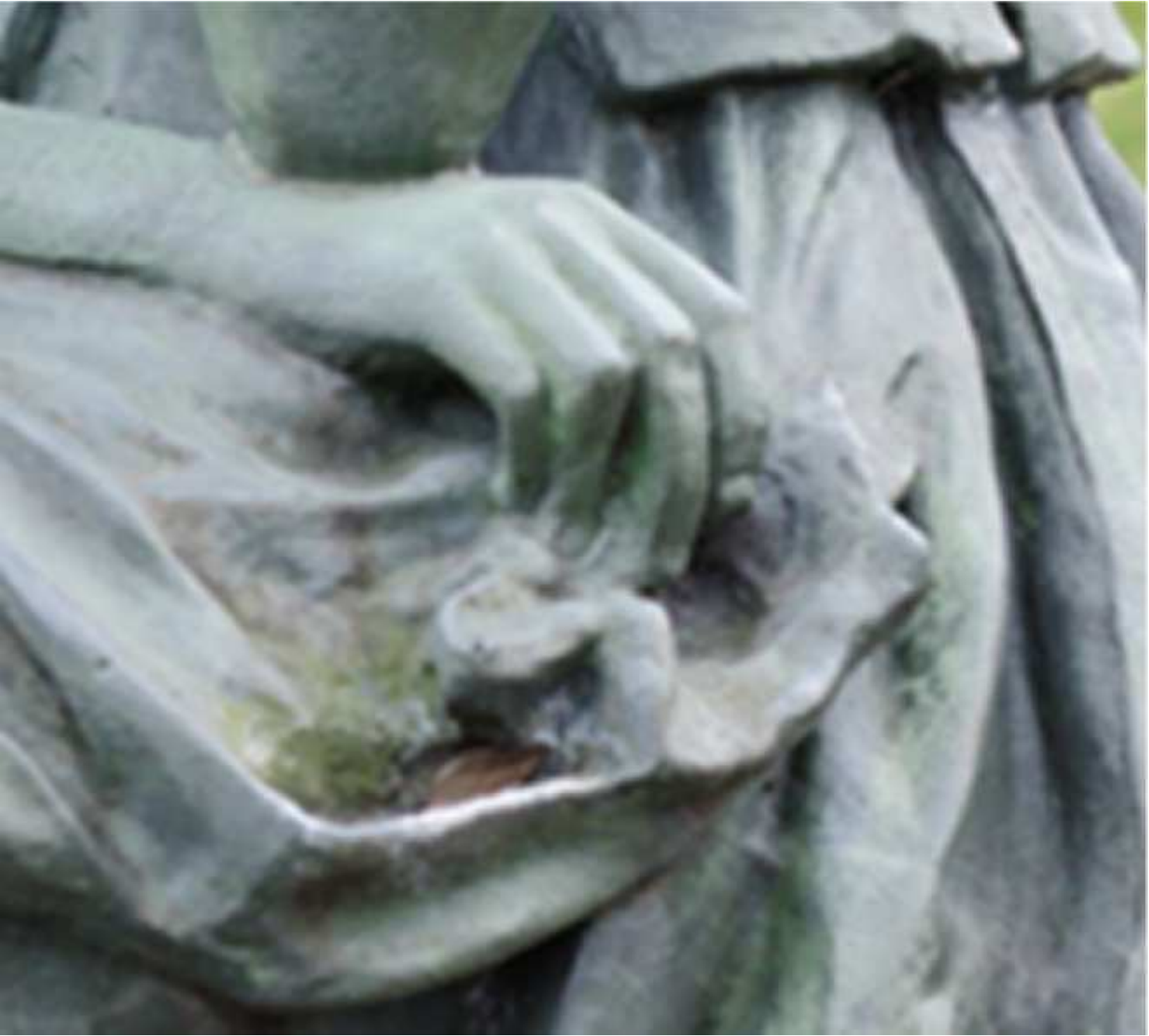}}
 \subfigure{
  \label{fig:subfig:resultstatuegirlf}
\includegraphics[height=25mm]{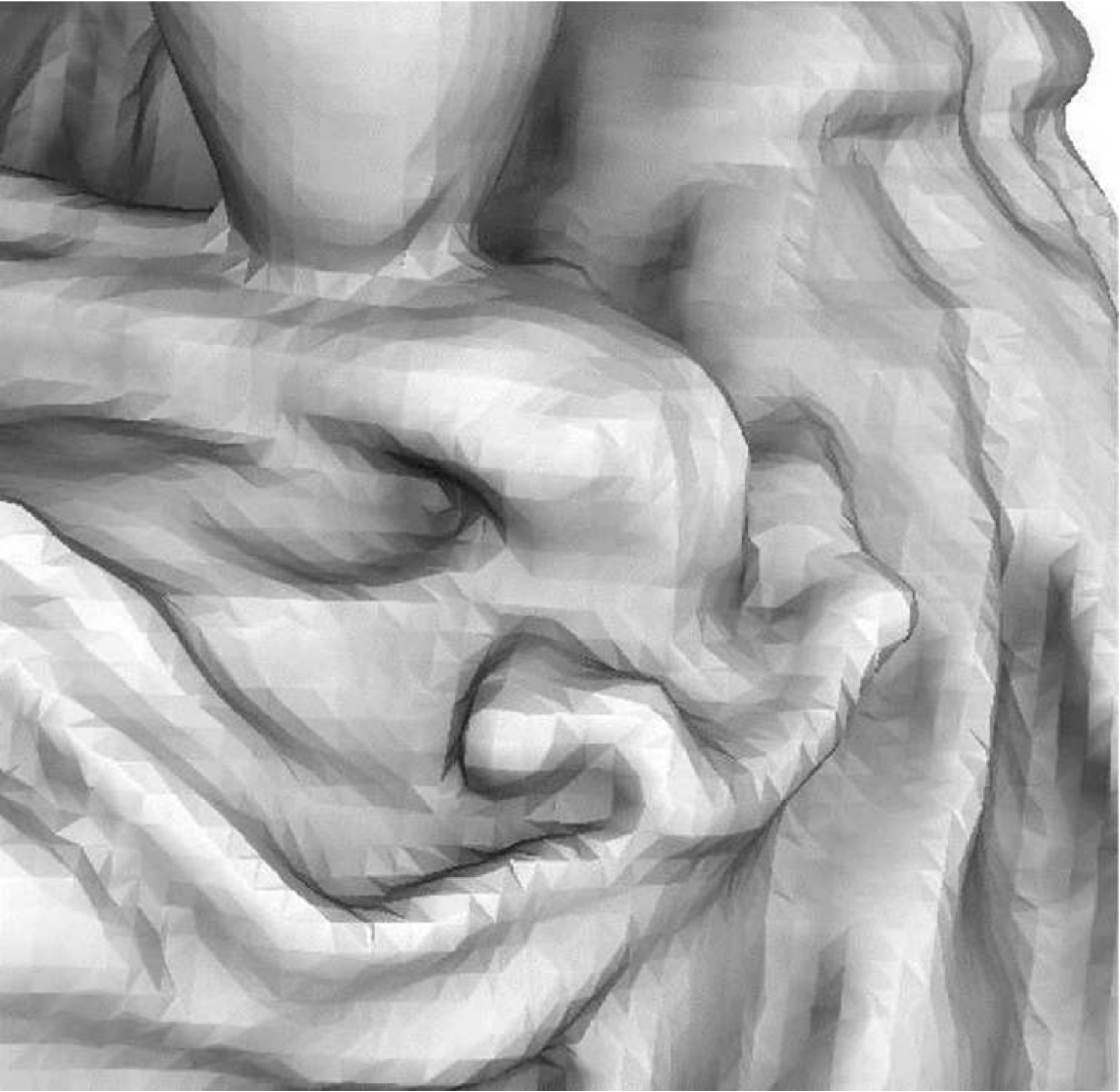}}
 \subfigure{
  \label{fig:subfig:resultstatuegirlg}
\includegraphics[height=25mm]{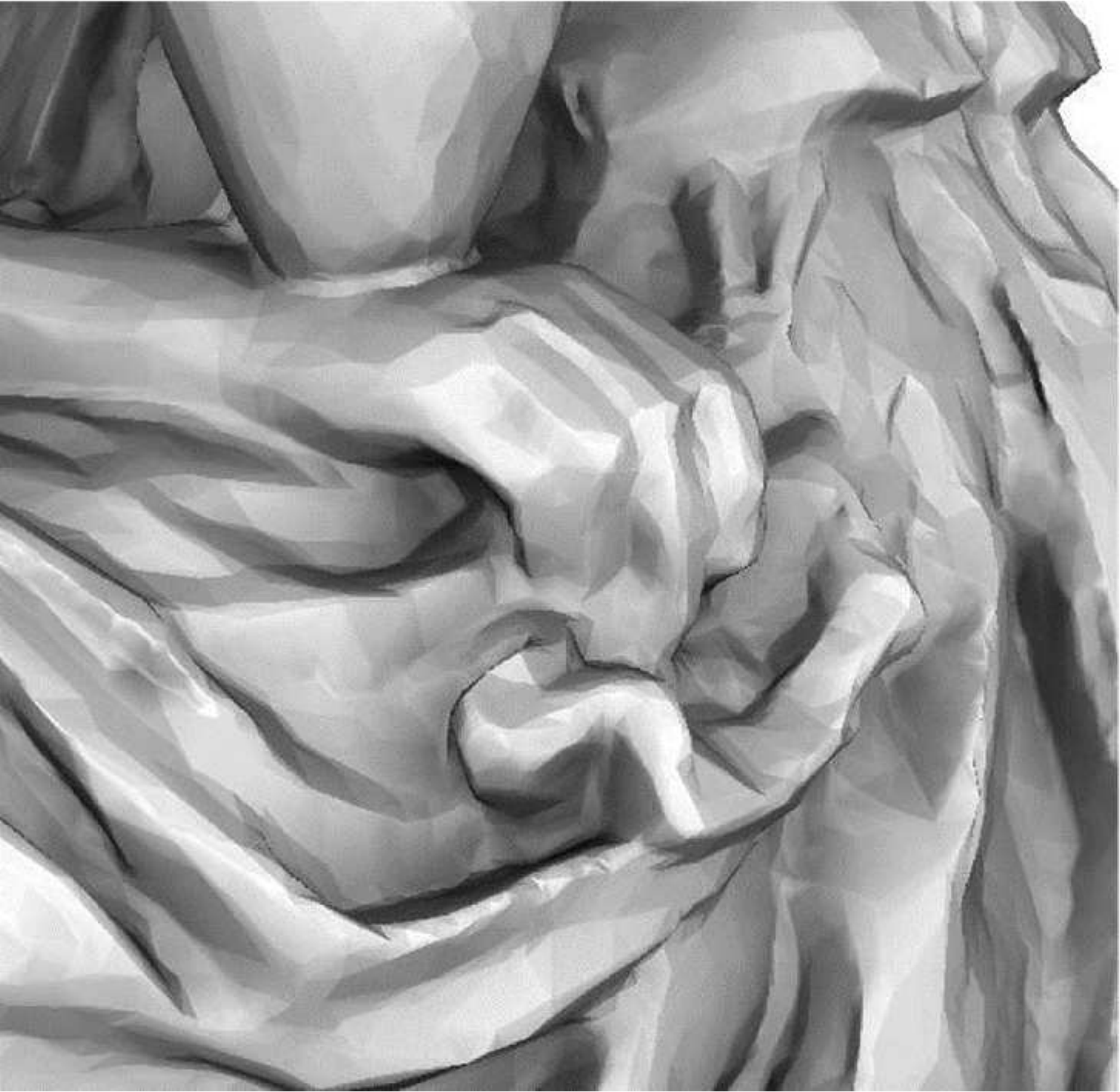}}
 \subfigure{
  \label{fig:subfig:resultstatuegirlh}
\includegraphics[height=25mm]{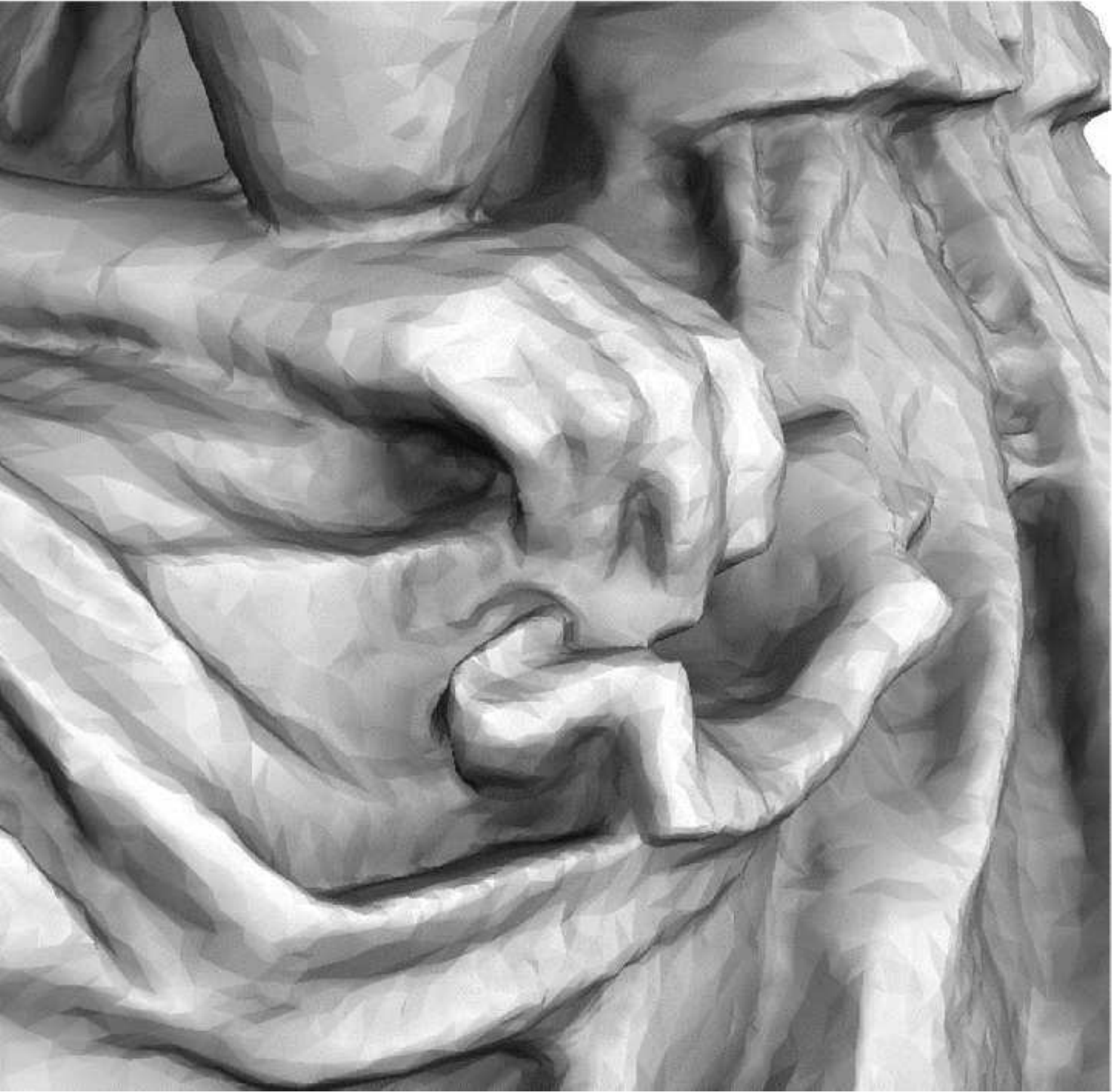}}

 \subfigure{
  \label{fig:subfig:resultstatuegirli}
\includegraphics[height=25mm]{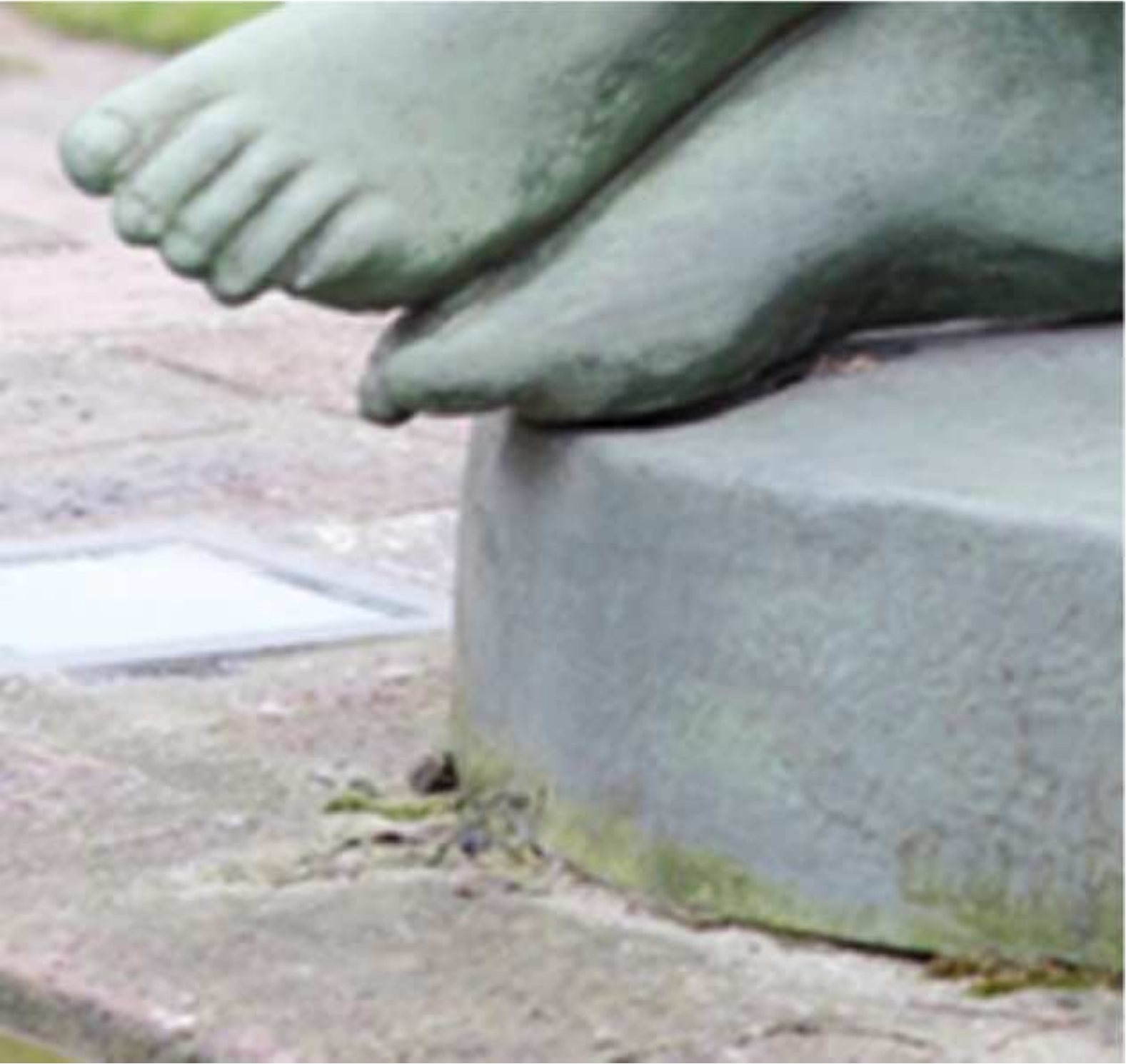}}
 \subfigure{
  \label{fig:subfig:resultstatuegirlj}
\includegraphics[height=25mm]{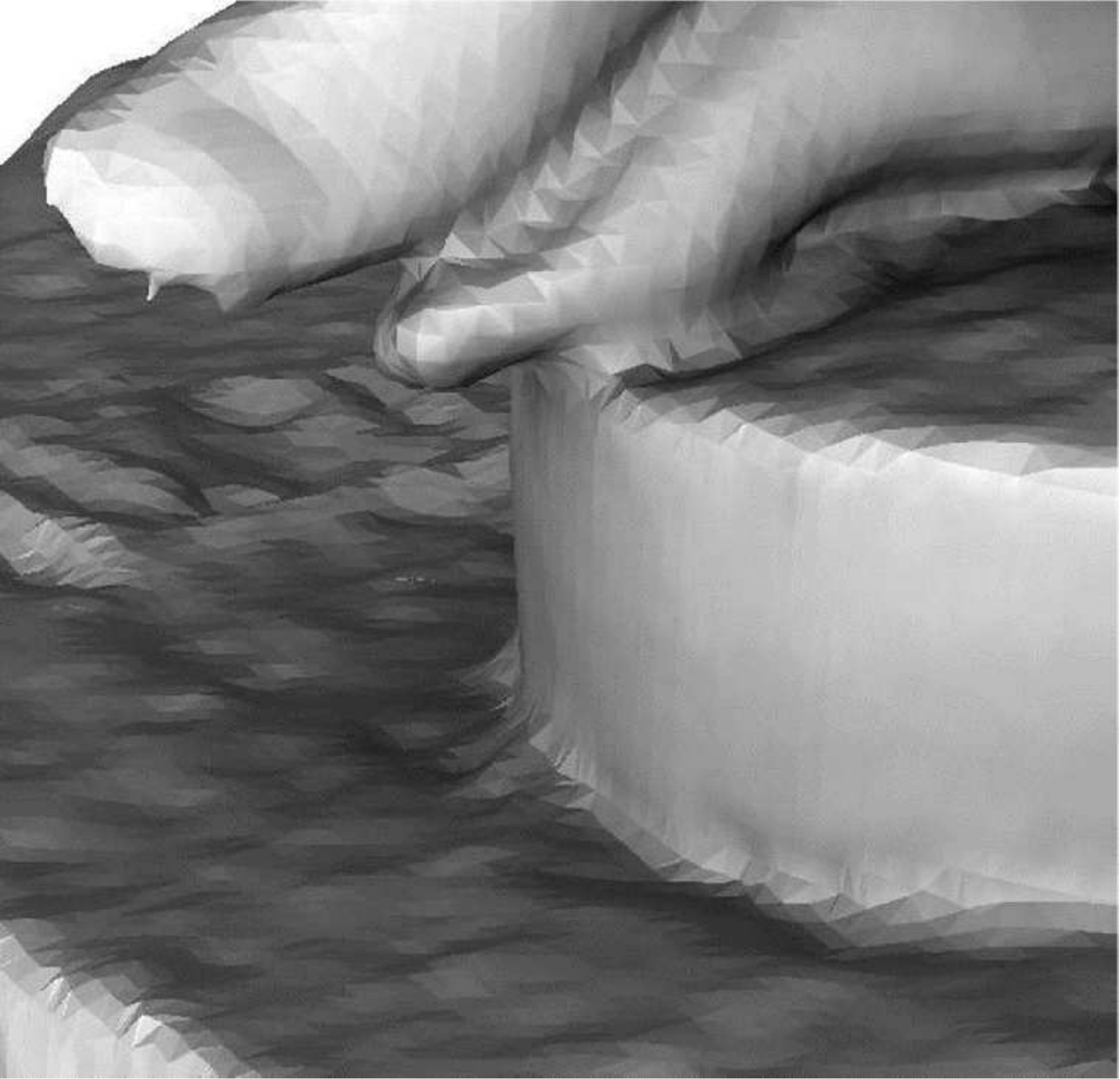}}
 \subfigure{
  \label{fig:subfig:resultstatuegirlk}
\includegraphics[height=25mm]{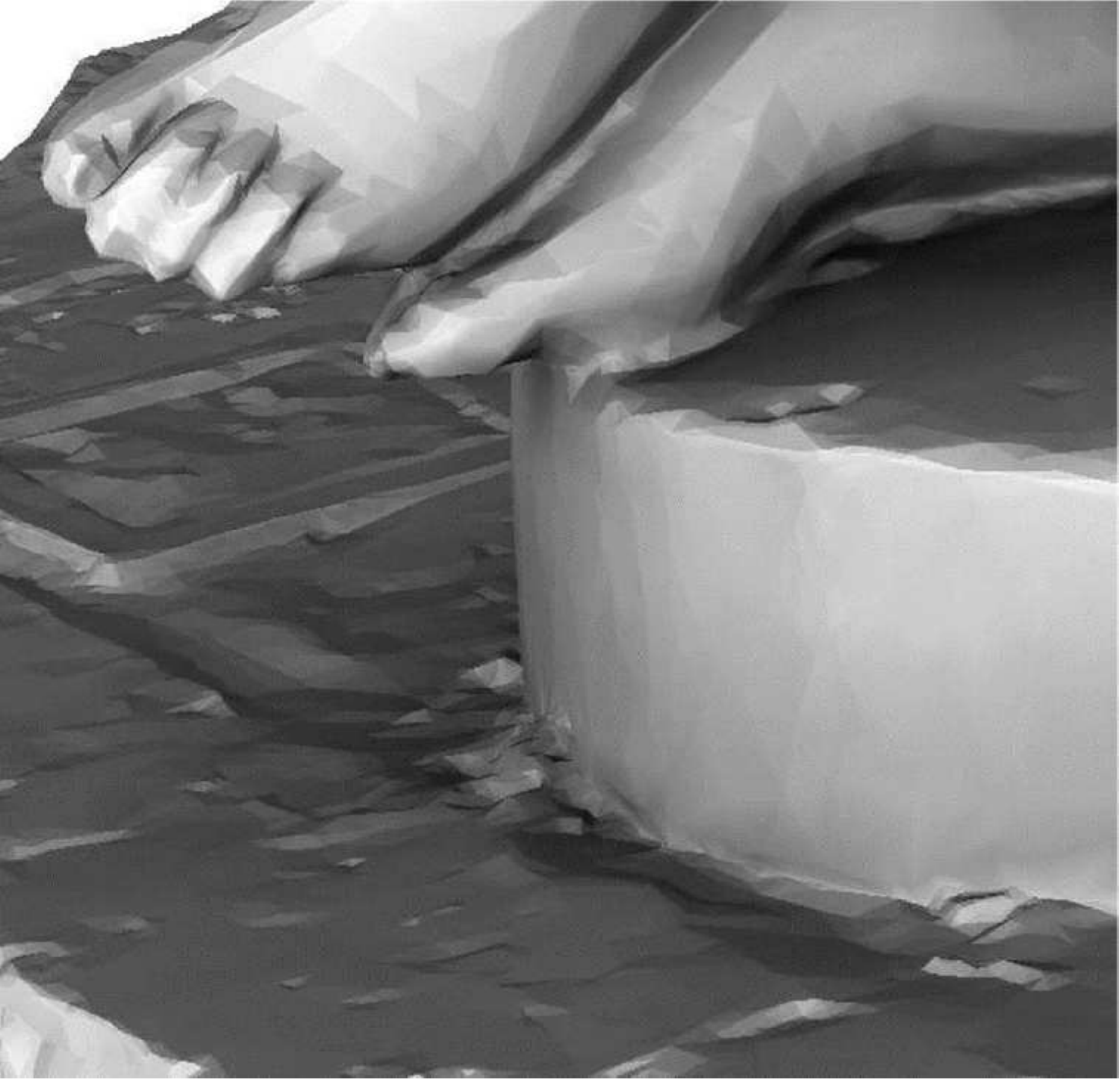}}
 \subfigure{
  \label{fig:subfig:resultstatuegirll}
\includegraphics[height=25mm]{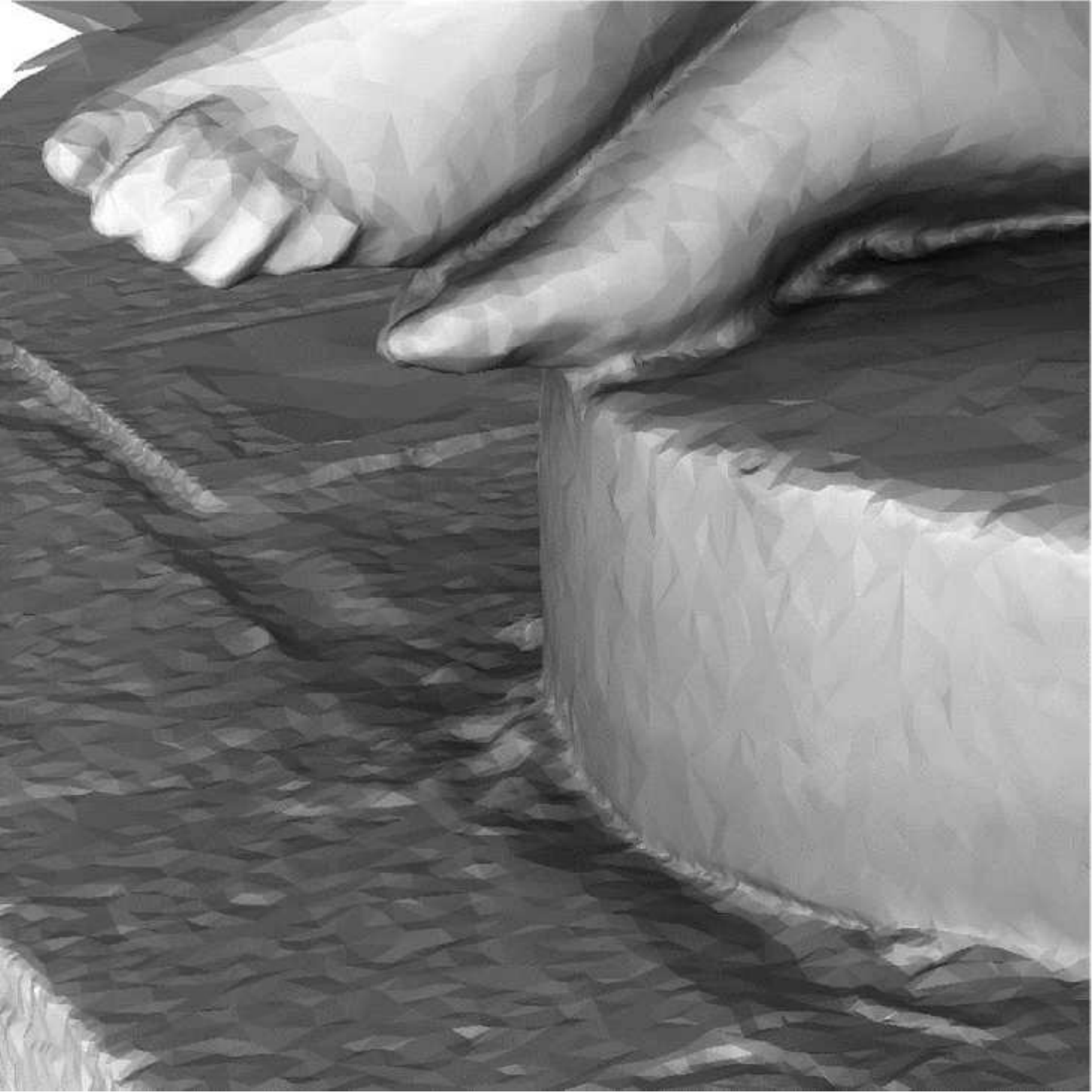}}

\vspace{-2mm}
\caption{Results on the {\it statuegirl} dataset. First row, from left to right: one of input images,  initial surface by PMVS+PSR, results by DCV and results by Smart3dCapture Free Edition with ultra high precision setting. Second and third row, from left to right: the close-ups of reconstruction results in the first row.}
\label{fig:resultstatuegirl}
\vspace{-5mm}
\end{figure*}

\begin{figure*}[tbp]
\centering
\subfigure[]{
\label{fig:subfig:resultbunnya}
\includegraphics[height=20mm]{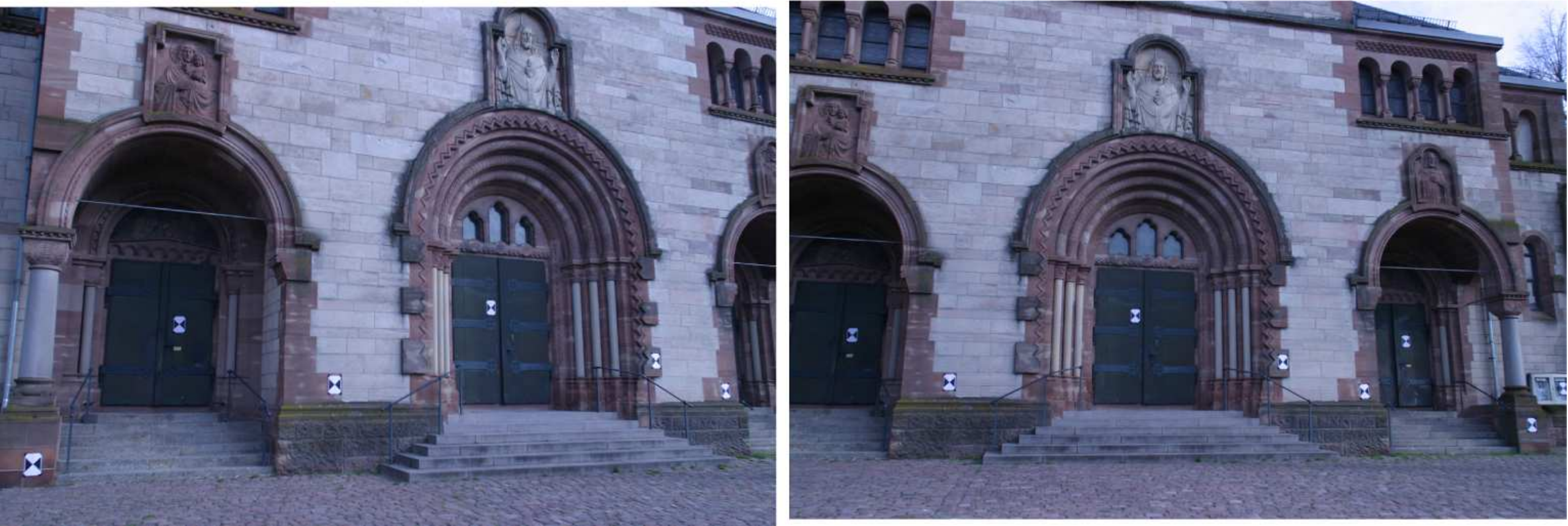}}
 \subfigure[]{
  \label{fig:subfig:resultbunnyb}
\includegraphics[height=25mm]{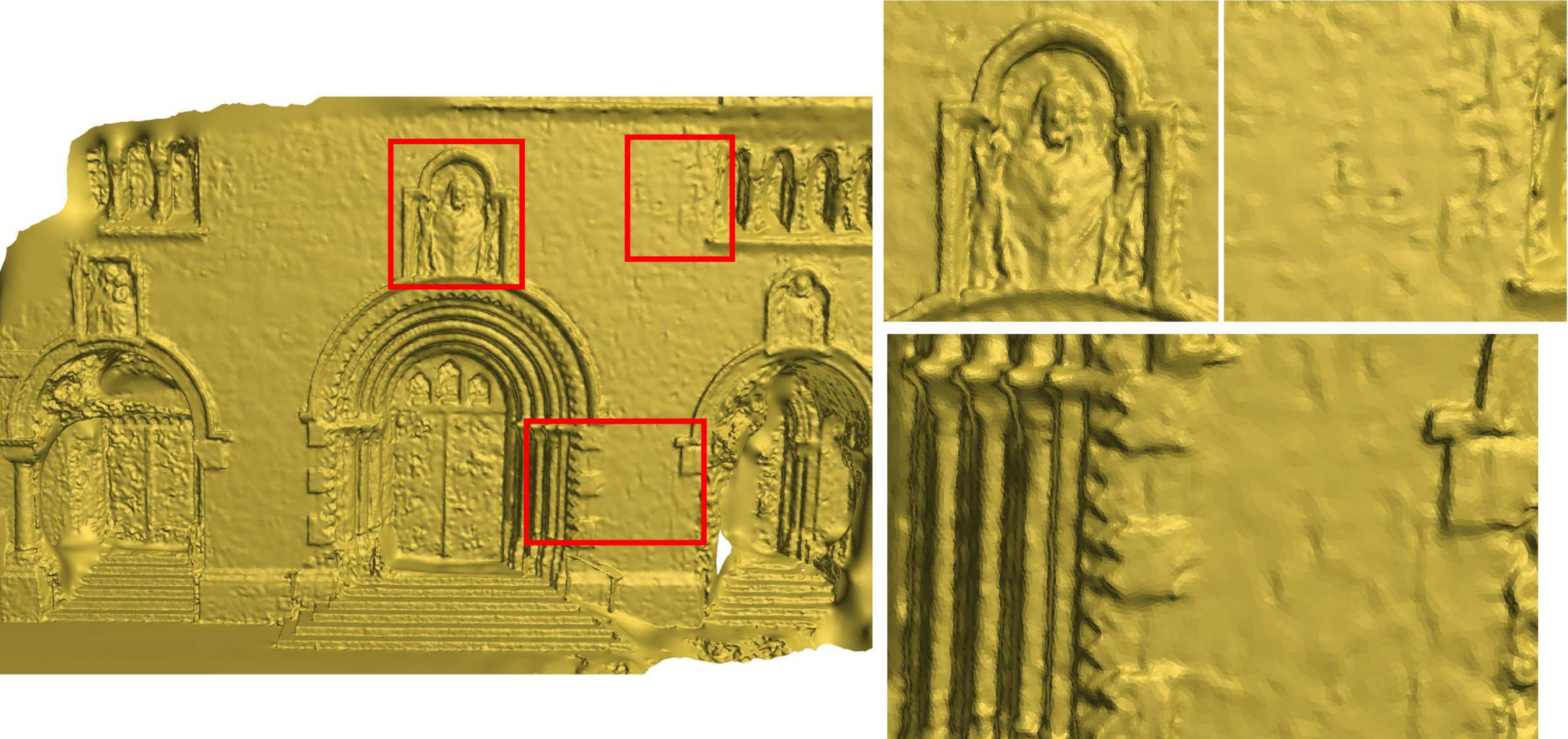}}
\subfigure[]{
\label{fig:subfig:resultbunnyd}
\includegraphics[height=25mm]{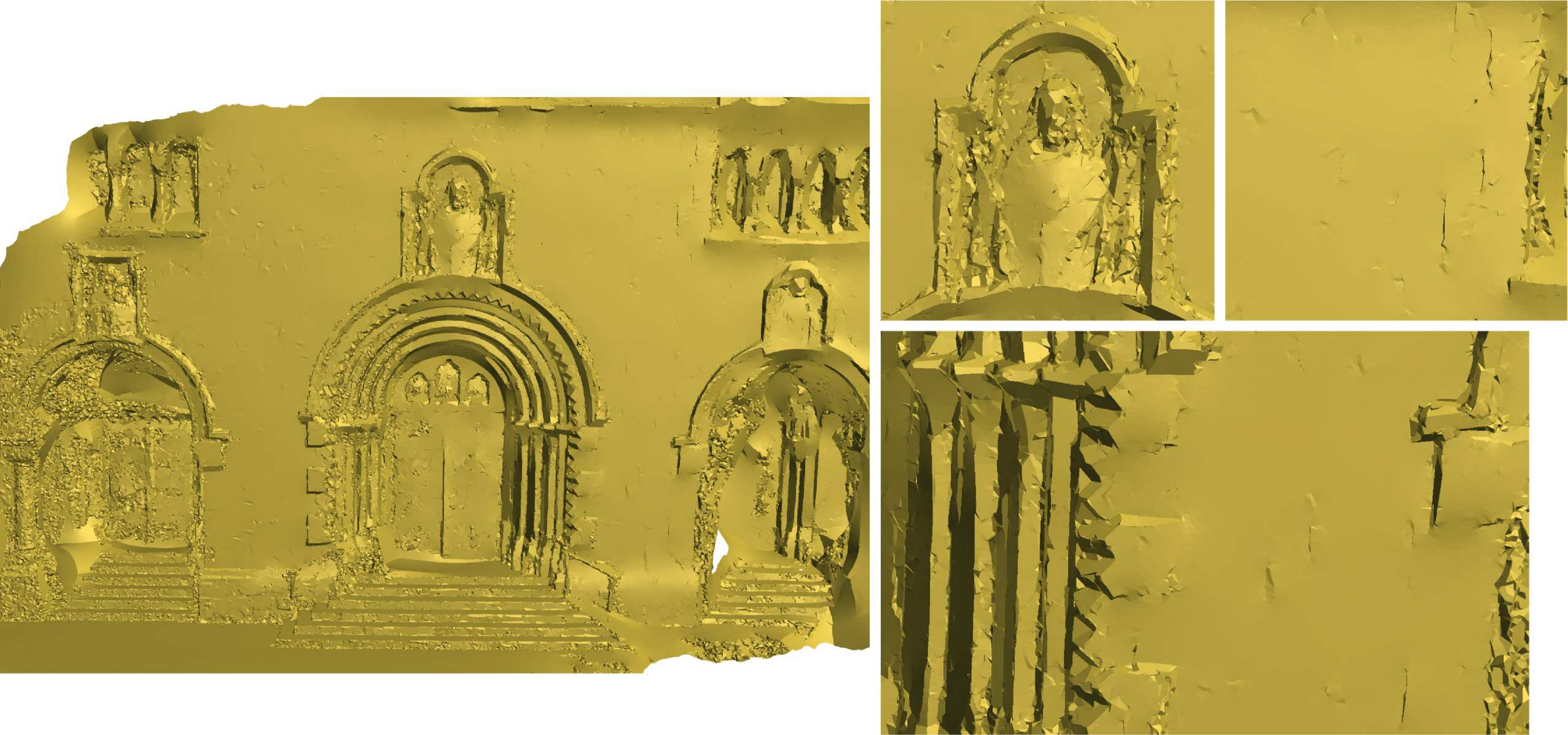}}
\subfigure[]{
 \label{fig:subfig:resultbunnye}
\includegraphics[height=25mm]{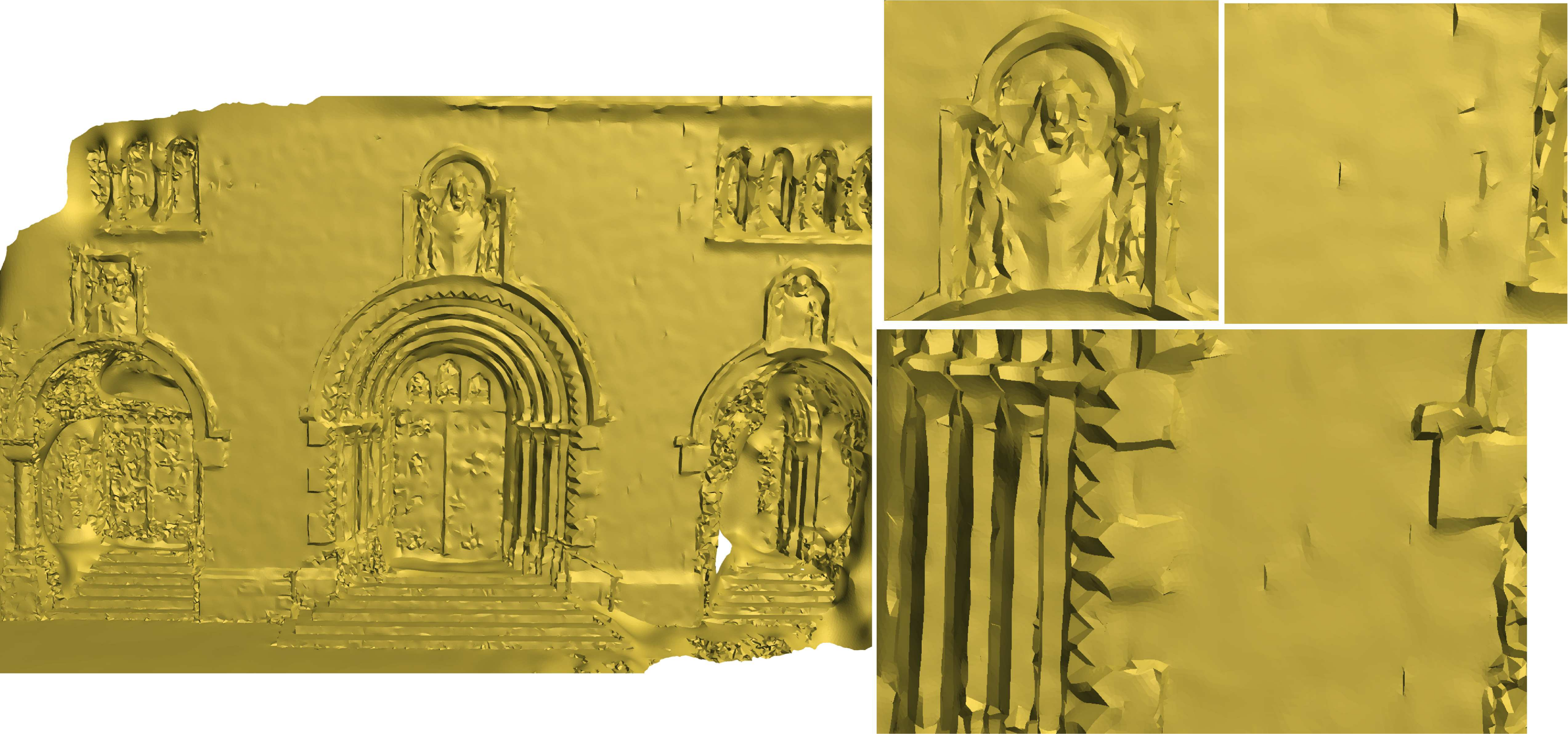}}
 \subfigure[]{
  \label{fig:subfig:resultbunnyf}
\includegraphics[height=25mm]{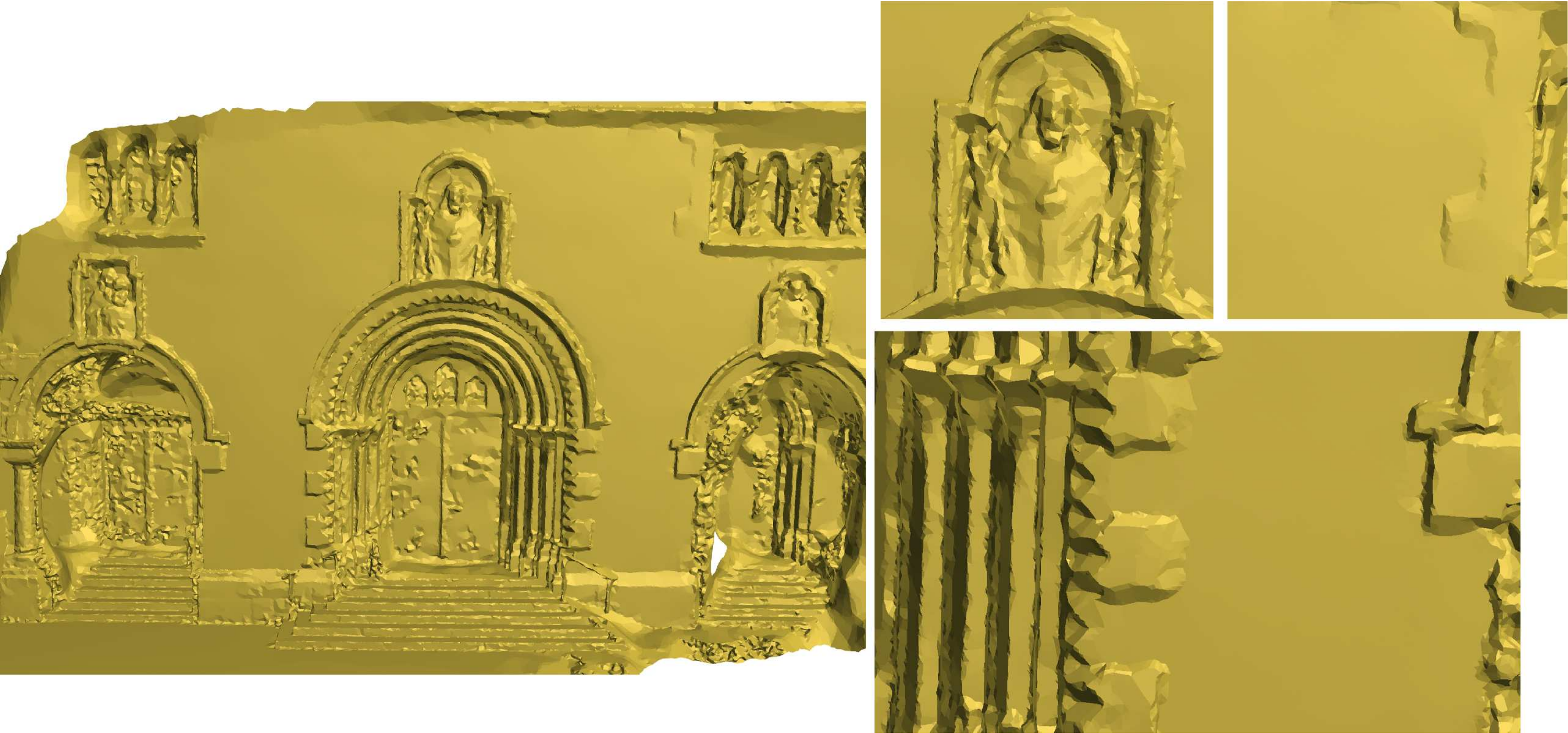}}
\subfigure[]{
\label{fig:subfig:resultbunnyg}
\includegraphics[height=25mm]{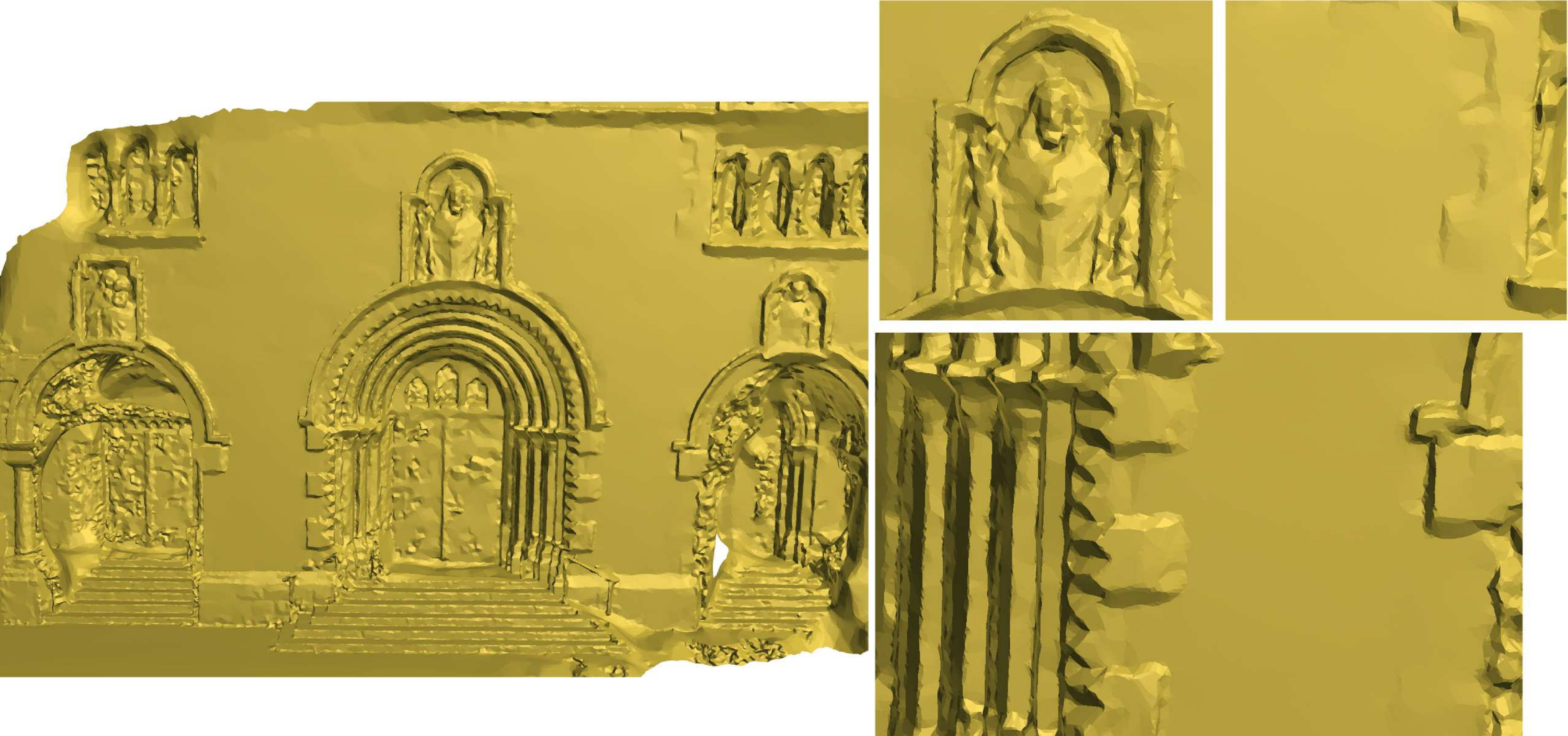}}
\caption{Results by using different mesh denoising methods on the {\it Herzjesu-P8} dataset. (a) Input images. (b) Results by the combination of first order and second order Laplacian~\cite{18,31a}. (c) Results by Sun et al.'s method~\cite{28}. (d) Results by Zhang et al.'s bilateral normal filtering \cite{29}. (e) Results by He et al.'s $L_{0}$ denoising \cite{30}. (f) Results by our $L_{p}$ denoising method. All the models are flat-shaded to show faceting effect.}
\label{fig:resultherzjesu}
 \vspace{-1mm}
\end{figure*}

\begin{figure*}[tbp]
\centering
\subfigure[]{
\label{fig:subfig:resultflowera}
\includegraphics[height=24mm]{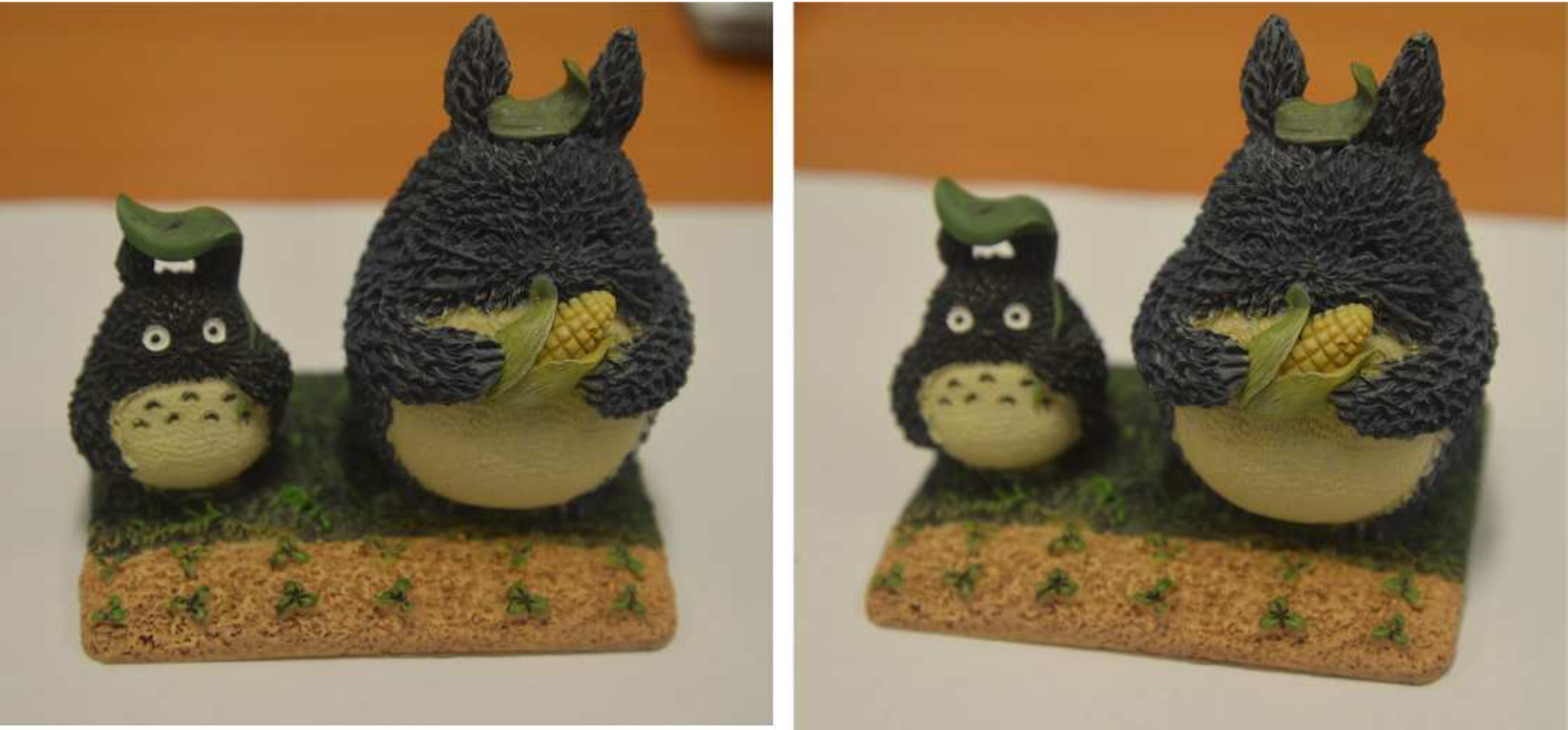}}
 \subfigure[]{
  \label{fig:subfig:resultflowerb}
\includegraphics[height=25mm]{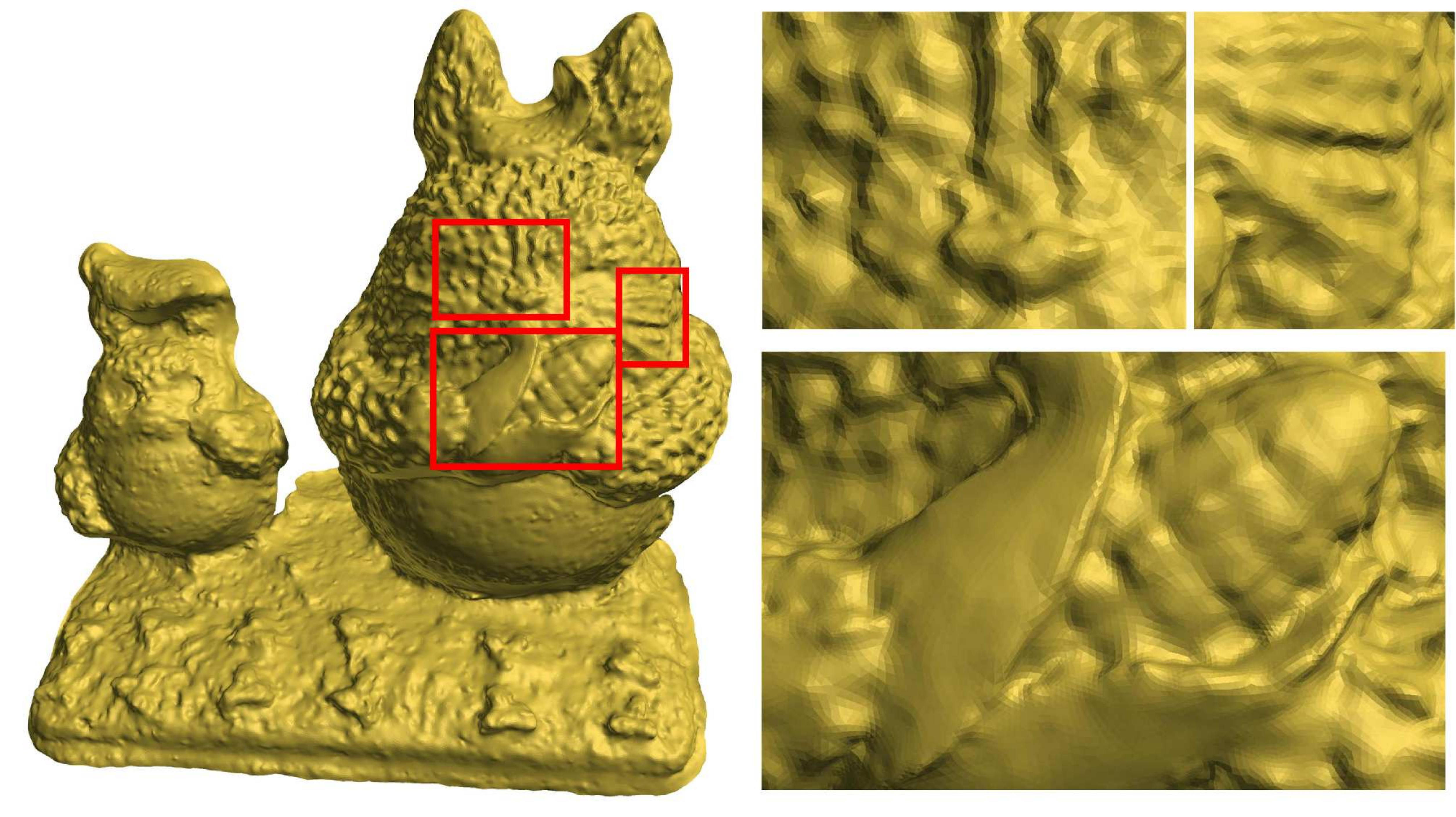}}
 \subfigure[]{
  \label{fig:subfig:resultflowerc}
\includegraphics[height=25mm]{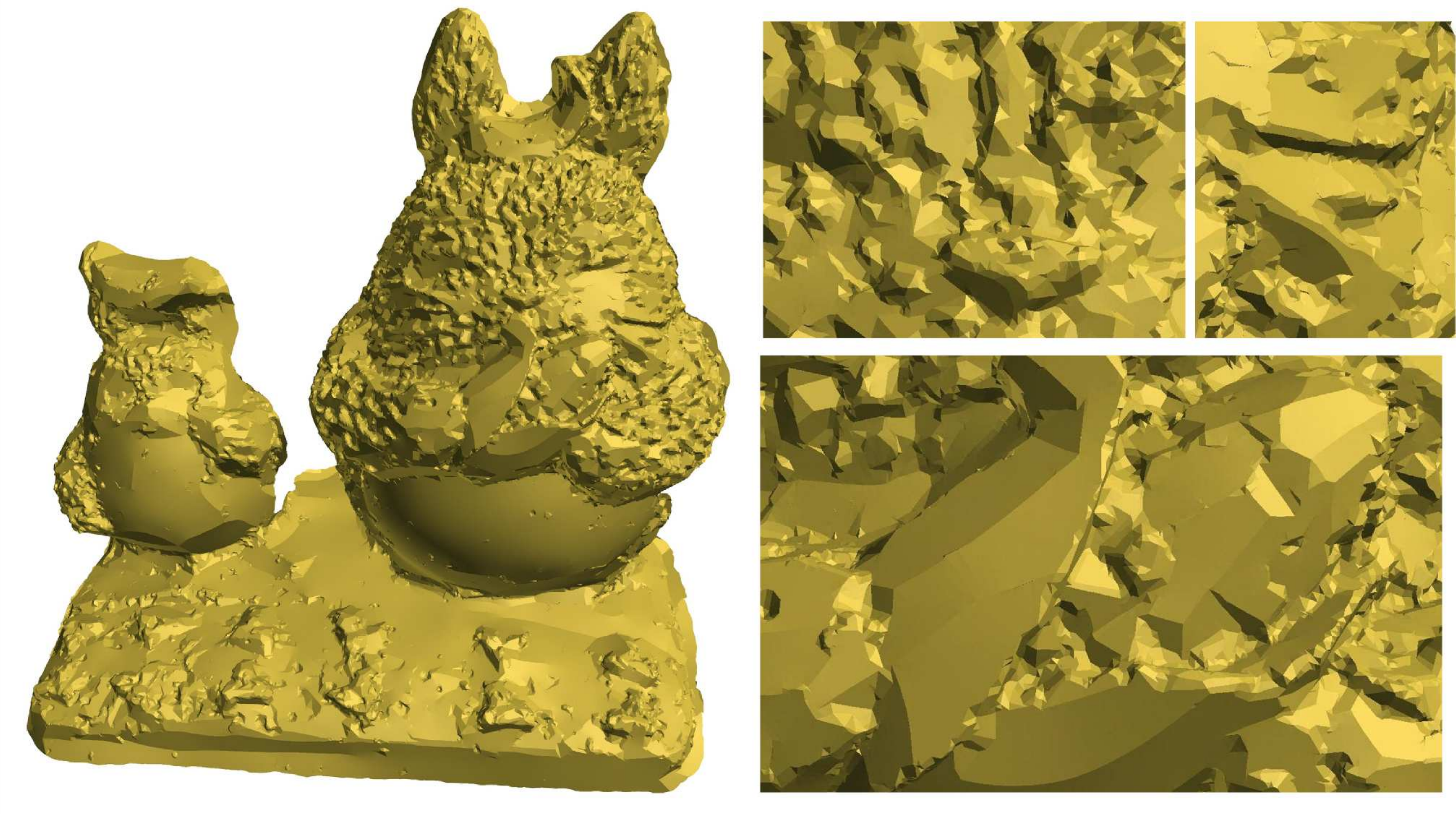}}
\subfigure[]{
 \label{fig:subfig:resultflowere}
\includegraphics[height=25mm]{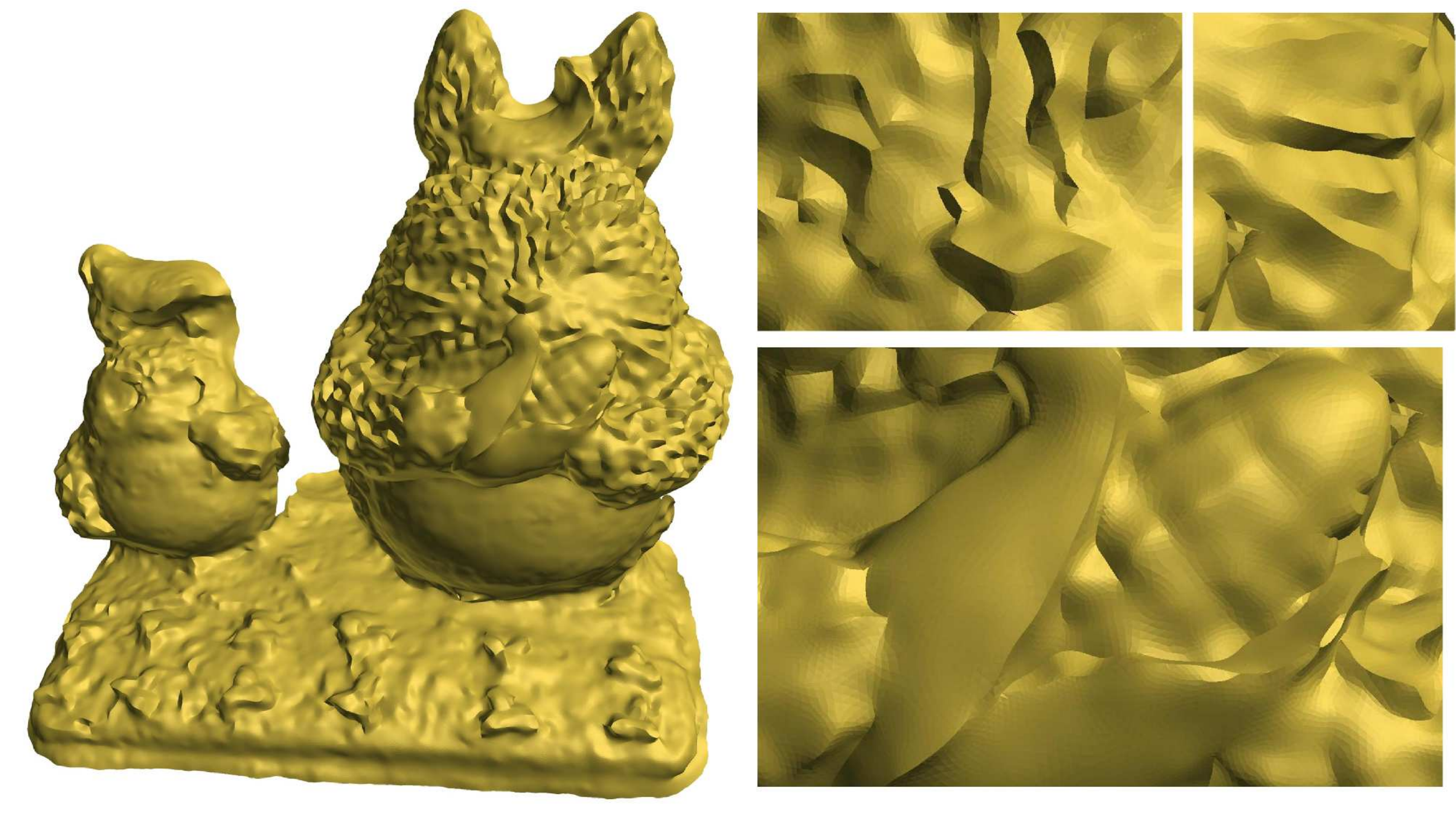}}
 \subfigure[]{
  \label{fig:subfig:resultflowerf}
\includegraphics[height=25mm]{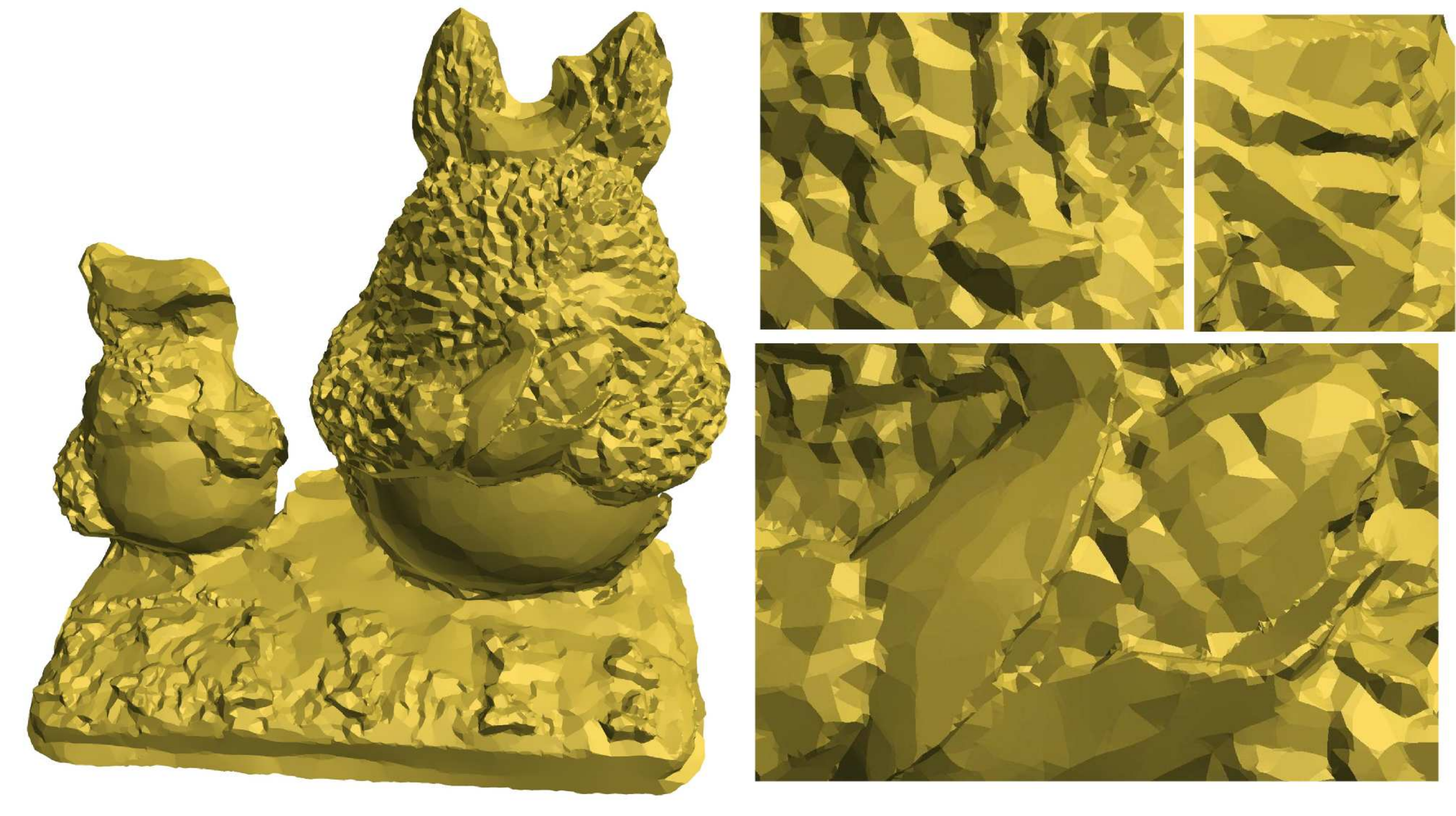}}
\subfigure[]{
\label{fig:subfig:resultflowerg}
\includegraphics[height=25mm]{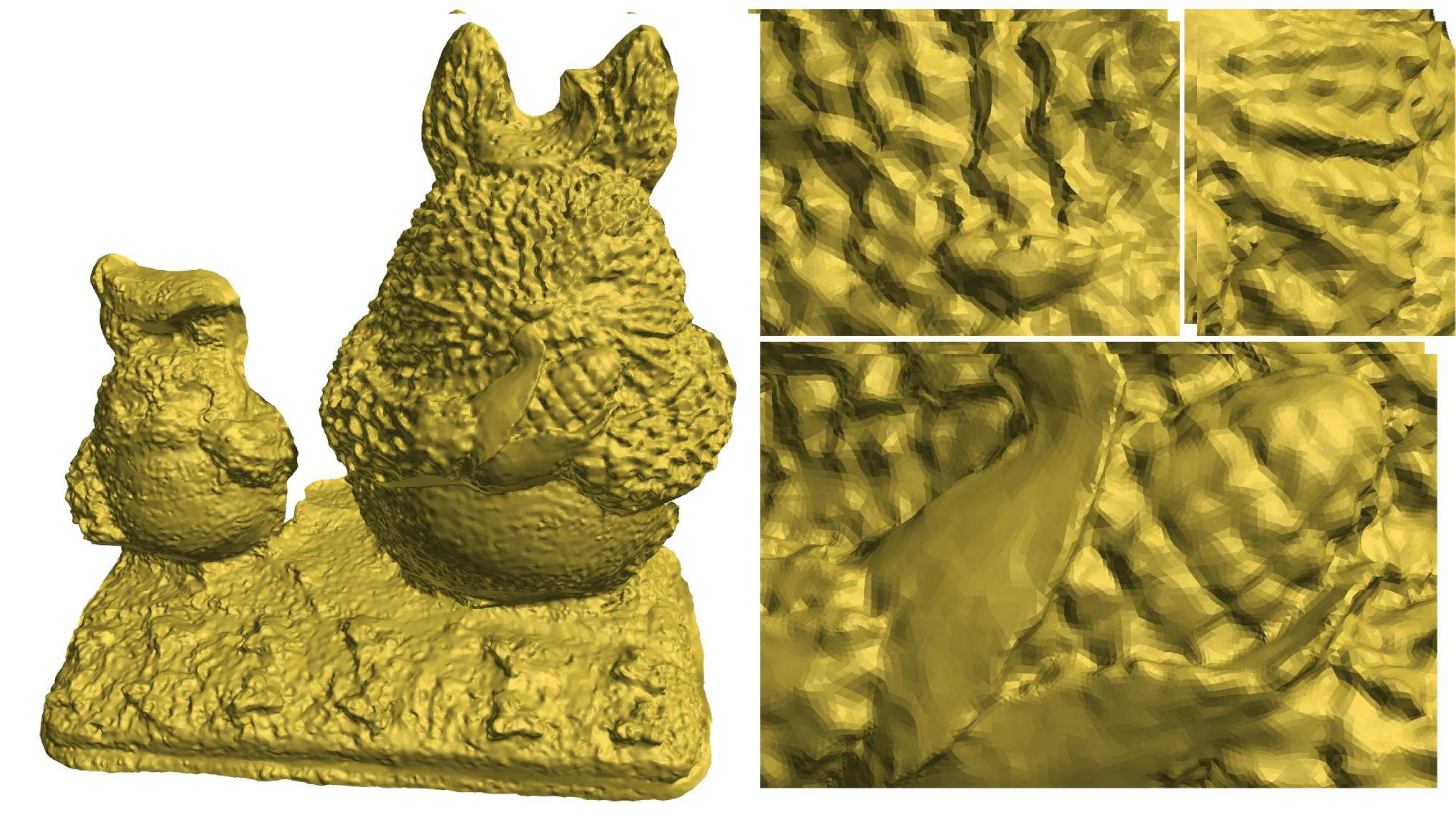}}
\vspace{-3mm}
\caption{Results by using different mesh denoising methods on the {\it Totoro} dataset. (a) Input images. (b) Results by the combination of first order and second order Laplacian~\cite{18,31a}. (c) Results by Sun et al.'s method~\cite{28}. (d) Results by Zhang et al.'s bilateral normal filtering \cite{29}. (e) Results by He et al.'s $L_{0}$ denoising \cite{30}. (f) Results by our content-aware $L_{p}$ denoising method. All the models are flat-shaded to show faceting effect.
}
\label{fig:resultpusa}
\vspace{-3mm}
\end{figure*}

\subsection{Evaluation on Content-aware Mesh Denoising}
Our DCV method consists of two components, i.e., detail-preserving similarity measure and content-aware $L_{p}$ mesh denoising. The effectiveness of the former component has been validated in Fig. \ref{fig:detailenhancefig} by comparing with isotropic ZNCC measure. To evaluate the effectiveness of the latter component, we implement four variants of DCV by substituting the content-aware $L_{p}$ mesh denoising with four competing denoising methods, including the isotropic mesh smoothing method (i.e., the one based on the combination of first order and second order Laplacian \cite{31a}) and the anisotropic mesh denoising methods (i.e., two-step normal filtering \cite{28}, bilateral normal filtering \cite{29}, and $L_{0}$ mesh denoising \cite{30}). Two datasets, i.e., \emph{Herzjesu-P8} and \emph{Totoro}, are used for evaluating the mesh denosing methods.

The \emph{Herzjesu-P8} dataset contains eight $3072\times 2048$ images of a building with very sparse sharp edges (porch and stairs) and many flat regions (wall). The \emph{Totoro} dataset contains eight $1504\times 1004$ images of a plastic status with many fine-scale details (grass in the ground, whiskers, fur and corn). As shown in Fig. \ref{fig:resultherzjesu}, on the \emph{Herzjesu-P8} dataset, the anisotropic methods have better performance than isotropic methods, and our content-aware $L_{p}$ denoising method achieves similar results to He et al.'s $L_{0}$ denoising method but is visually more pleasant. As shown in Fig. \ref{fig:resultpusa}, on the \emph{Totoro} dataset, our content-aware $L_{p}$ denoising method obtains much better results than the other methods. Unlike the competing denoising methods, the proposed $L_{p}$ denoising method is content-aware and is able to reconstruct the object with flat regions, sharp edges, and fine-scale details.

\section{Conclusion}
\label{section6}

In this paper, we proposed a  detail-preserving and content-aware variational (DCV) method for multi-view stereo (MVS) reconstruction. First, by connecting guided image filtering with image registration, a novel similarity measure was proposed to preserve the fine-scale details in reconstruction. Second, by the hyper-Laplacian modelling of surface gradients, a content-aware mesh denoising method based on $L_{p}$ minimization was presented to suppress the noises and outliers while preserving sharp features. Compared with state-of-the-art MVS methods, the proposed DCV method is capable of reconstructing a smooth and clean surface with finely preserved details and sharp features. The running time of our single-thread CPU implementation of DCV method on the datasets used in the paper is from twenty minutes to several hours. In the future, GPU-based parallel implementation on the main parts of gradient computation will be adopted to improve the speed of DCV by using the Nvidia CUDA framework.

\section*{Acknowledgement}
The authors would like to thank Dr. Daniel Scharstein for evaluating our results on the Middlebury datasets, and Dr. Hoang Hiep Vu for providing the {\it statuegirl} dataset.
\ifCLASSOPTIONcaptionsoff
  \newpage
\fi

\end{document}

%% file: table2_DeCoS_double.tex
\begin{table}[htbp]
\centering
\caption{Datasets used in our experiments}
\footnotesize
\label{table2_DeCoS}
\tabcolsep=0.06in
\begin{tabular}{|>{\centering}p{1.7cm}|>{\centering}p{1.3cm}|>{\centering}p{1.7cm}|>{\centering}p{1.7cm}|>{\centering}p{1cm}|}
\hline 
Dataset & Number of images & Resolution & Initialization & Time (min)\tabularnewline
\hline 
dino sparse & 16 & 640$\times$480 & visual hull & 90\tabularnewline
dino ring & 48 & 640$\times$480 & visual hull & 150\tabularnewline
temple sparse & 16 & 640$\times$480 & pmvs+psr & 105\tabularnewline
temple ring & 47 & 640$\times$480 &  pmvs+psr & 170\tabularnewline
Beethoven & 33 & 1024$\times$768 & visual hull & 180\tabularnewline
bird & 21 & 1024$\times$768 & visual hull & 160\tabularnewline
fountain-P11 & 11 & 3072$\times$2048 & pmvs+psr & 210\tabularnewline
Herzjesu-P8 & 8 & 3072$\times$2048 & pmvs+psr & 150\tabularnewline
Totoro  & 8 & 1504$\times$1004 & pmvs+psr & 45\tabularnewline
Buddha & 5 & 2400$\times$1800  & pmvs+psr  & 30\tabularnewline
bell  & 3  &  1504$\times$1004 & pmvs+psr & 20\tabularnewline
statuegirl  & 50  &  2592$\times$3888 & pmvs+psr & 750\tabularnewline
\hline 
\end{tabular}
\vspace{-0.6cm}
\end{table}

%% file: table3_DeCoS.tex
\begin{table*}[htbp]
\centering
\caption{Quantitative comparison between DCV and several state-of-the-art methods on the Middlebury data sets in terms of accuracy/completeness}
\footnotesize
\label{table3_DeCoS}
\tabcolsep=0.06in
\begin{tabular}{|>{\centering}p{2.5cm}|>{\centering}p{2cm}|>{\centering}p{2cm}|>{\centering}p{2cm}|>{\centering}p{2.5cm}|}
\hline 
Method & dino sparse & dino ring & temple sparse & temple ring \tabularnewline
\hline 
DCV               & \textbf{0.3mm/100\%}    & \textbf{0.28mm/100\%}   & 0.66mm/97.3\% & 0.73mm/98.2\%\tabularnewline
Vu\cite{20}         & $-$       & 0.53mm/99.7\%  & $-$      & \textbf{0.45mm/99.8\%}\tabularnewline
Kostrikov \cite{2d} & 0.37mm/99.3\%  & 0.35mm/99.6\%  & 0.79mm/95.8\% & 0.57mm/99.1\%\tabularnewline
Zaharescu\cite{18}  & 0.45mm/99.2\%  & 0.42mm/98.6\%  & 0.78mm/95.8\% & 0.55mm/99.2\%\tabularnewline
Kolev2\cite{16a}      & 0.53mm/98.3\%  & 0.43mm/99.4\%  & 1.04mm/91.8\% & 0.72mm/97.8\%\tabularnewline
Kolev3\cite{16}      & 0.48mm/98.6\%  & 0.42mm/99.5\%  & 0.97mm/92.7\% & 0.7mm/98.3\%\tabularnewline
Gargallo\cite{23}      & 0.76mm/90.7\%  & 0.6mm/92.9\%  & 1.05mm/81.9\% & 0.88mm/84.3\%\tabularnewline
Delaunoy\cite{24}   & 0.89mm/93.9\%  & $-$       & 0.73mm/95.9\% & $-$\tabularnewline
Furukawa3\cite{31a}  & 0.37mm/99.2\%  & 0.28mm/99.8\%  & \textbf{0.63mm/99.3\%} & 0.47mm/99.6\%\tabularnewline
\hline 
\end{tabular}
\vspace{-0.0cm}
\end{table*}